\documentclass[journal]{IEEEtran}

\usepackage{graphicx}
\usepackage{amssymb}
\usepackage{multirow}
\usepackage{booktabs}
\usepackage{verbatim}
\usepackage[colorlinks,linkcolor=black, citecolor=black]{hyperref}
\begin{document}

\title{Single Image Dehazing with \\ An Independent Detail-Recovery Network}

\iftrue
\author{Yan Li,
        De Cheng$^{\sharp}$,
        Jiande Sun$^{\sharp}$,
        Dingwen Zhang, 
        Nannan~Wang~\IEEEmembership{Member,~IEEE,}
        Xinbo~Gao~\IEEEmembership{Senior Member,~IEEE}
\thanks{Yan Li, Jiande Sun, are with the School of Information Science and Engineering, Shandong Normal University, Jinan 250358, China.}
\thanks{De Cheng, Nannan Wang are with the Department of Communication Engineering, Xidian University, Xi'an 710071, China.}
\thanks{Dingwen Zhang is with the School of Automation, Northwestern University, Xi'an 710071, China.}
\thanks{Xinbo Gao is with the Chongqing University of Posts and Telecommunications, Chongqing 400065, China.}
\thanks{The corresponding authors are: De Cheng$^{\sharp}$ and Jiande Sun{$^{\sharp}$}.}}
\fi
%



\maketitle

\begin{abstract}
Single image dehazing is a prerequisite which affects the performance of many computer vision tasks and has attracted increasing attention in recent years. However, most existing dehazing methods emphasize more on haze removal but less on the detail recovery of the dehazed images. In this paper, we propose a single image dehazing method with an independent Detail Recovery Network (DRN), which considers capturing the details from the input image over a separate network and then integrates them into a coarse dehazed image. The overall network consists of two independent networks, named DRN and the dehazing network respectively. Specifically, the DRN aims to recover the dehazed image details through local and global branches respectively. The local branch can obtain local detail information through the convolution layer and the global branch can capture more global information by the Smooth Dilated Convolution (SDC). The detail feature map is fused into the coarse dehazed image to obtain the dehazed image with rich image details. Besides, we integrate the DRN, the physical-model-based dehazing network and the reconstruction loss into an end-to-end joint learning framework. Extensive experiments on the public image dehazing datasets (RESIDE-Indoor, RESIDE-Outdoor and the TrainA-TestA) illustrate the effectiveness of the modules in the proposed method and show that our method outperforms the state-of-the-art dehazing methods both quantitatively and qualitatively. The code is released in \url{https://github.com/YanLi-LY/Dehazing-DRN}.
\end{abstract}


\section{Introduction}
\IEEEPARstart{S}{ingle} image dehazing aims to recover the clean image from the given hazy image.
Usually, the hazy images are subject to blurring, low contrast, color distortion, and other visible quality degradation. Taking those degraded images as inputs will impede the real performance of many tasks in the computer vision research fields. 
Generally, image dehazing acts as a prerequisite and can affect a wide range of subsequent applications, such as content-based image retrieval \cite{min2020two}, object detection \cite{ren2020salient}, and so on \cite{li2020bottom, yao2020adaptive}.
Hence, image dehazing has become one hot topic in recent years.  

In the image dehazing literatures \cite{he2010single, cai2016dehazenet, wang2017fast, ling2017optimal}, the atmospheric scattering model \cite{mccartney1976optics} is widely used to describe the formation of a hazy image:
\begin{equation}
I(x) = J(x)t(x) + A(x)(1-t(x)), \label{eq:1}
\end{equation}
where $x$ denotes the pixel of the image, $I(x)$ denotes the observed hazy image, $J(x)$ denotes the haze-free image, $A(x)$ denotes the global atmospheric light which is a 2D map, and $t(x)$ denotes the transmission map. 
Moreover, the transmission map can be represented as $t(x) = e^{- \gamma d(x)}$,
where $d(x)$ and $\gamma$ denote the scene depth and the atmospheric scattering coefficient, respectively. Image dehazing aims to recover the haze-free image $J(x)$ from the given hazy image $I(x)$.
We can rewrite Eq. (\ref{eq:1}) as:
\begin{equation}
J(x) = \frac{I(x)-A(x)}{t(x)} + A(x). \label{eq:2}
\end{equation}

In recent decades, a series of methods \cite{tan2008visibility, he2010single, shen2018iterative, li2019pdr, dong2020multi} have been proposed to tackle the single image dehazing problem. These methods can be broadly segregated into two classes, which are prior-based methods and learning-based methods. In prior-based methods \cite{fattal2014dehazing, he2010single, zhu2015fast}, the transmission map $t(x)$ and the global atmospheric light $A(x)$ are characterized by hand-crafted priors, which are manufactured by exploiting the statistical properties of clean images. These priors are supposed to distinguish between hazy images and haze-free images, such as dark-channel prior \cite{he2010single}, color attenuation prior \cite{zhu2015fast}, haze-line prior \cite{berman2016non}, and color-lines \cite{fattal2014dehazing}. Although prior-based methods work well in some scenarios, there are the others in which they are ineffective. For example, dark channels prior don't work for white objects or scenes which are closer to atmospheric light, where it is difficult to accurately estimate the transmission map. 

With the development of deep Convolutional Neural Network (CNN) and the emergence of large-scale datasets, learning-based image dehazing methods have gradually attracted more and more attentions. Learning-based methods \cite{cai2016dehazenet, li2017aod, zhang2018multi, liu2019griddehazenet, dong2020multi} can be further classified into two branches. Early works \cite{li2018single, ren2018gated, zhang2018multi, engin2018cycle, liu2019griddehazenet} are based on physical models, which learn the transmission map from the training data and then the haze-free image can be recovered via the physical atmospheric scattering model. These methods have largely overcome the shortcomings of prior-based methods. Since the encoder-decoder structure has been proved to achieve good results in image reconstruction fields \cite{mao2016image, ren2018gated}, many methods \cite{zhang2018densely, li2020zero} adopt the encoder-decoder structure as the estimation network for the transmission map. The skip connections are usually established between the encoder and decoder in order to enable the decoder to take full advantage of the information in the encoder network. Therefore, we follow the network proposed by \cite{zhang2018densely} as our dehazing network. By means of the shortcut connection, the feature information of shallow layers can be transmitted to the deep layers, so as to realize the supplement of lost information. In addition, many methods lack the learning of atmospheric light and can not accurately estimate it, which becomes a key obstacle to the performance improvement for physical-model-based haze removal algorithms. Therefore, we raise awareness of this deficiency and attempt to correct it in our method. We choose a simple but effective U-Net \cite{ronneberger2015u} to learn the atmospheric light. Because there is less high-level semantic information in atmospheric light, low-level features are particularly important. U-Net combines high-level features and low-level features to learn atmospheric light more accurately. 

Subsequently, some end-to-end learning-based methods have been proposed, these methods \cite{cai2016dehazenet, ren2016single, zhang2018densely, li2017aod} leverage CNN to extract features and learn the mappings from hazy images to haze-free images by directly using enormous training data. For both types of learning-based methods, we choose the learning-based dehazing model based on the physical atmosphere scattering model because the physical model can specify task-specific optimization objectives for the network training and make the entire training process explicable.

Though tremendous improvements have been made in existing methods, there still contain several limitations that hinder their performance. Most existing methods only devote themselves to remove haze but lose sight of the detail recovery. Although some approaches \cite{li2020deep, gao2018detail, ju2019idgcp, qu2019enhanced} allow for the measures in the dehazing backbone network to preserve the details of dehazed images, the dehazing performance is limited. Because there is image information lost in the dehazing process, it is difficult to retain plenty of image details in the dehazing backbone network. In this paper, we propose a single image dehazing method with an independent detail-recovery network. Different from previous methods, the designed DRN is independent of the dehazing backbone, which takes the hazy image as input and outputs the detail feature map. Then we fuse the detail feature map into the coarse dehazed image from the dehazing backbone network and finally output a high-quality dehazed image. 

The main contribution of this work can be concluded as follows:
\begin{itemize}
	\item[$\bullet$]We build an independent detail recovery network (DRN) to recover the intrinsic image details contained in the hazy images, which consists of the convolution-based local branch and the dilated convolution-based global branch, to extract the local and global image contextual features at the same time. It succeeds in recovering the original image details clearly by detail feature obtained from the DRN.
	\item[$\bullet$]We integrate the proposed DRN, the physical atmospheric scattering model and the reconstruction loss into an end-to-end joint learning framework. In particular, our proposed method applies a multi-faceted loss function, which not only includes the pixel level loss and the feature level loss, but also considers the loss between reconstructed hazy images and input image i.e. reconstruction loss. Reconstruction loss is beneficial to the improvement of dehazing performance and the stability of network training.
	\item[$\bullet$]Extensive experiments conducted on both synthetic datasets and real-world hazy images, demonstrate the effectiveness of the proposed method and show that the proposed method performs favorably against the state-of-the-art dehazing approaches both in terms of qualitative and quantitative metrics.
\end{itemize}

\begin{figure}[]
	\centering  
	\includegraphics[width=2cm,height=2cm]{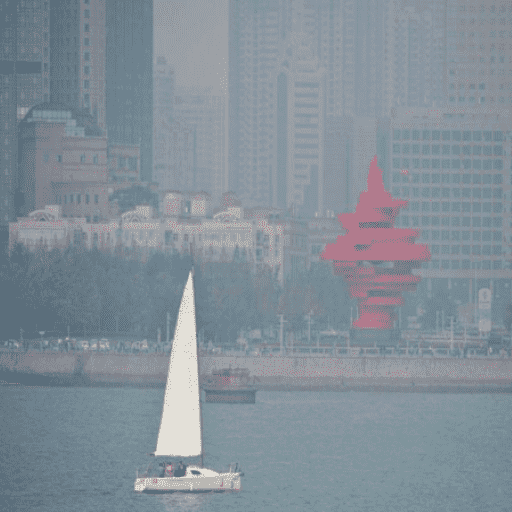}  
	\includegraphics[width=2cm,height=2cm]{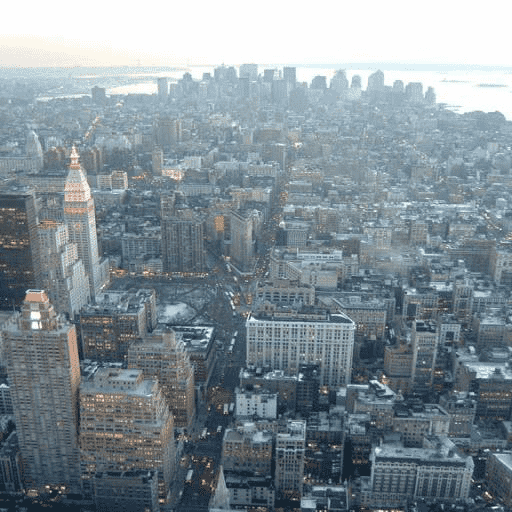}
	\includegraphics[width=2cm,height=2cm]{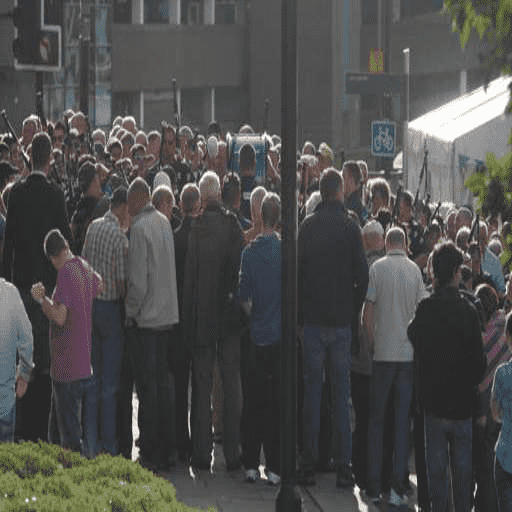}  
	\includegraphics[width=2cm,height=2cm]{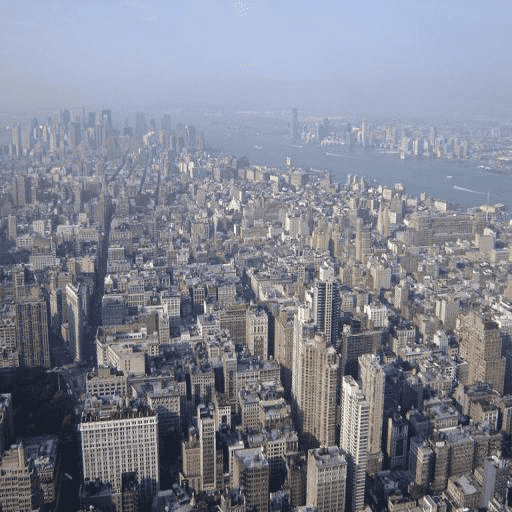}			
	\\
	\vspace{3pt}
	\includegraphics[width=2cm,height=2cm]{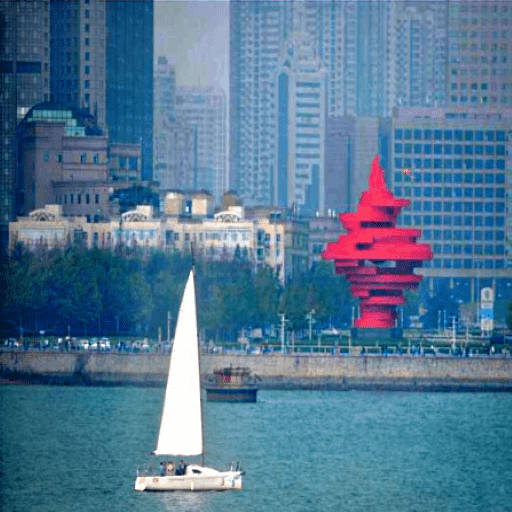} 
	\includegraphics[width=2cm,height=2cm]{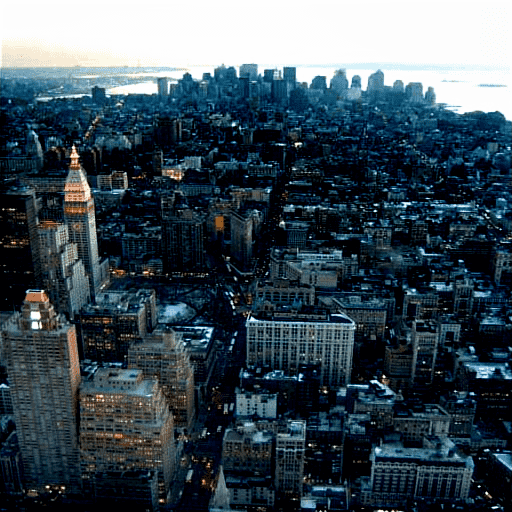} 
	\includegraphics[width=2cm,height=2cm]{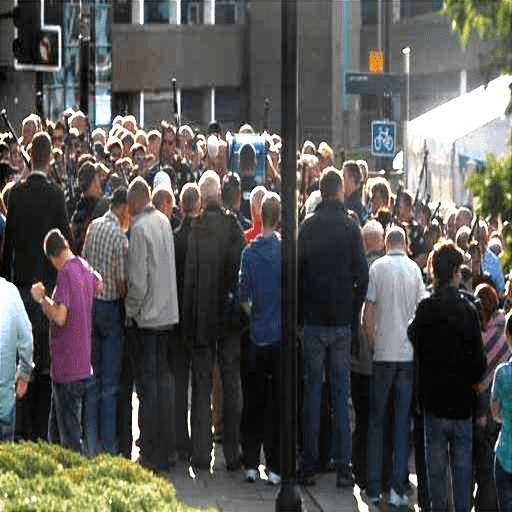}
	\includegraphics[width=2cm,height=2cm]{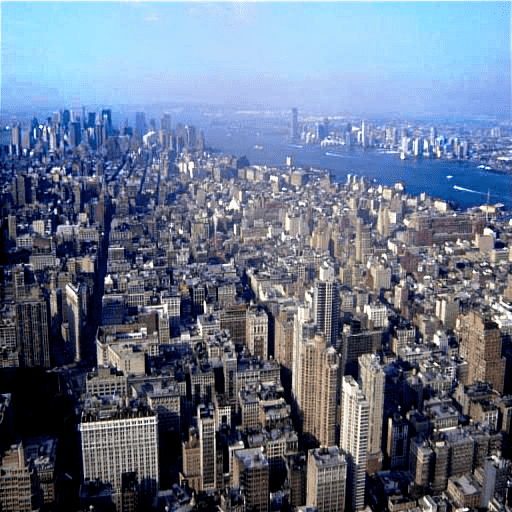}
	\caption{The dehazing examples on our proposed method. The top line is the input hazy images and the bottom line is the dehazed images.}
	\label{haze-dehaze}
\end{figure}

\section{Related Work}
In the past, a variety of prior-based \cite{tan2008visibility, fattal2008single, fattal2014dehazing, he2010single, zhu2015fast} and learning-based \cite{li2017aod, zhang2018densely, mei2018progressive, qin2020ffa, dong2020multi} methods have been proposed to handle the single image dehazing problem. In this section, we retrospect related works on single image dehazing, starting from the traditional prior-based methods to the recent learning-based methods.

\subsection{Prior-based methods}
In many prior-based methods, the priors that can distinguish the haze-free image from the hazy image are designed as constraints for image dehazing. Tan et al. \cite{tan2008visibility} propose the image dehazing method based on Markov Random Fields, which is based on the observation that haze-free images are apt to have more contrast than their hazy counterpart and attempts to remove haze by maximizing the local contrast of each patch in the hazy image. The image dehazing method proposed by Fattal et al. \cite{fattal2008single} estimates the albedo of the scene under the prior that transmission map is not correlated with surface shading. He et al. \cite{he2010single} present the Dark Channel Prior (DCP) for the estimation of transmission map, which asserts that in the case of haze-free image patches, some pixels have low intensity even close to zero in at least one color channel. Zhu et al. \cite{zhu2015fast} create a linear model to recover depth information of hazy images via color attenuation prior. The color-line prior \cite{fattal2014dehazing} is introduced for image dehazing based on the observation that small patches of image typically exhibit a one-dimensional distribution in the RGB color space. Even if the prior-based methods have achieved certain results, these priors are designed manually according to the statistical features of images, which cannot completely meet the inherent attributes of images. Therefore, these methods do not apply to all scenarios and will be limited in some cases.

\subsection{Learning-based methods}
\textbf{Pysical-model-based methods.}
Since CNN has achieved exemplary achievements in the field of computer vision, it has also been widely used for image reconstruction including image dehazing. Early works are based on the physical model, which first estimate the transmission map and global atmospheric light and then reconstruct the haze-free image based on the atmospheric scattering model. Ren et al. \cite{ren2016single} design a coarse-to-fine Multi-Scale Convolutional Neural Network (MSCNN) to estimate the transmission map and then refine it locally. Cai et al. \cite{cai2016dehazenet} create a novel CNN architecture, named DehazeNet, to estimate the transmission map of hazy image. All-in-One Dehazing Network (AOD-Net) \cite{li2017aod} integrates the transmission map and the atmospheric light into one variable and then estimates it. Zhang et al. \cite{zhang2018densely} develop a Densely Connected Pyramid Dehazing Network (DCPDN) to estimate the transmission map, atmospheric light, and dehazed image. Some of the earlier approaches design simpler networks for transmission map estimation, with performance limitations. Latter methods are developed to design more accurate networks. Many methods adopt the encoder-decoder structure as the estimation network for transmission map and supplement the information in the encoder to the decoder through various connections to alleviate the information loss. Hence, the encoder-decoder structure with residual connections is used in our dehazing network which is capable to get the utmost information and transmit feature information from the shallow layers to the deep layers. In addition, many methods only focus on the estimation of transmission map without paying attention to the accuracy of atmospheric light, which becomes an important factor that hinders the performance of dehazing. And unlike other methods, we choose a simple but effective U-Net \cite{ronneberger2015u} to learn the atmospheric light. Since U-Net can combine high-level features with low-level features, it is advantageous for atmospheric light, which has less semantic information but more low-level features.

\textbf{End-to-end methods.}
Recently, end-to-end CNN-based dehazing methods have been designed to directly learn the haze-free image from a hazy image without the estimation of the transmission map and the atmospheric light. Ren et al. \cite{ren2018gated} propose a Gated Fusion Network (GFN) with a novel fusion strategy that takes the original hazy image and its three derived images as input by applying White Balance, Contrast Enhancing, and Gamma Correction. Mei et al. \cite{mei2018progressive} design a Progressive Feature Fusion Network (PFFNet) that directly learns the nonlinear transformation function from observed hazy image to haze-free one. Enhanced Pix2pix Dehazing Network (EPDN) 
 \cite{qu2019enhanced} attempts to follow the dehazing network with an enhancer to improve the dehazing performance. Dong et al. \cite{dong2020physics} propose an effective Feature Dehazing Unit (FDU) to explore useful features in the deep feature space. Liu et al. \cite{liu2019griddehazenet} propose GridDehazeNet by integrating multi-scale estimation with the attention mechanism. Qin et al. \cite{qin2020ffa} propose an end-to-end feature fusion attention network, whose Attention-based Feature Fusion (FFA) structure can retain the information from shallow layers and pass it into deep layers. Dong et al. \cite{dong2020multi} propose the Multi-Scale Boosted Dehazing Network (MSBDN), which incorporates the boosting strategy and the back-projection technique for image dehazing. Dong et al. \cite{dong2020fd} propose a Generative Adversarial image dehazing Network with Fusion-Discriminator (FD-GAN), which treats the frequency information as extra constraints. End-to-end learning is a blind learning process without the support of the physical model. The physical model can specify task-specific optimization objectives for the network training process and makes the entire training process explicable. 

The learning-based methods are no longer bound by the priors, so they overcome the shortcomings of the prior-based methods. But in spite of that, they only focus on the dehazing process and ignore the detail recovery of dehazed images. A few approaches consider the detail-preserving measures in the dehazing backbone network. For example, EPDN \cite{qu2019enhanced} feeds the dehazed image into an enhancer in an attempt to enhance the details of the dehazed image. However, the input hazy image will inevitably be accompanied by details loss in the dehazing process, so it is difficult to retain information in the dehazing process or to recover it after dehazing. Different from existing methods, we build an independent DRN to recover the intrinsic image details contained in the hazy images, which takes the hazy image as input and outputs the detail feature map. The high-quality dehazed image is obtained by consolidating the coarse dehazed image with the detail feature map.

\begin{figure*}[t]
	\centering
	\includegraphics[width=\linewidth]{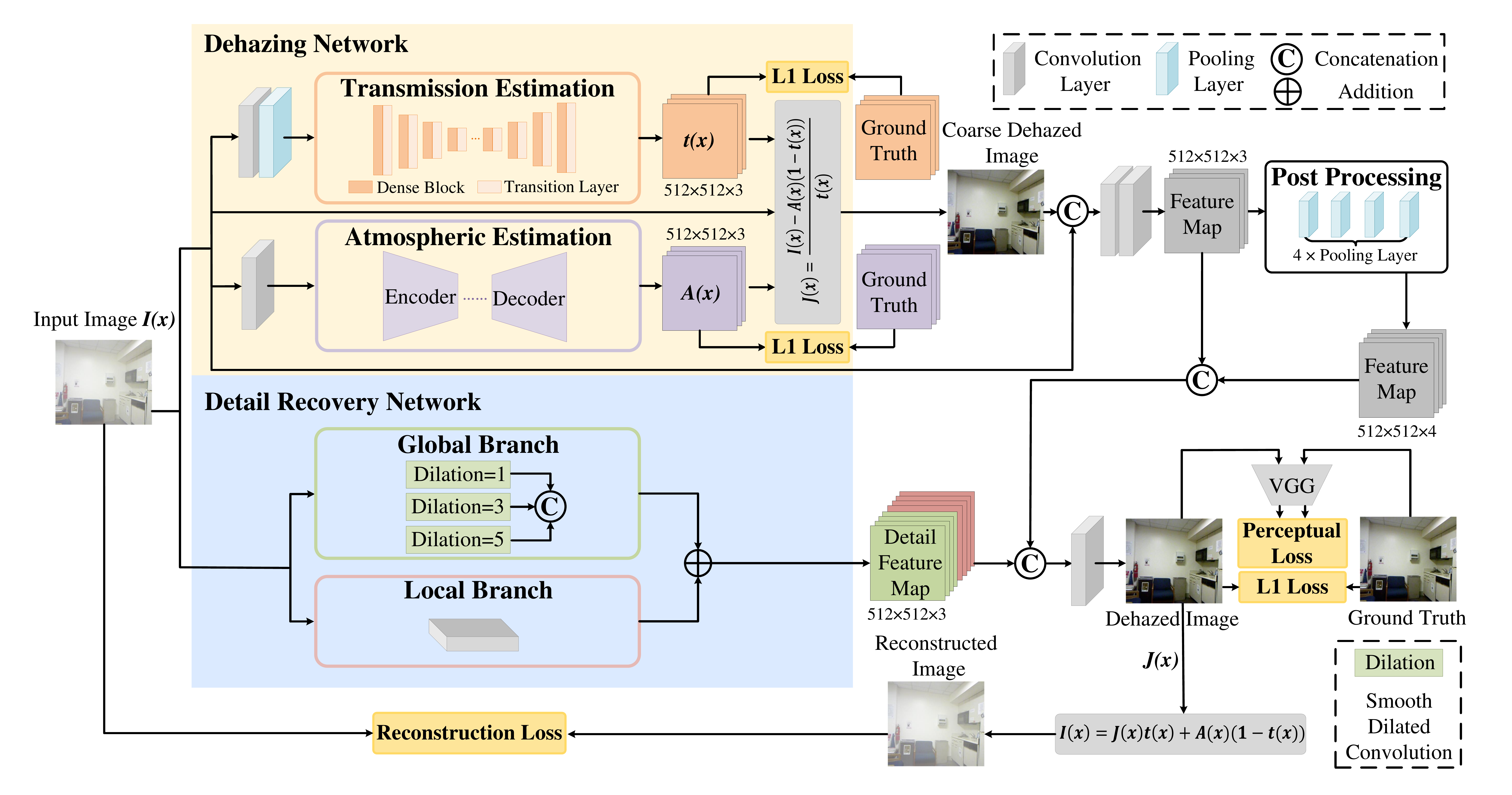}
	\caption{The overall architecture of our proposed network. The whole framework consists of two branch networks. The top one is independent DRN, which is used to recover the intrinsic image details contained in the hazy images. It consists of a local branch and a global branch to extract the local details and global image features. The specific structure of the network is described in Fig. \ref{detail_network}. The bottom one is the image dehazing network which achieves the estimation of the transmission map and the atmospheric light. Then we utilize them to estimate the coarse dehazed image via Eq. (\ref{eq:2}). A detailed description of the dehazing network is shown in Fig. \ref{dehazing_network}. The detail feature map is integrated into the coarse dehazed image to obtain the high-quality dehazed image finally. We formulate the DRN, the physical atmospheric scattering model and the reconstruction loss into an end-to-end joint learning framework.}
	\label{total-network}
\end{figure*}

\begin{figure}[t]
	\centering
	\includegraphics[width=9cm, height=6cm]{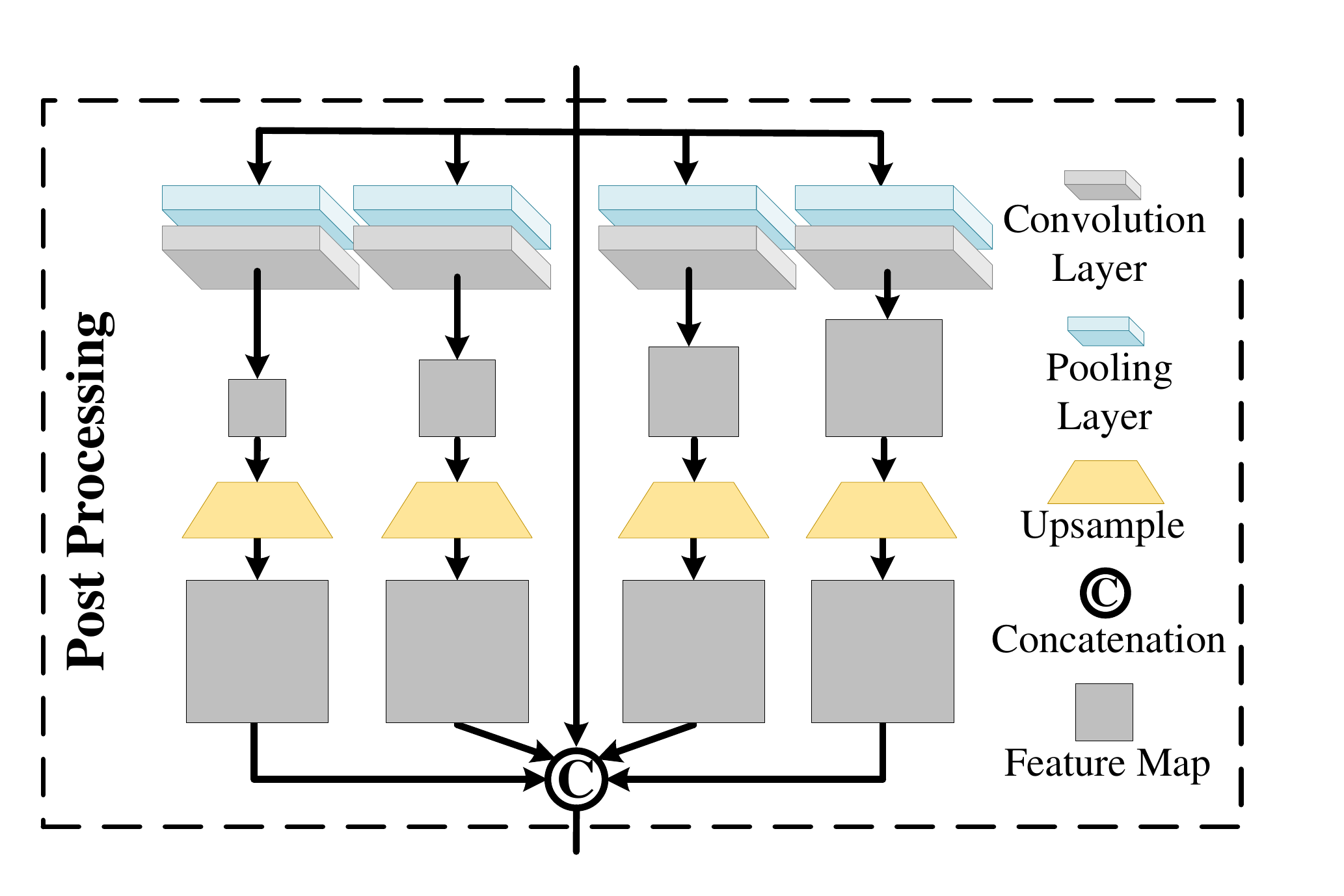}
	\caption{The post-processing module in Fig. \ref{total-network}. The feature maps of the coarse dehazed image are firstly input into four scale pooling layers to obtain feature maps with different levels. Then, the dimension of multi-level feature maps is adjusted by a 1$\times$1 convolution layer respectively. Finally, the multi-level feature maps are adjusted to the size of the input image by upsampling operation and are concatenated with the initial input as the final output of the module.}
	\label{total-part}
\end{figure}

\section{Method}
Most existing methods generally focus more on image dehazing but less on details recovery for the dehazed image. However, the detail information is of great significance for image reconstruction. Some methods attempt to preserve details by taking measures to the dehazing backbone network. However, detail loss is inevitable in the process of dehazing, so it is difficult to retain information in the dehazing process or restore it after dehazing. Therefore, we build an independent DRN to recover the intrinsic image details contained in the hazy images, which takes the hazy image as input and outputs the detail feature map. We fuse the detail feature map into the coarse dehazed image from the dehazing backbone network and finally output a high-quality dehazed image. Since the encoder-decoder structure has been proved to achieve good results in image reconstruction fields \cite{mao2016image, ren2018gated}, many methods \cite{zhang2018densely, li2020zero} adopt the encoder-decoder structure as the estimation network for transmission map. Therefore, we use the encoder-decoder structure with residual connections as the dehazing network. It takes a dense block as the basic unit for the sake of its capacity of maximizing the information flow and adds the shortcut connection on the dense blocks to transmit feature information of the shallow layers to the deep layers. In addition, many methods only focus on the estimation of transmission map but ignore the accuracy of atmospheric light, which is also critical to the performance of image dehazing. Instead, we choose a simple but effective U-Net \cite{ronneberger2015u} to learn atmospheric light. Because there is less high-level semantic information in atmospheric light, low-level features are particularly important. U-Net can integrate shallow features into deep layers so that the final output image can make better use of low-level features and high-level semantic information.

\subsection{Overall Framework}
Fig. \ref{haze-dehaze} shows the overall architecture of the proposed network. The overall network consists of DRN and the physical-model-based dehazing network. On one hand, the dehazing network takes a hazy image as input and outputs a coarse dehazed image. Specifically, it consists of two parallel branches i.e. transmission map estimation network and atmospheric light estimation network, which are used to learn the estimated transmission map and the estimated atmospheric light, respectively. For both branch networks, we utilize L1 loss to constrain the training process. Further, the coarse dehazed image is obtained by calculating the output results of both branches and the original input hazy image by Eq. \ref{eq:2}. Then, the coarse dehazed image is enhanced by a post-processing module to integrate multi-scale feature information. On the other hand, DRN takes a hazy image as input and outputs a detail feature map. It also has two branches, global branch and local branch, which are used to obtain the global features and local details of the image respectively. Finally, the enhanced coarse dehazed image is concatenated with the detail feature map to obtain the final detailed dehazed image. For the final output dehazed image, we employ two kinds of loss constraints i.e. perceptual loss and L1 loss. Perceptual loss can calculate the loss of image feature level and L1 loss can calculate the loss of image pixel level. In order to make the training of the whole network more stable, we reconstruct the hazy image from the final dehazed image through Eq. \ref{eq:1} and use the loss between the reconstructed hazy image and the input hazy image to further constrain the training of the network. We introduce the DRN and the dehazing network in more details in the following sections.

\begin{figure}[]
	\centering
	\includegraphics[width=\linewidth]{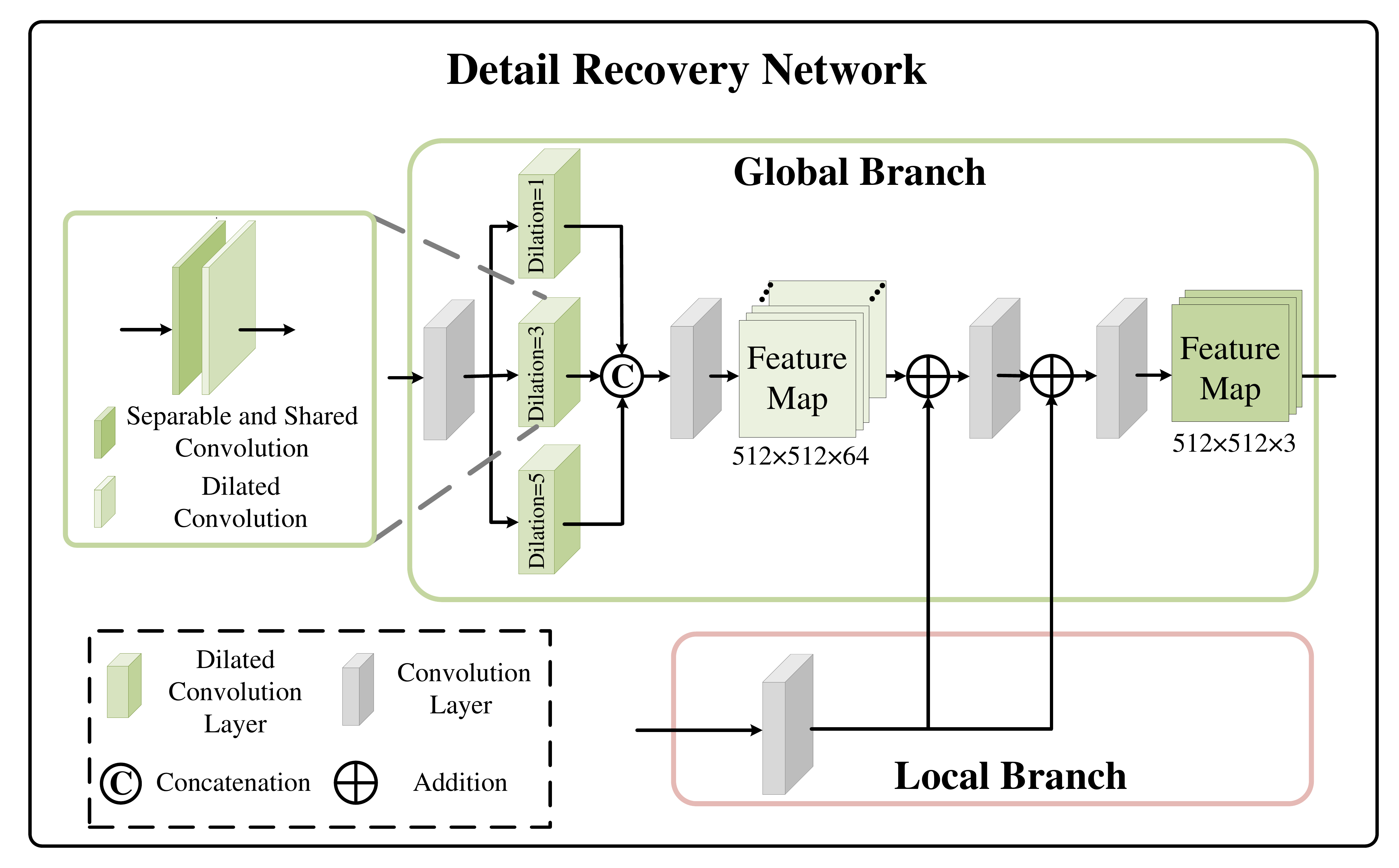}
	\caption{An overview of DRN. The independent DRN aims to recover the details of the dehazed image, which consists of the local and global branches to extract the local details and global image features. The local branch can obtain more detail information through the convolution layer and the global branch can capture more global information by SDC with larger reception fields. Local feature information is repeatedly incorporated into the global branch for more details. The final output detail feature will be integrated into the coarse dehazed image to obtain a high-quality dehazed image.}
	\label{detail_network}
\end{figure}

\subsection{Detail-Recovery Network}
During the image dehazing process, details are inevitably lost and the dehazed images with missing details are not desirable. Most existing approaches pay more attention to reconstruct the haze-free images but overlook the recovery of dehazed image details. A few approaches consider the detail-preserving measures in the dehazing backbone network. However, the input hazy image will inevitably be accompanied by the loss of details, so it is difficult to retain information in the dehazing process or to recover it after dehazing. Instead, we build an independent DRN to recover the intrinsic image details contained in the hazy images, which takes the hazy image as input and outputs the detail feature map.

Inspired by the work in \cite{deng2020detail}, we design DRN to recover the details of the reconstructed image. However, \cite{deng2020detail} only considers obtaining more global features through dilated convolution in the detail recovery process. Although global features play an important role in image reconstruction, which can obtain more global information of a whole image, local features are more advantageous for the local texture and the color of the image, which is crucial for the recovery of image details as well. 

Therefore, we propose a dual-branch DRN containing global branch and local branch, where the former is used to extract the global features while the latter aims at the local detail features of the image. As shown in Fig. \ref{detail_network}, DRN fed the input hazy image into the global branch and the local branch, respectively. 

As is well-known, the dilated convolution can expand the receptive field without losing resolution or damaging the structures. Therefore, in the global branch, the input feature maps first pass through a 1$\times$1 convolution layer, and then follow
three parallel dilated convolutions with different dilated factors. The feature maps with different receptive fields can be obtained and then they are concatenated together. A 1$\times$1 convolution layer \cite{lin2013network} is followed to adjust the number of channels, which can realize cross-channel interaction and information integration at the same time. Further, we concatenate the input feature maps with the output feature maps of the above 1$\times$1 convolution layer by considering the original input image information. In order to get more local details, we build local branch. Since there will be information loss with the deepening of the network, and considering that deep convolution often has a larger receptive field which cannot pay attention to details, we only use one convolution layer as our local branch. Local feature information is incorporated into the global branch twice through the addition operation, so that the local details can be better supplemented. Each addition operation is followed by a convolution layer. The final output feature maps are obtained by sending the concatenated feature maps into the two convolution layers for further channel adjustment and information fusion.

Most important of all, since general dilated convolution produces the gridding artifacts \cite{wang2018smoothed}, we replace them with smooth dilated convolution, which consists of a separable and shared convolution and a dilated convolution. The ablation experiments later validate that these modifications are beneficial to improve the performance of our detail-recovery network.   
The final detail feature maps contains both global feature information and local feature information of the original input image.

\begin{figure}[]
	\centering
	\includegraphics[width=\linewidth]{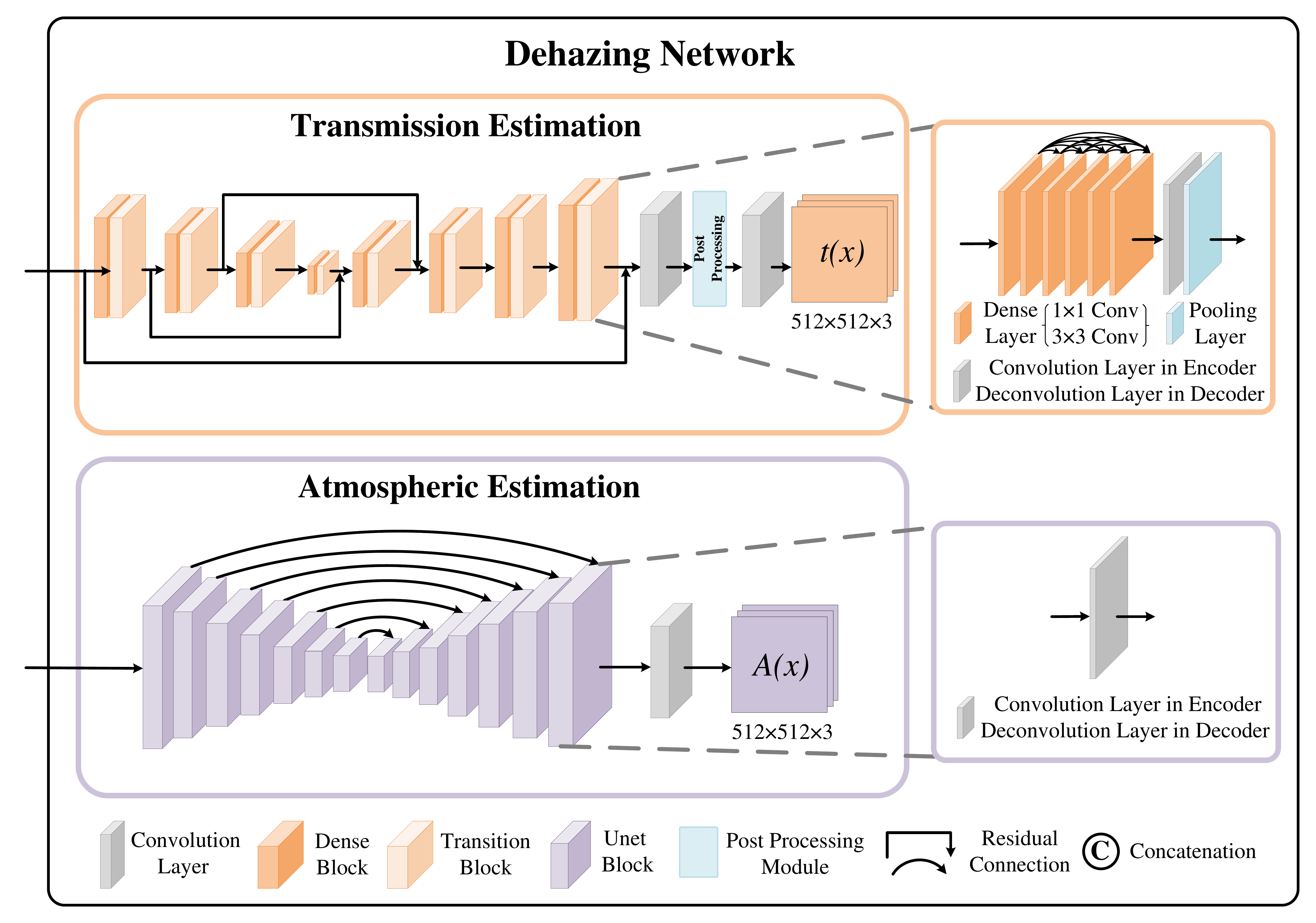}
	\caption{An overview of the physical-model-based dehazing network. The network consists of two subnetworks, which are the transmission map estimated network and the atmospheric light estimated network, respectively.}
	\label{dehazing_network}
\end{figure}

\subsection{Dehazing Network}
As shown in Fig. \ref{dehazing_network}, the dehazing network is mainly composed of two parallel branches, transmission map estimation network and atmospheric light estimation network. The dehazing network inputs a hazy image into two parallel branches. The upper branch outputs the estimated transmission map and the lower branch outputs the estimated atmospheric light. Combining the generated transmission map, atmospheric light and the input image, the coarse dehazed image can be generated by Eq. \ref{eq:2}.

Since the encoder-decoder structure has been proved to achieve good results in image reconstruction fields \cite{mao2016image, ren2018gated}, many methods \cite{zhang2018densely, li2020zero} adopt the encoder-decoder structure as the estimation network for transmission map. Similarly, we construct a transmission map estimation network with an encoder-decoder structure. In order to obtain more feature information from the decoder, many methods supplement the information from the encoder to the decoder through various connections. Our dehazing network takes a dense block as the basic unit to maximize information flow, and then the shortcut connection is established on the dense block to realize the transmission of feature information from shallow layers to deep layers. Every dense block has six dense layers with a residual connection between each layer. After the dense block, there is a transition block, which consists of a convolution layer and a pooling layer. The decoder also takes the dense block as the basic unit. As shown in Fig. \ref{total-network}, a hazy image is first transformed into the feature maps through a convolution layer and a pooling layer, and then these feature maps are sent into the transmission estimation network. To make up the loss of information, we establish the residual connections between the encoder and the decoder, which enables the decoder to obtain more feature information of the original image. After the feature maps output by the decoder, they pass through a convolution layer to adjust the number of channels. In order to obtain multi-scale features, a post processing module is added after the convolution layer, which will be described in detail below. Finally, the transmission map is output through a convolution layer.

Since there is little semantic information in the atmospheric light, more emphasis is placed on low-level features. However, the deepening of the network will be accompanied by the loss of low-level features, so we choose the simple but effective U-Net \cite{ronneberger2015u} to learn the atmospheric light. U-Net can transmit low-level features in the shallow layers to the deep layers, so as to estimate the atmospheric light more accurately.

\subsection{Post-processing Network}
As presented in \cite{ren2016single, zhang2018densely}, the multi-scale features for estimating the transmission map are rewarding. Therefore, we input the coarse image into a post-processing network for further enhancement. The post-processing network structure is shown in Fig. \ref{total-part}. The network is composed of four parallel branches and each of which is composed of a pooling layer and a convolution layer. The downsampling factors of the four pooling layers are 32, 16, 8, and 4, respectively. Therefore, four-scale feature maps with different receptive fields are generated, which can provide multi-scale features and are more conducive to the next generation of clearer images. Finally, the feature maps of four different scales are upsampled to the same scale, which is the same as the input image scale, and then concatenated with the initial input as the output.

\subsection{Loss Functions}
\textbf{Perceptual Loss: }
Since the perceptual loss is proposed by \cite{johnson2016perceptual}, it has been widely applied in image reconstruction. The perceptual loss calculates the visual difference between the dehazed image and the ground truth quantificationally by measuring the gap of their high-level feature representations extracted from a pre-trained deep neural network. In our method, we leverage VGG16 \cite{simonyan2014very} pre-trained on ImageNet dataset \cite{ILSVRC15} as the loss network and extract the features from different layers to calculate the loss of different levels semantics. We introduce a perceptual loss for the sake of the improvement in the perceptual similarity between the output dehazed image $I_{pred}$ and the ground truth $\hat{I}$, which is defined as:
\begin{equation}
L_{per} = \sum_{j=3,8,15} \sum_{i=1}^{N} ||\phi_{j}(I_{pred}) - \phi_{j}(\hat{I})||, \label{eq:3}
\end{equation}
where $\phi_{j}(\cdot)$ denotes the feature maps from the $j$-th layer of the VGG16, $N$ represents the batch size and $||\cdot||$ is the L1 norm. 

\textbf{Pixel-wise Loss: }
Except for the perceptual loss on feature level, we introduce the pixel-wise loss particularly, which can compute the pixel-level difference between the dehazed image and ground truth. B. Lim et al \cite{lim2017enhanced} point out that L1 loss achieved superior performance for many images reconstruction in terms of Peak Signal-to-Noise Ratio (PSNR) \cite{huynh2008scope} and Structural Similarity (SSIM) \cite{wang2004image}. We implement the pixel-wise loss by the L1 loss function. We utilize the pixel-wise loss to measure the residual of the transmissions map, the atmospheric light and dehazed image estimated by our method. 
The loss for dehazed image is defined as Eq. (\ref{eq:4}).
\begin{equation}
L_{d} = \sum_{i=1}^{N}||I_{pred} - \hat{I}||, \label{eq:4}  
\end{equation}
where $N$ represents the batch size and $||\cdot||$ is the L1 norm.

Specifically, the loss of transmission map is defined as Eq. (\ref{eq:5}), where $t_{pred}$ is the estimated transmission map and $\hat{t}$ is the ground truth transmission map.
\begin{equation}
L_{t} = \sum_{i=1}^{N}||t_{pred} - \hat{t}||. \label{eq:5}  
\end{equation}

The loss of the atmospheric light is defined as Eq. (\ref{eq:6}), where $A_{pred}$ is the estimated atmospheric light and $\hat{A}$ is the ground truth atmospheric light.
\begin{equation}
L_{a} = \sum_{i=1}^{N}||A_{pred} - \hat{A}||. \label{eq:6}  
\end{equation}

\textbf{Reconstruction Loss: }
The reconstruction loss is employed to make the whole network be jointly optimized. We utilize the estimated transmission map $t_{pred}$, atmospheric light $A_{pred}$ and dehazed image $I_{pred}$ to reconstruct the image $I_{rec}$ on the basis of Eq. (\ref{eq:1}), and then the reconstruction loss is built up between the reconstruct image $I_{rec}$ and the input image $I$. At the same time, reconstruction loss can make use of the potential correlations between transmission map, atmospheric light and dehazed image during the training, so as to improve the network performance. The reconstruction loss is defined as Eq.(7): 
\begin{equation}
L_{rec} = \sum_{i=1}^{N}||I_{rec} - I||. \label{eq:7}  
\end{equation}

\textbf{Total Loss: }
Ultimately, the total loss of our method is a multi-faceted loss function which can think about the perceptual similarity, the pixel-level similarity and the reconstruction gap. The total loss is defined by combing the five loss functions mentioned above.
\begin{equation}
L_{total} = L_{per} + L_{d} + L_{t} + L_{a} + L_{rec}. \label{eq:8}
\end{equation}

\begin{table}[]
	\scriptsize
	\centering
	\caption{Quantitative comparisons on RESIDE for different methods.}
	\vspace{10pt}
	\setlength\tabcolsep{8pt}
	\scalebox{1.1}[1.1]{
		\begin{tabular}{lcccc}
			\toprule
			\multicolumn{1}{c}{\multirow{3}{*}{Method}} & \multicolumn{2}{c}{RESIDE-Indoor} & \multicolumn{2}{c}{RESIDE-Outdoor} \\ \cmidrule(l){2-5} 
			& PSNR           & SSIM           & PSNR            & SSIM           \\ 
			\midrule
			DCP \cite{he2010single}        & 16.62 & 0.82  & 19.13 & 0.81  \\
			DehazeNet \cite{cai2016dehazenet}   & 21.14 & 0.85  & 22.46 & 0.85  \\
			AOD-Net \cite{li2017aod}    & 19.06 & 0.85  & 20.29 & 0.88 \\
			DCPDN \cite{zhang2018densely}       & 19.00 & 0.84  & 19.71 & 0.83 \\
			GFN \cite{ren2018gated}       & 22.30 & 0.88  & 21.55 & 0.84 \\
			PFFNet \cite{mei2018progressive}      & 26.58 & 0.92  & \textbf{27.65} & \underline{0.91} \\
			EPDN \cite{qu2019enhanced}       & 25.06 & 0.92  & 22.57 & 0.86 \\ 
			FDU \cite{dong2020physics}         & \underline{32.68} & \underline{0.98}  & --- & --- \\
			\midrule
			\multicolumn{1}{c}{\textbf{Ours}}       & \textbf{33.01} & \textbf{0.98} &\underline{24.44} & \textbf{0.94}   \\
			\bottomrule 
	\end{tabular}}
	\label{SOTS-table}
\end{table}

\begin{figure*}[h]
\centering
	\footnotesize
	\begin{minipage}{0.09\linewidth}
		\vspace{3pt}
		\centerline{\includegraphics[width=\textwidth]{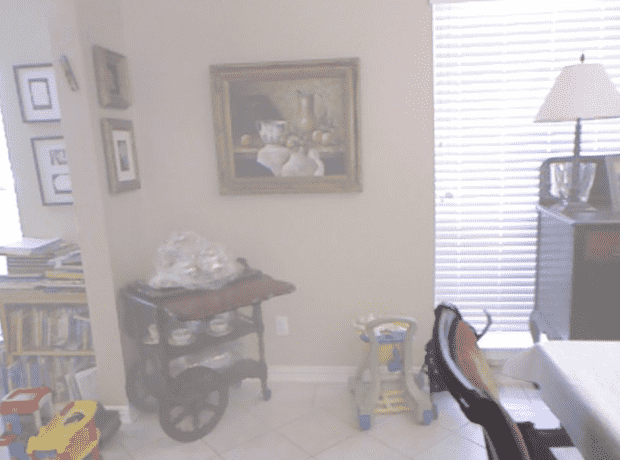}}
		\vspace{3pt}
		\centerline{\includegraphics[width=\textwidth]{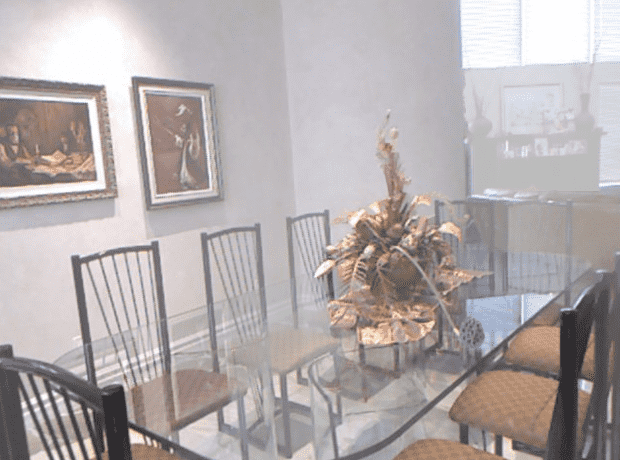}}
		\vspace{3pt}
		\centerline{\includegraphics[width=\textwidth]{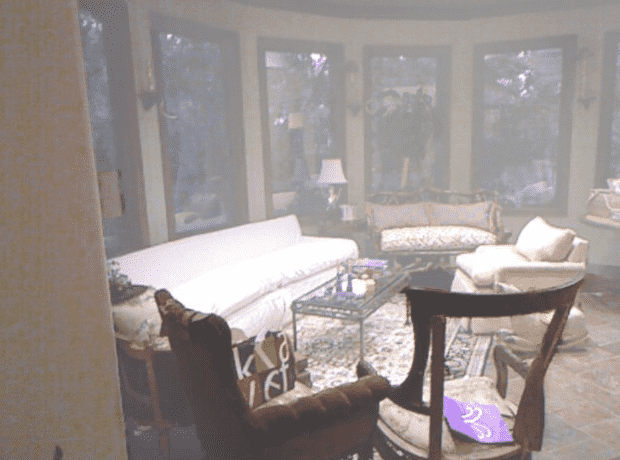}}
		\vspace{3pt}
		\centerline{\includegraphics[width=\textwidth]{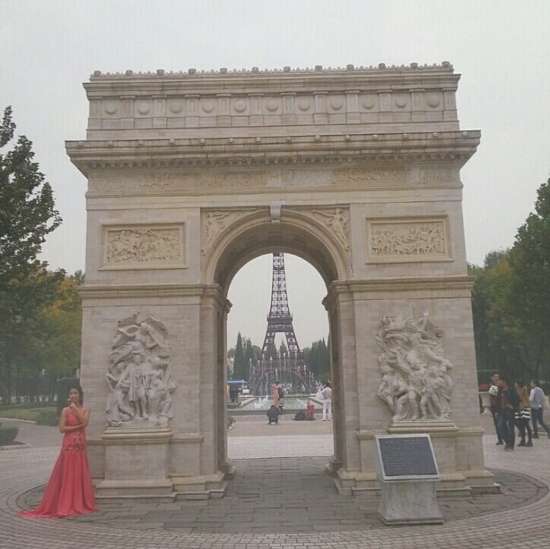}}
		\vspace{3pt}
		\centerline{\includegraphics[width=\textwidth]{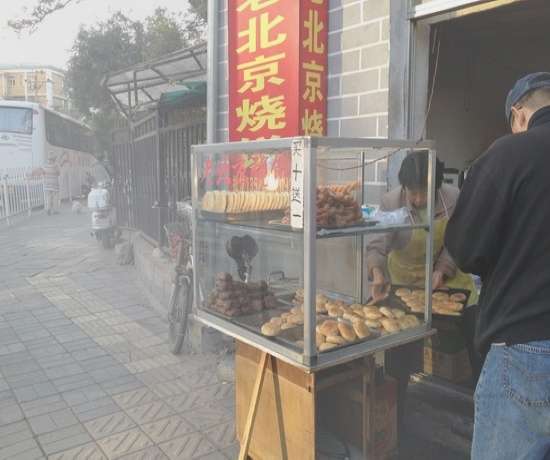}}
		\vspace{3pt}
		\centerline{\includegraphics[width=\textwidth]{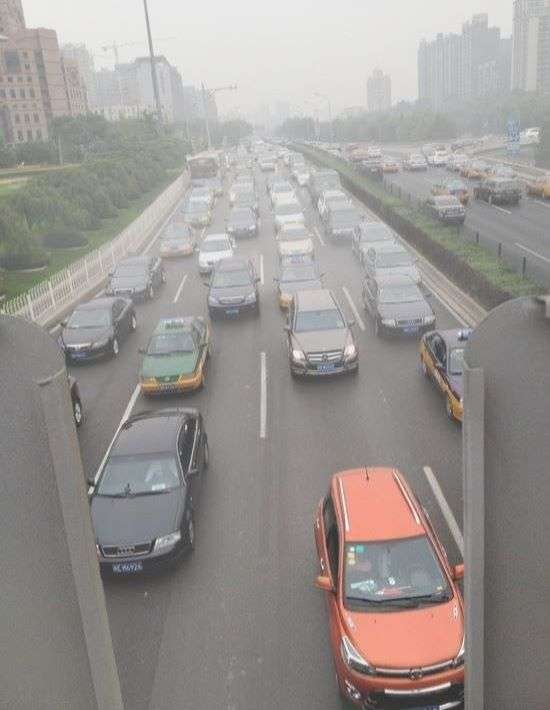}}
		\vspace{3pt}
		\centerline{(a)}
	\end{minipage}
	\begin{minipage}{0.09\linewidth}
		\vspace{3pt}
		\centerline{\includegraphics[width=\textwidth]{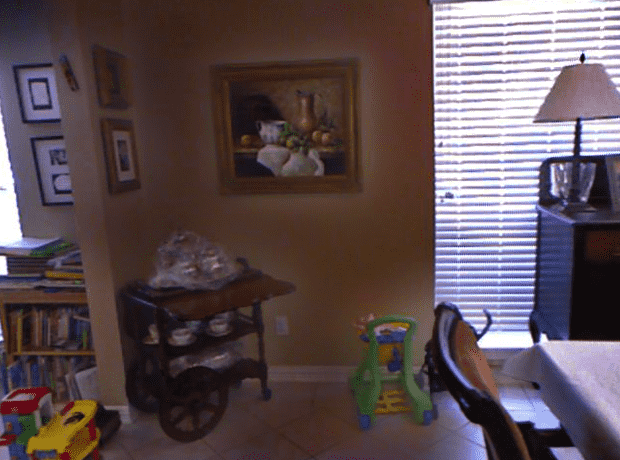}}
		\vspace{3pt}
		\centerline{\includegraphics[width=\textwidth]{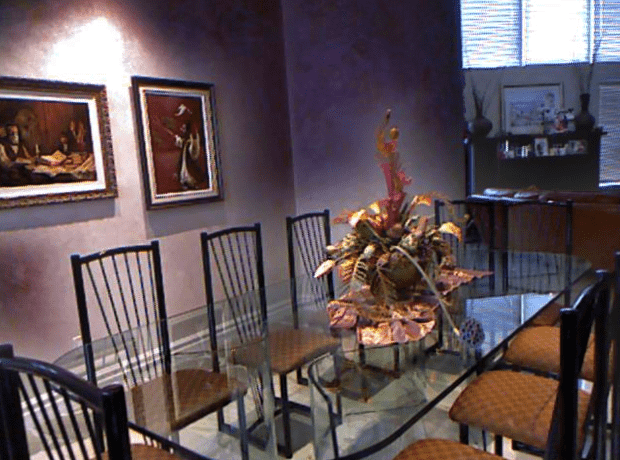}}
		\vspace{3pt}
		\centerline{\includegraphics[width=\textwidth]{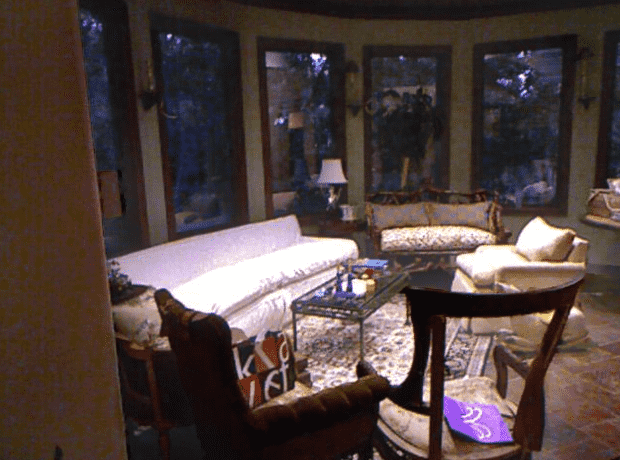}}
		\vspace{3pt}
		\centerline{\includegraphics[width=\textwidth]{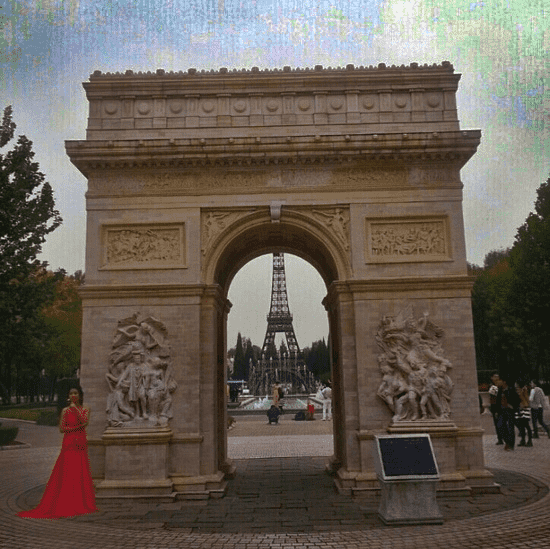}}
		\vspace{3pt}
		\centerline{\includegraphics[width=\textwidth]{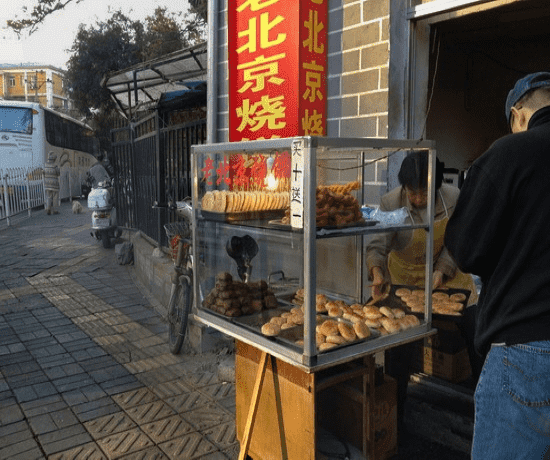}}
		\vspace{3pt}
		\centerline{\includegraphics[width=\textwidth]{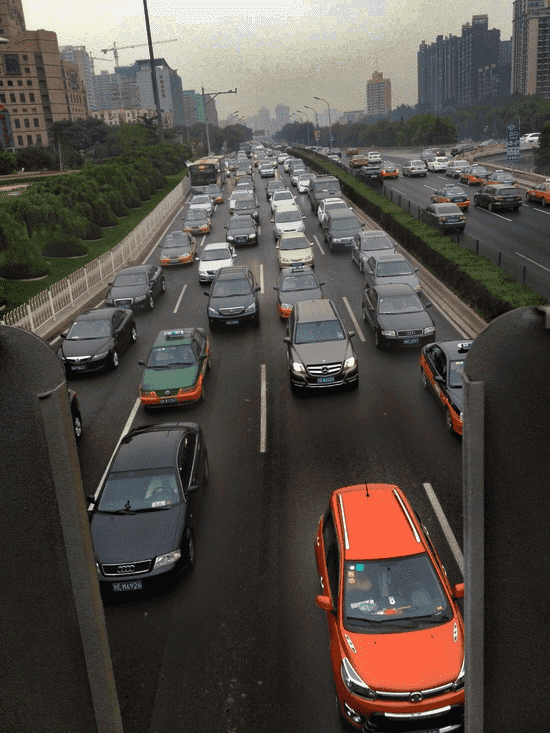}}
		\vspace{3pt}
		\centerline{(b)}
	\end{minipage}
	\begin{minipage}{0.09\linewidth}
		\vspace{3pt}
		\centerline{\includegraphics[width=\textwidth]{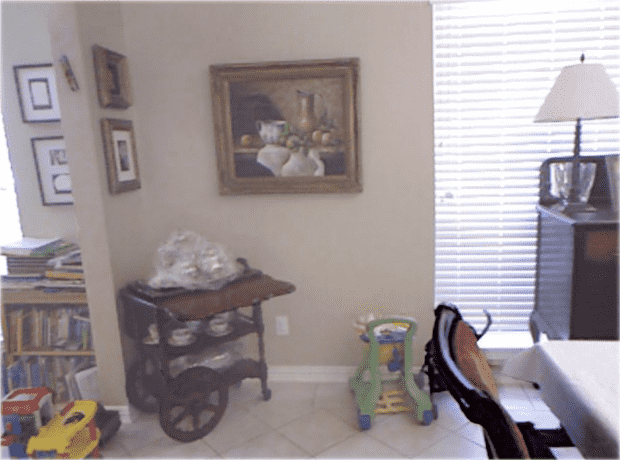}}
		\vspace{3pt}
		\centerline{\includegraphics[width=\textwidth]{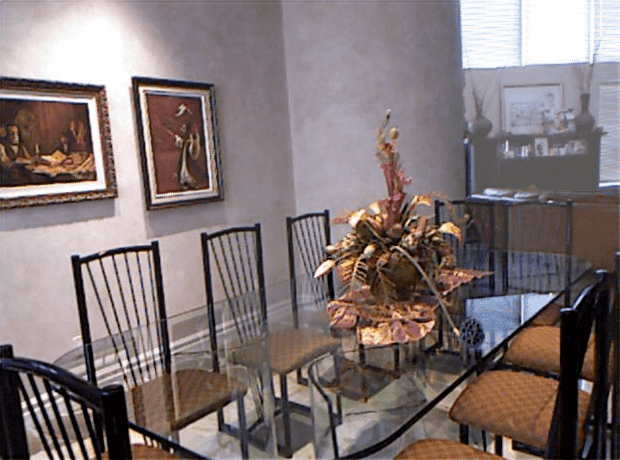}}
		\vspace{3pt}
		\centerline{\includegraphics[width=\textwidth]{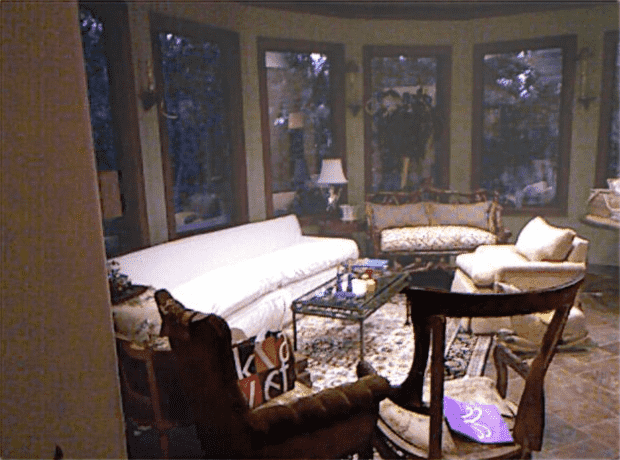}}
		\vspace{3pt}
		\centerline{\includegraphics[width=\textwidth]{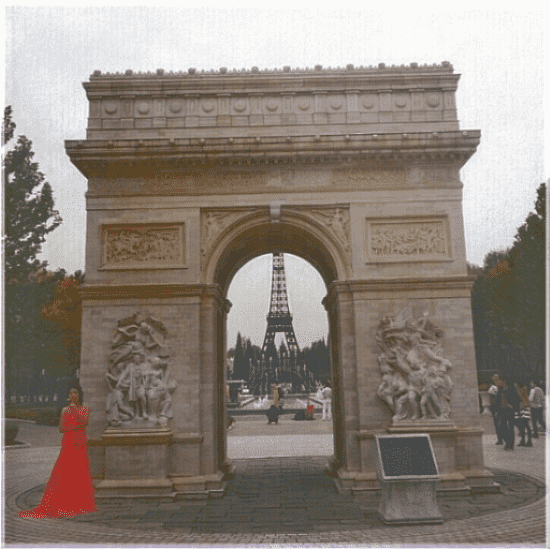}}
		\vspace{3pt}
		\centerline{\includegraphics[width=\textwidth]{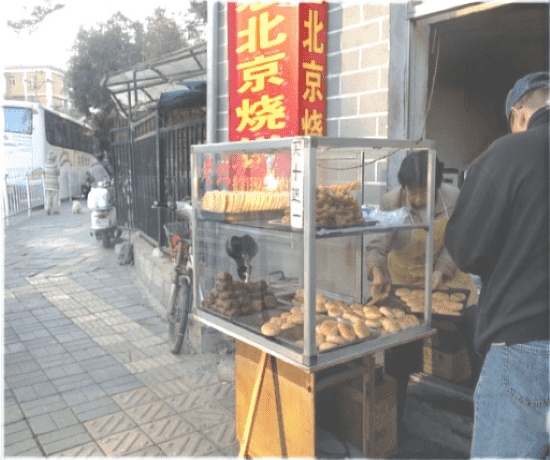}}
		\vspace{3pt}
		\centerline{\includegraphics[width=\textwidth]{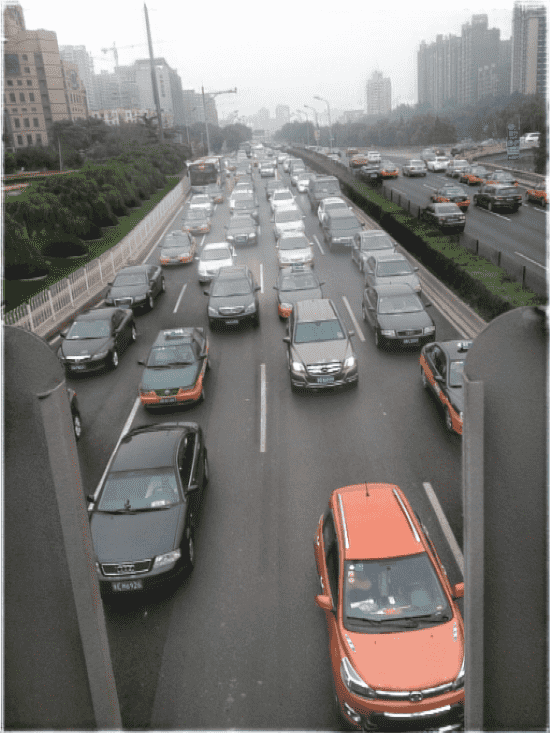}}
		\vspace{3pt}
		\centerline{(c)}
	\end{minipage}
	\begin{minipage}{0.09\linewidth}
		\vspace{3pt}
		\centerline{\includegraphics[width=\textwidth]{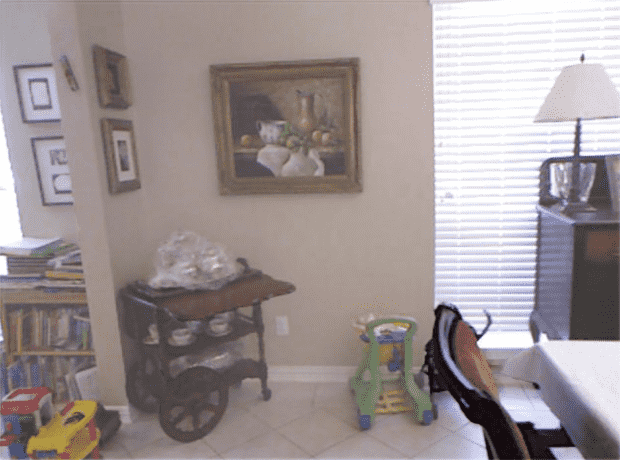}}
		\vspace{3pt}
		\centerline{\includegraphics[width=\textwidth]{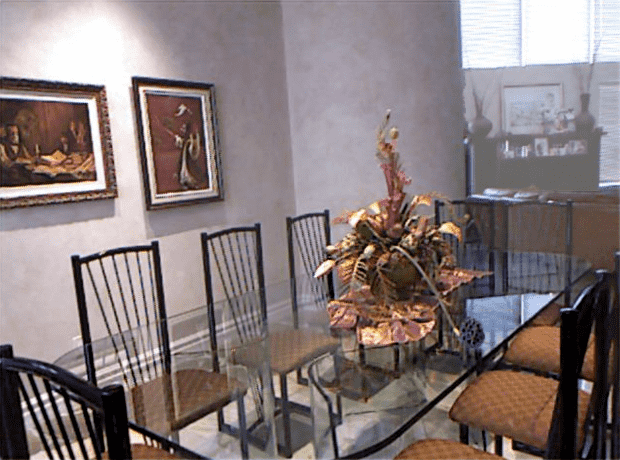}}
		\vspace{3pt}
		\centerline{\includegraphics[width=\textwidth]{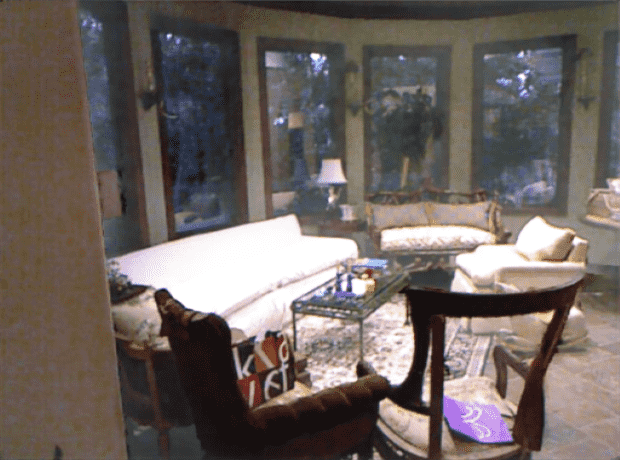}}
		\vspace{3pt}
		\centerline{\includegraphics[width=\textwidth]{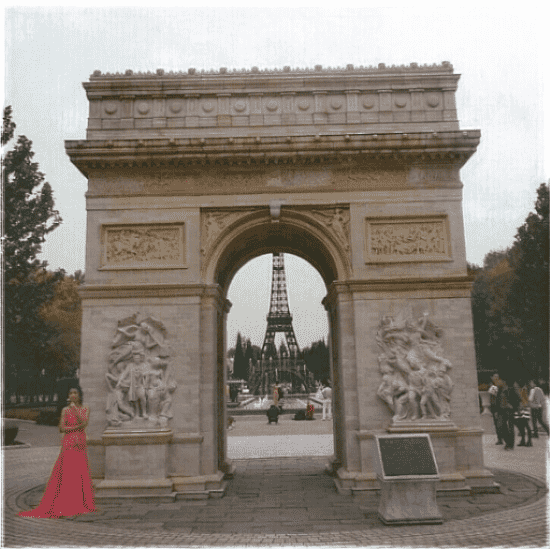}}
		\vspace{3pt}
		\centerline{\includegraphics[width=\textwidth]{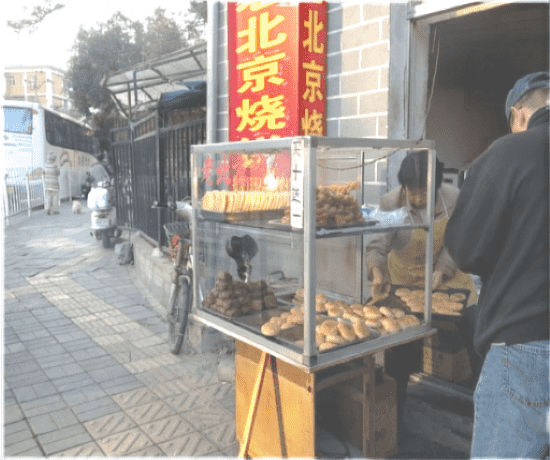}}
		\vspace{3pt}
		\centerline{\includegraphics[width=\textwidth]{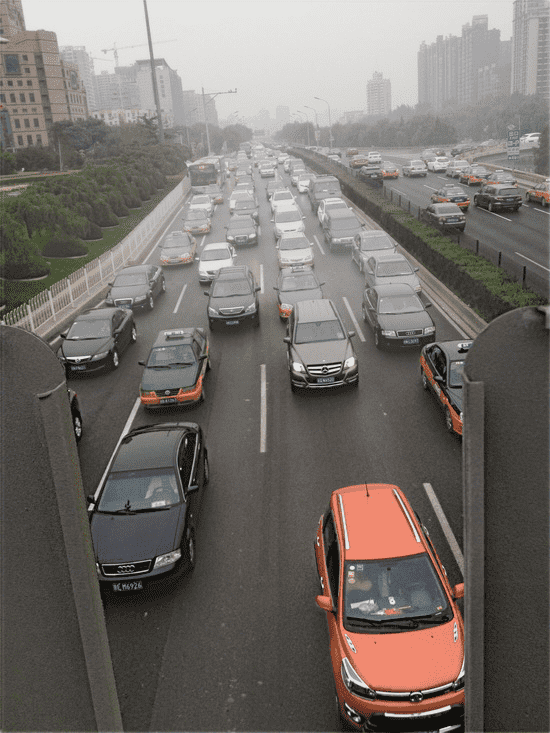}}
		\vspace{3pt}
		\centerline{(d)}
	\end{minipage}
	\begin{minipage}{0.09\linewidth}
		\vspace{3pt}
		\centerline{\includegraphics[width=\textwidth]{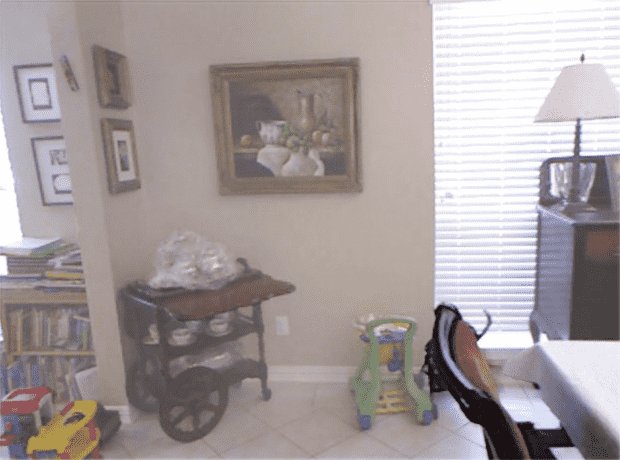}}
		\vspace{3pt}
		\centerline{\includegraphics[width=\textwidth]{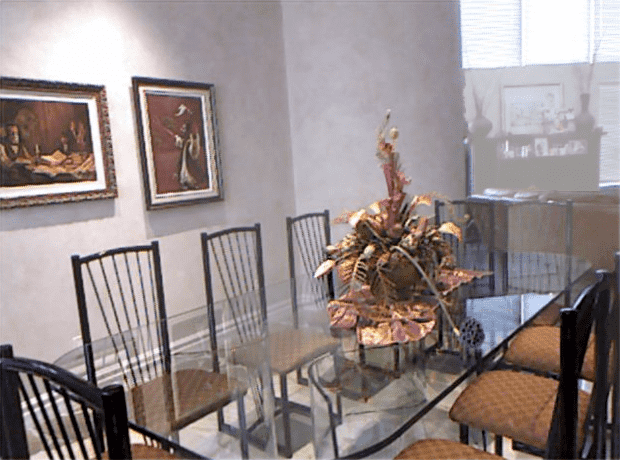}}
		\vspace{3pt}
		\centerline{\includegraphics[width=\textwidth]{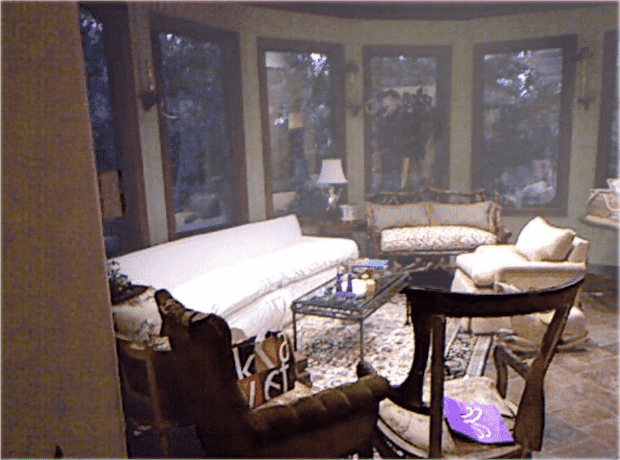}}
		\vspace{3pt}
		\centerline{\includegraphics[width=\textwidth]{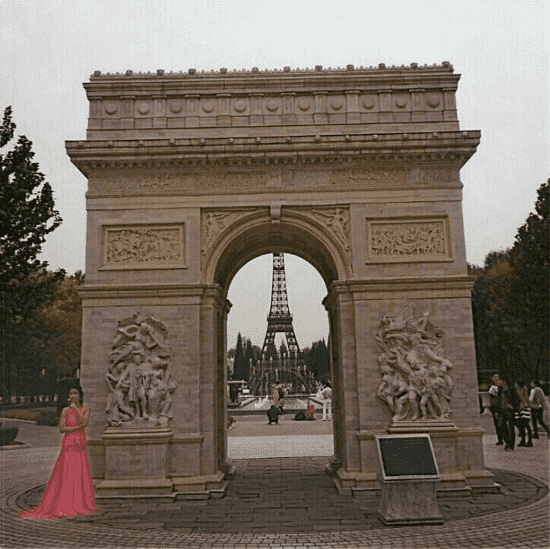}}
		\vspace{3pt}
		\centerline{\includegraphics[width=\textwidth]{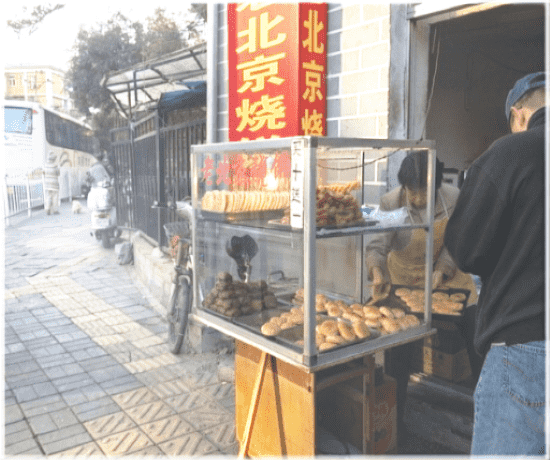}}
		\vspace{3pt}
		\centerline{\includegraphics[width=\textwidth]{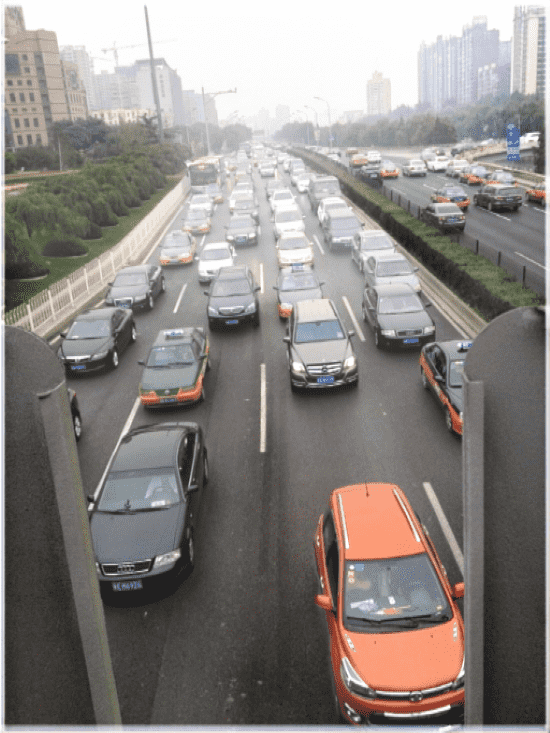}}
		\vspace{3pt}
		\centerline{(e)}
	\end{minipage}
	\begin{minipage}{0.09\linewidth}
		\vspace{3pt}
		\centerline{\includegraphics[width=\textwidth]{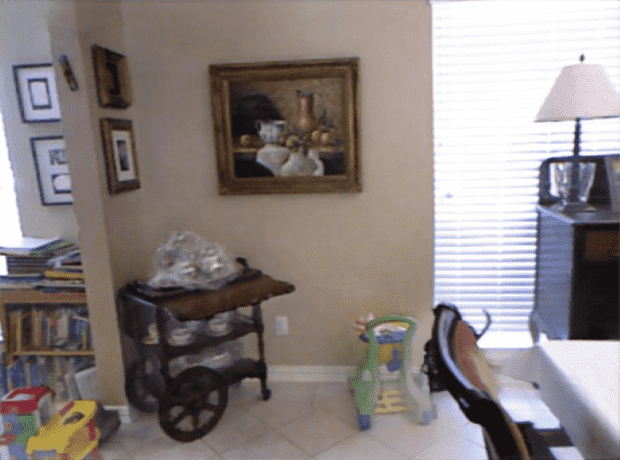}}
		\vspace{3pt}
		\centerline{\includegraphics[width=\textwidth]{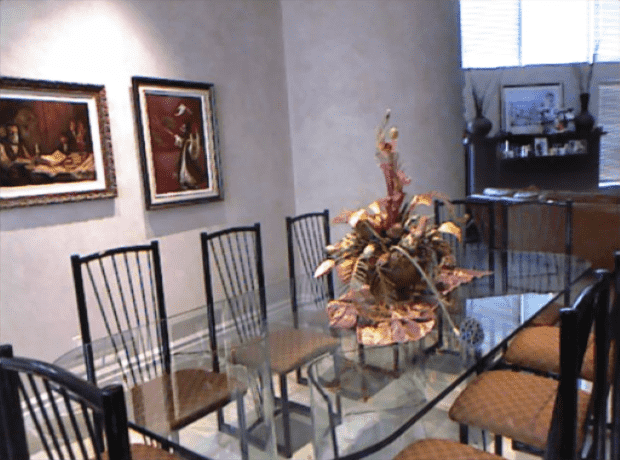}}
		\vspace{3pt}
		\centerline{\includegraphics[width=\textwidth]{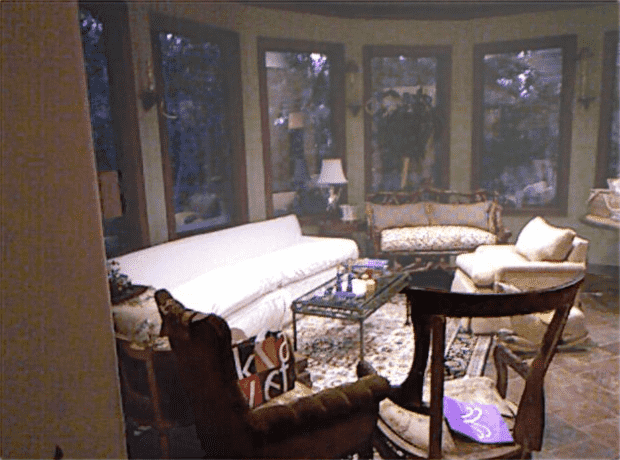}}
		\vspace{3pt}
		\centerline{\includegraphics[width=\textwidth]{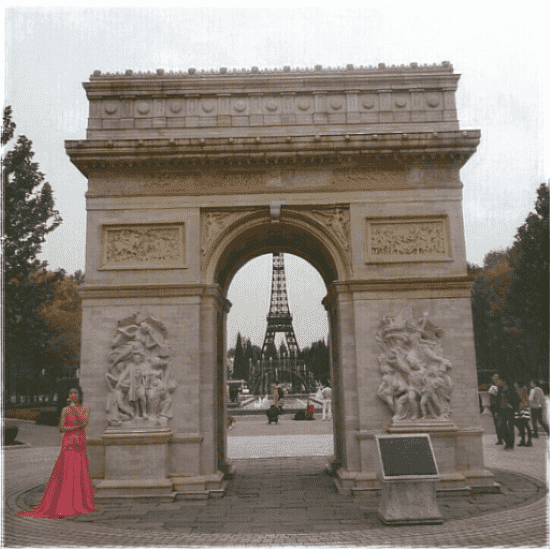}}
		\vspace{3pt}
		\centerline{\includegraphics[width=\textwidth]{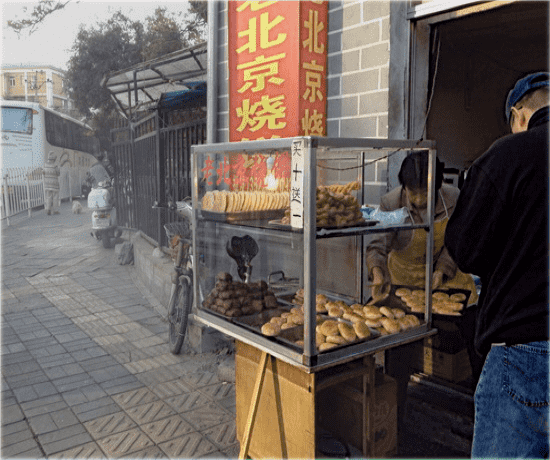}}
		\vspace{3pt}
		\centerline{\includegraphics[width=\textwidth]{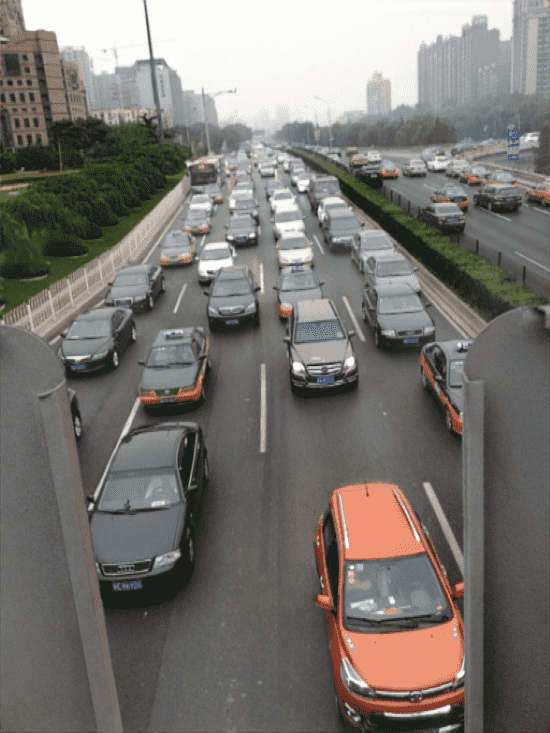}}
		\vspace{3pt}
		\centerline{(f)}
	\end{minipage}
	\begin{minipage}{0.09\linewidth}
		\vspace{3pt}
		\centerline{\includegraphics[width=\textwidth]{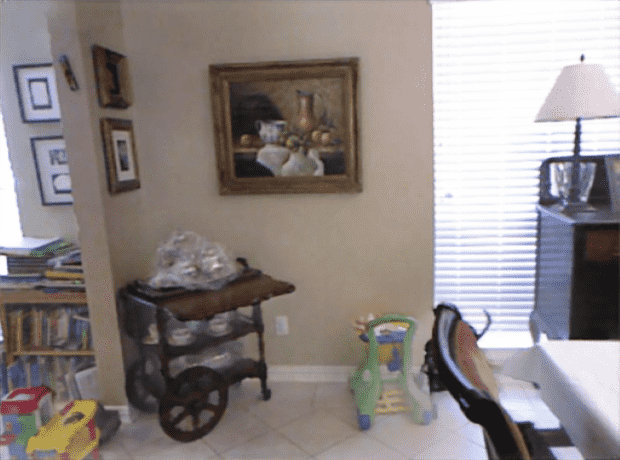}}
		\vspace{3pt}
		\centerline{\includegraphics[width=\textwidth]{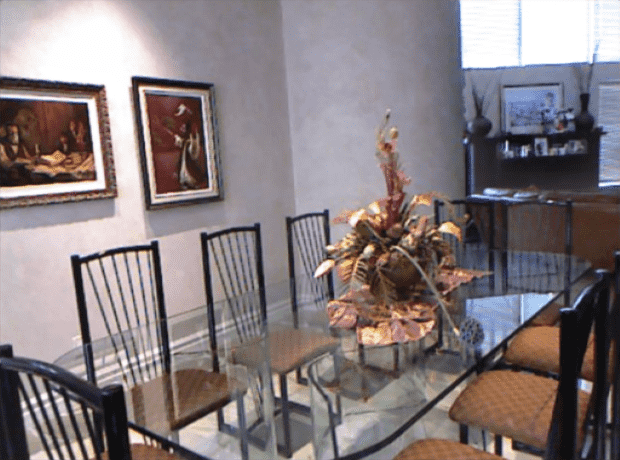}}
		\vspace{3pt}
		\centerline{\includegraphics[width=\textwidth]{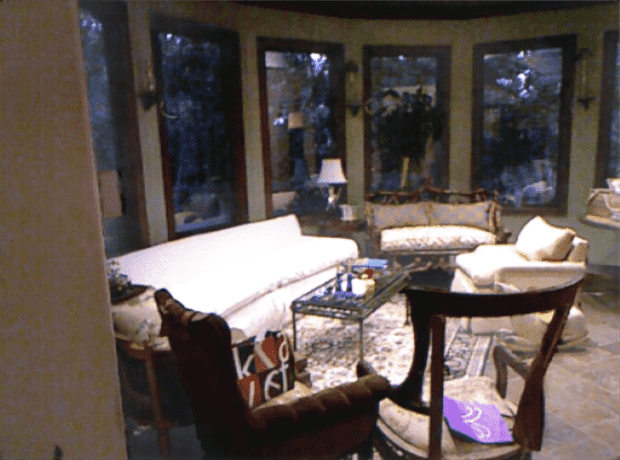}}
		\vspace{3pt}
		\centerline{\includegraphics[width=\textwidth]{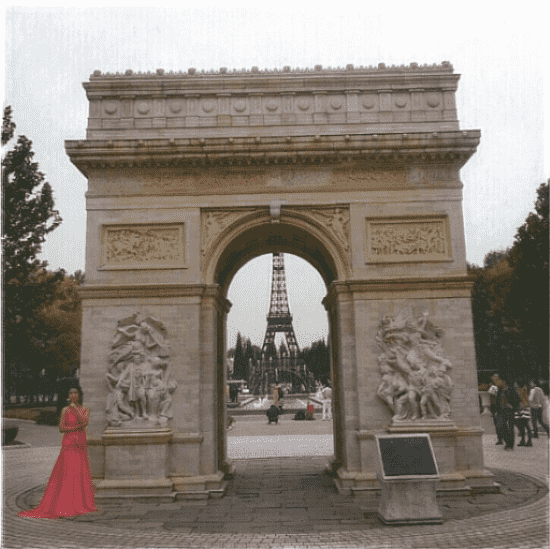}}
		\vspace{3pt}
		\centerline{\includegraphics[width=\textwidth]{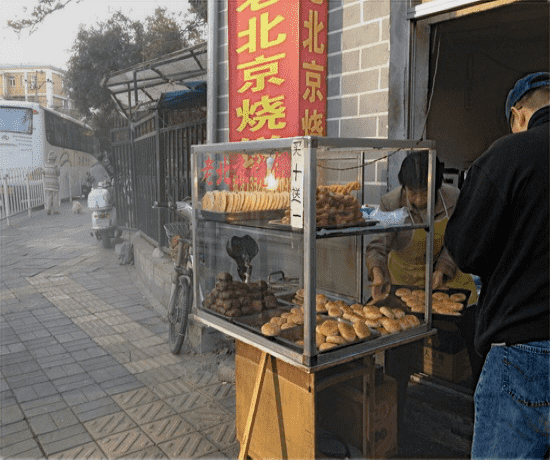}}
		\vspace{3pt}
		\centerline{\includegraphics[width=\textwidth]{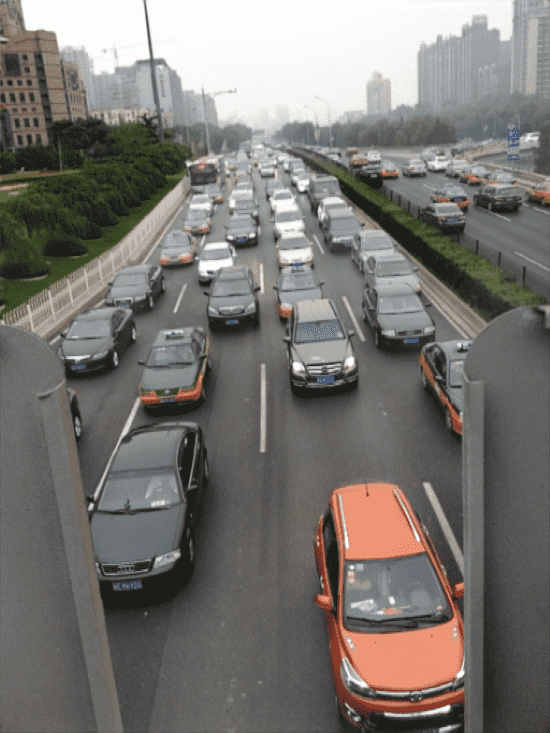}}
		\vspace{3pt}
		\centerline{(g)}
	\end{minipage}
	\begin{minipage}{0.09\linewidth}
		\vspace{3pt}
		\centerline{\includegraphics[width=\textwidth]{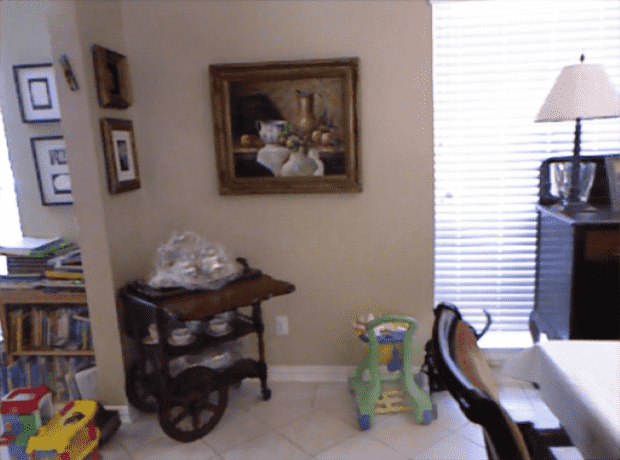}}
		\vspace{3pt}
		\centerline{\includegraphics[width=\textwidth]{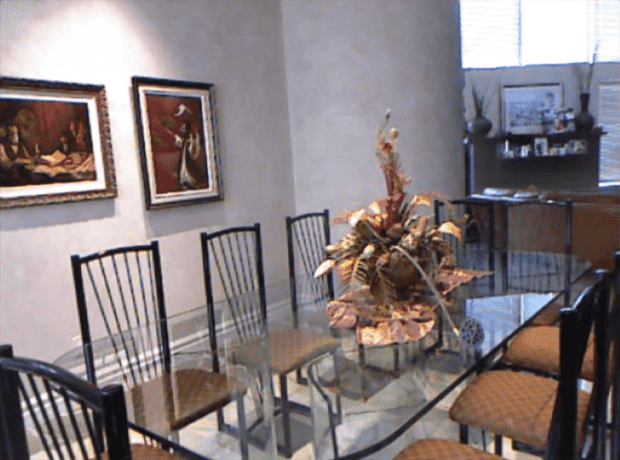}}
		\vspace{3pt}
		\centerline{\includegraphics[width=\textwidth]{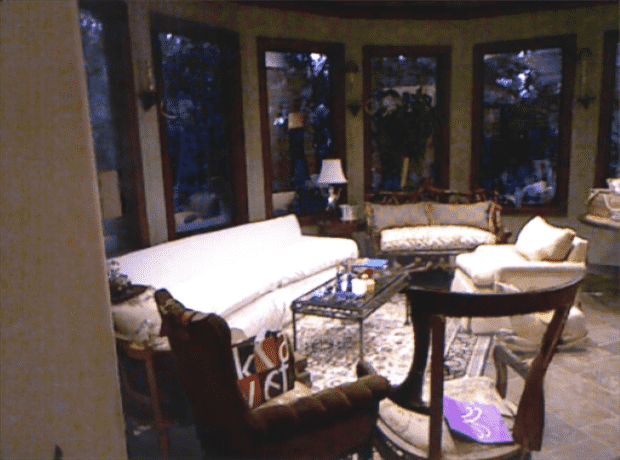}}
		\vspace{3pt}
		\centerline{\includegraphics[width=\textwidth]{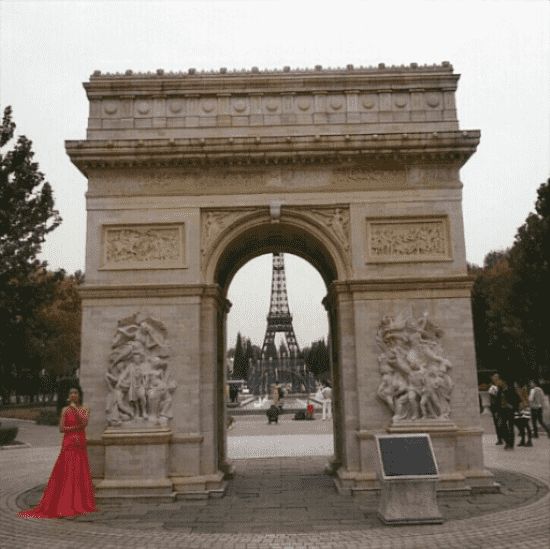}}
		\vspace{3pt}
		\centerline{\includegraphics[width=\textwidth]{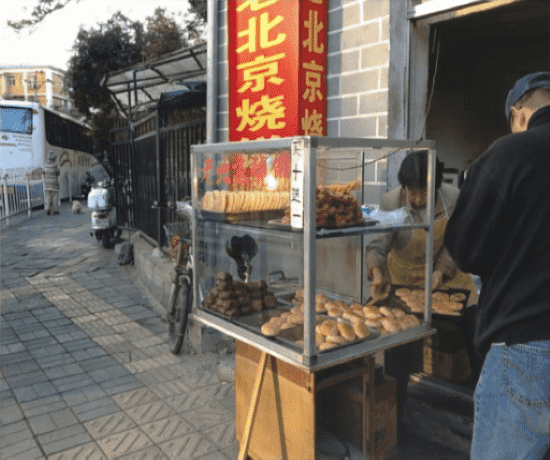}}
		\vspace{3pt}
		\centerline{\includegraphics[width=\textwidth]{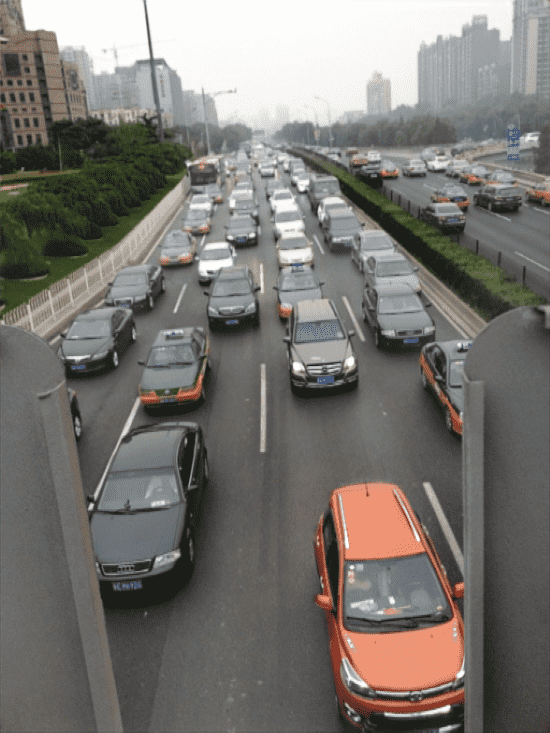}}
		\vspace{3pt}
		\centerline{(h)}
	\end{minipage}
	\begin{minipage}{0.09\linewidth}
		\vspace{3pt}
		\centerline{\includegraphics[width=\textwidth]{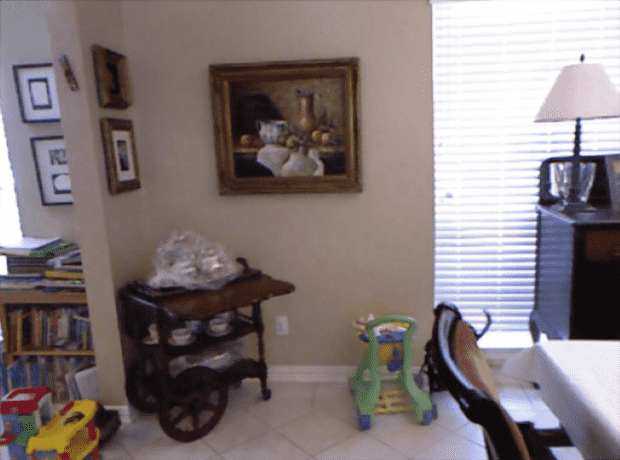}}
		\vspace{3pt}
		\centerline{\includegraphics[width=\textwidth]{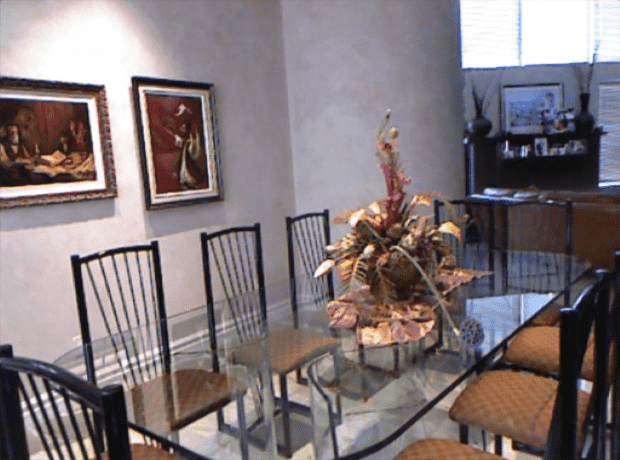}}
		\vspace{3pt}
		\centerline{\includegraphics[width=\textwidth]{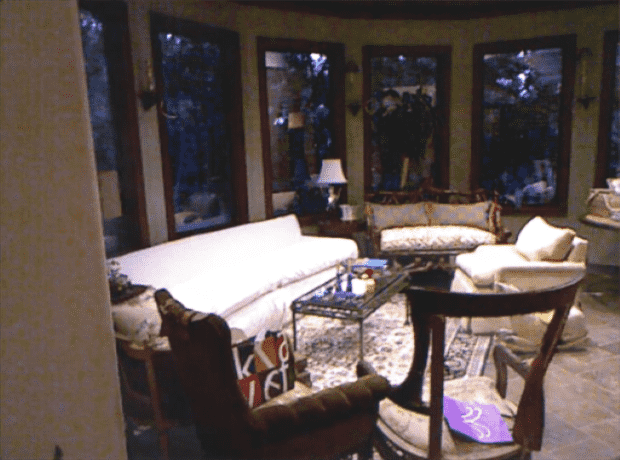}}
		\vspace{3pt}
		\centerline{\includegraphics[width=\textwidth]{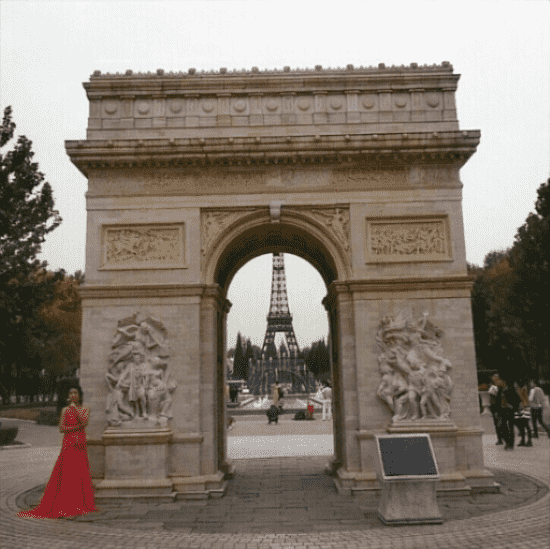}}
		\vspace{3pt}
		\centerline{\includegraphics[width=\textwidth]{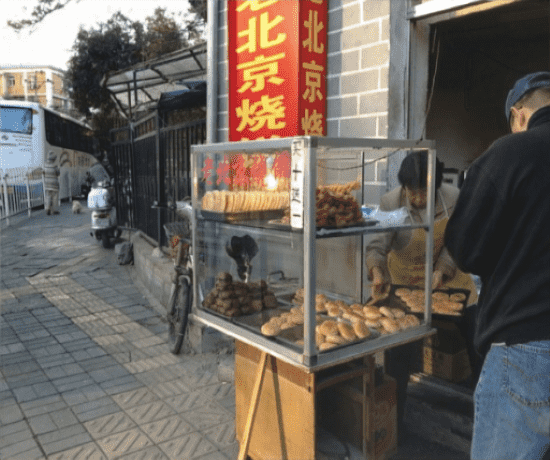}}
		\vspace{3pt}
		\centerline{\includegraphics[width=\textwidth]{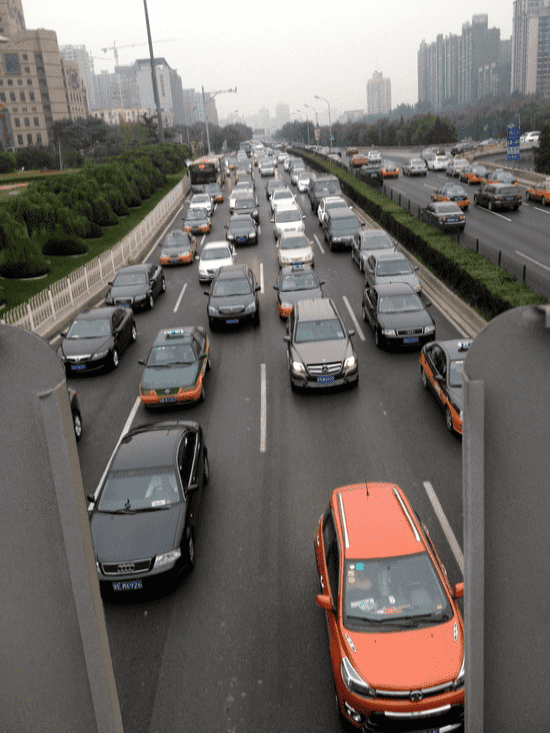}}
		\vspace{3pt}
		\centerline{(i)}
	\end{minipage}
	\begin{minipage}{0.09\linewidth}
		\vspace{3pt}
		\centerline{\includegraphics[width=\textwidth]{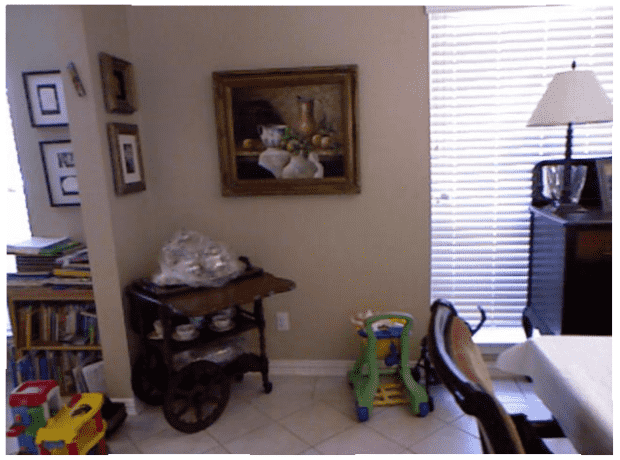}}
		\vspace{3pt}
		\centerline{\includegraphics[width=\textwidth]{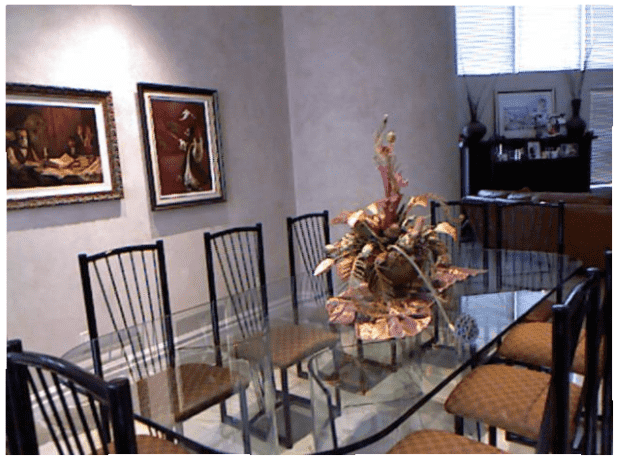}}
		\vspace{3pt}
		\centerline{\includegraphics[width=\textwidth]{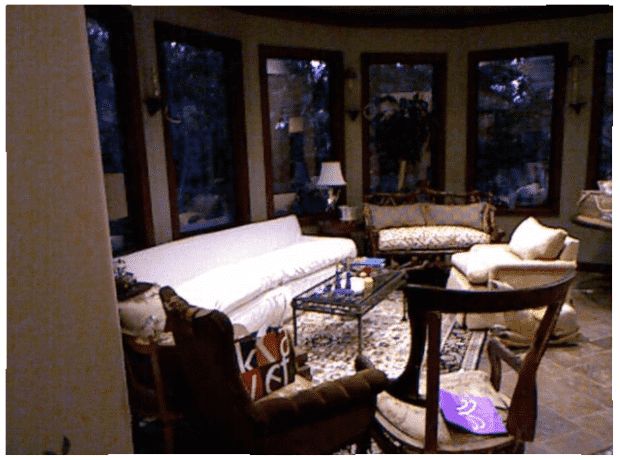}}
		\vspace{3pt}
		\centerline{\includegraphics[width=\textwidth]{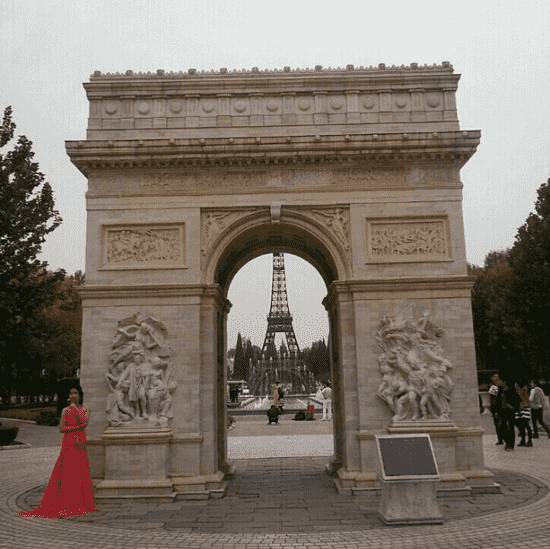}}
		\vspace{3pt}
		\centerline{\includegraphics[width=\textwidth]{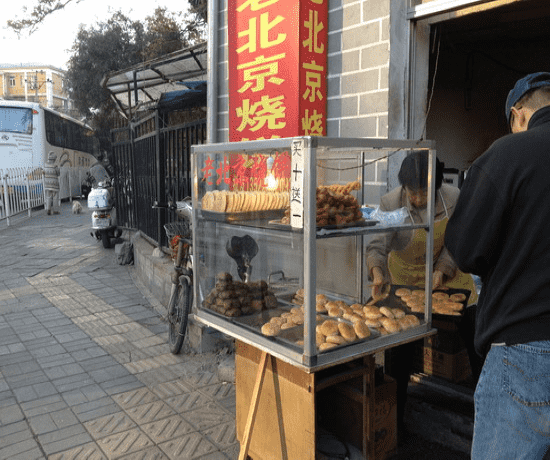}}
		\vspace{3pt}
		\centerline{\includegraphics[width=\textwidth]{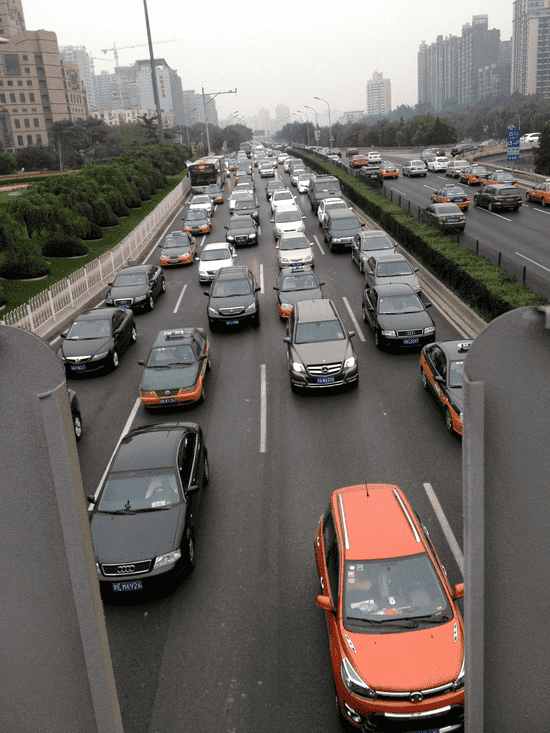}}
		\vspace{3pt}
		\centerline{(j)}
	\end{minipage}
	\caption{Qualitative comparisons on RESIDE for different methods. (a) Input hazy image, (b) DCP \cite{he2010single}, (c) DehazeNet \cite{cai2016dehazenet}, (d) AOD-Net \cite{li2017aod}, (e) DCPDN \cite{zhang2018densely}, (f) GFN \cite{ren2018gated}, (g) PFFNet \cite{mei2018progressive}, (h) EPDN \cite{qu2019enhanced}, (i) Our method, (j) Ground Truth.}
	\label{SOTS}
\end{figure*}

\begin{table}[t]
	\tiny
	\centering
	\caption{Quantitative comparisons on TrainA-TestA for different methods.}
	\vspace{12pt}
	\setlength\tabcolsep{12pt}
	\scalebox{1.5}[1.5]{
		\begin{tabular}{lcc}
			\toprule
			\multicolumn{1}{c}{\multirow{2}{*}{Method}} & \multicolumn{2}{c}{TrainA-TestA} \\ \cmidrule(l){2-3} 
			& PSNR           & SSIM                     \\ 
			\midrule
			DCP \cite{he2010single}       & 13.91 & 0.86  \\
			DehazeNet \cite{cai2016dehazenet}  & 19.92 & 0.86  \\
			MSCNN \cite{ren2016single}    & 17.98 & 0.82  \\
			GFN \cite{ren2018gated}      & 25.59 & 0.80  \\
			AOD-Net \cite{li2017aod}     & 20.46 & 0.88 \\
			DCPDN \cite{zhang2018densely}      & \underline{29.27} & \underline{0.96} \\
			\midrule
			\multicolumn{1}{c}{\textbf{Ours}}      & \textbf{30.02} & \textbf{0.97}    \\
			\bottomrule 
	   \end{tabular}}
	\label{TestA-table}
\end{table}

\begin{figure*}[h]
	\footnotesize
	\centering
	\begin{minipage}{0.1\linewidth}
		\vspace{3pt}
		\centerline{\includegraphics[width=\textwidth]{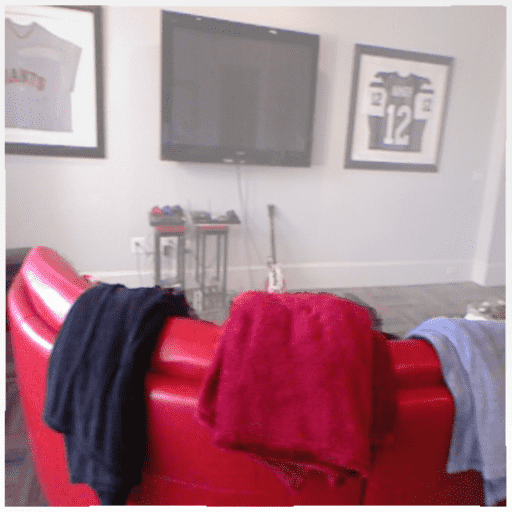}}
		\vspace{3pt}
		\centerline{\includegraphics[width=\textwidth]{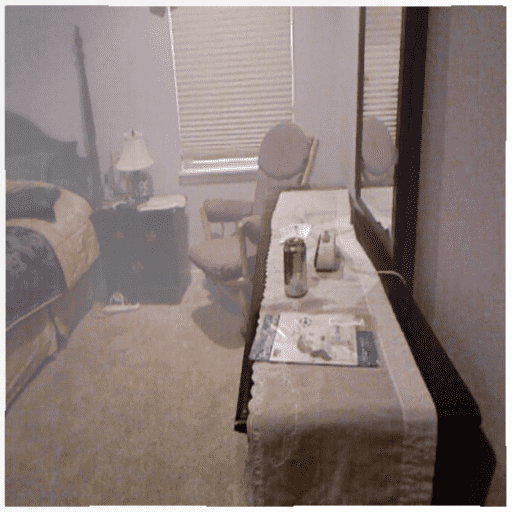}}
		\vspace{3pt}
		\centerline{\includegraphics[width=\textwidth]{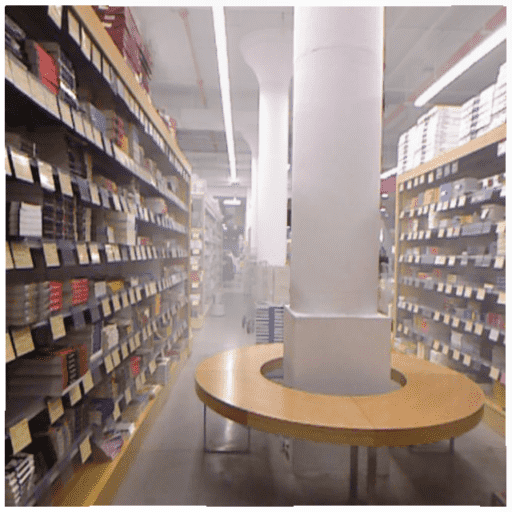}}
		\vspace{3pt}
		\centerline{\includegraphics[width=\textwidth]{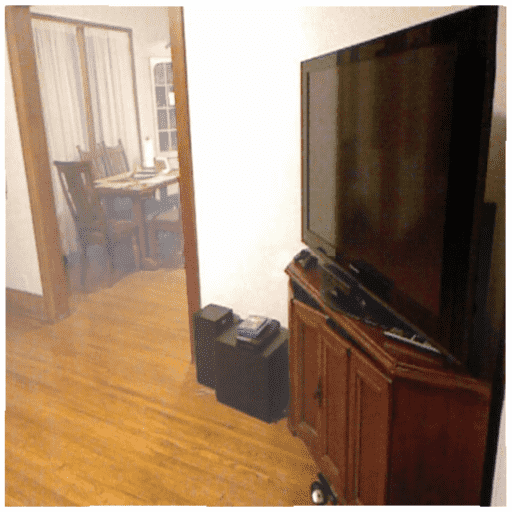}}
		\vspace{3pt}
		\centerline{(a)}
	\end{minipage}
	\begin{minipage}{0.1\linewidth}
		\vspace{3pt}
		\centerline{\includegraphics[width=\textwidth]{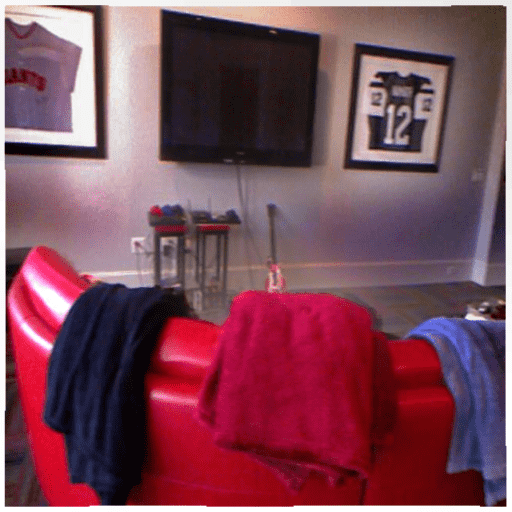}}
		\vspace{3pt}
		\centerline{\includegraphics[width=\textwidth]{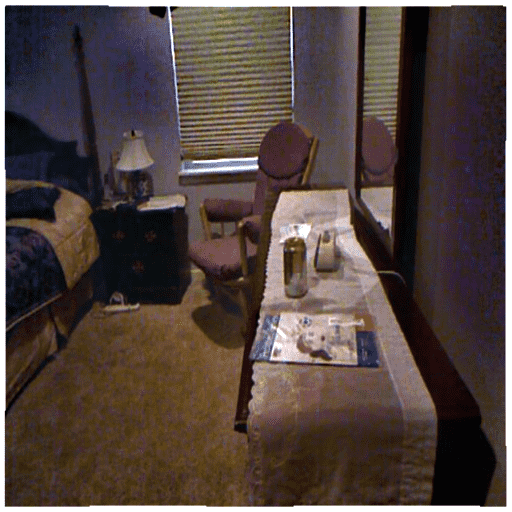}}
		\vspace{3pt}
		\centerline{\includegraphics[width=\textwidth]{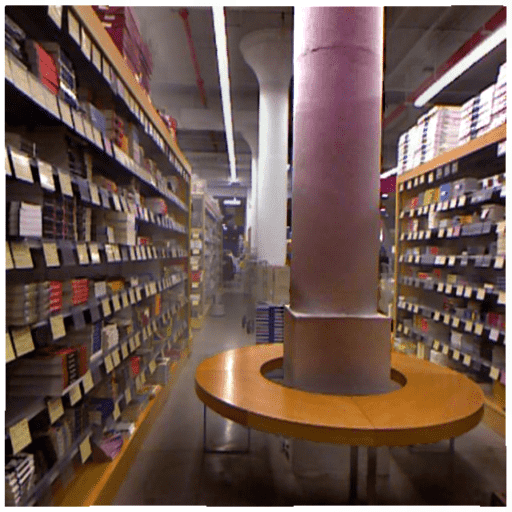}}
		\vspace{3pt}
		\centerline{\includegraphics[width=\textwidth]{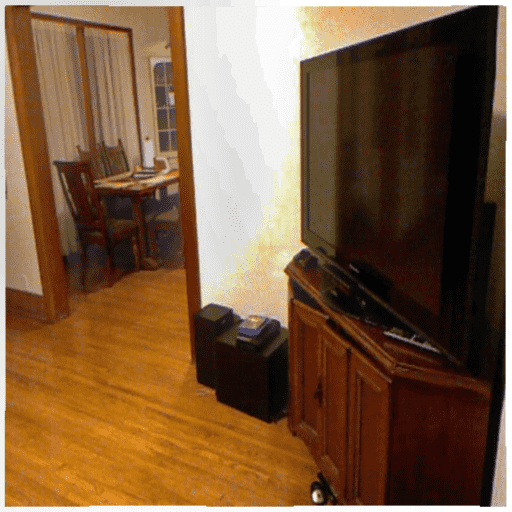}}
		\vspace{3pt}
		\centerline{(b)}
	\end{minipage}
	\begin{minipage}{0.1\linewidth}
		\vspace{3pt}
		\centerline{\includegraphics[width=\textwidth]{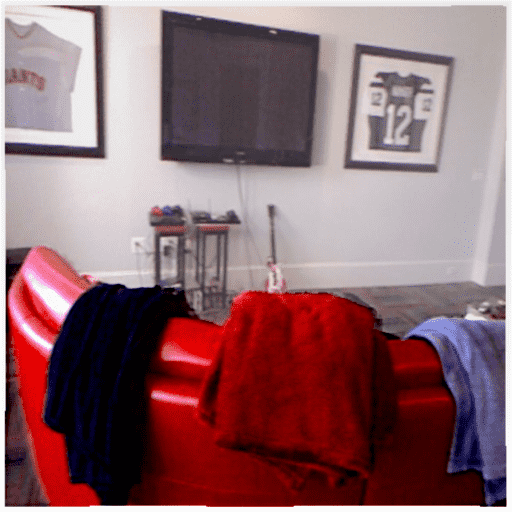}}
		\vspace{3pt}
		\centerline{\includegraphics[width=\textwidth]{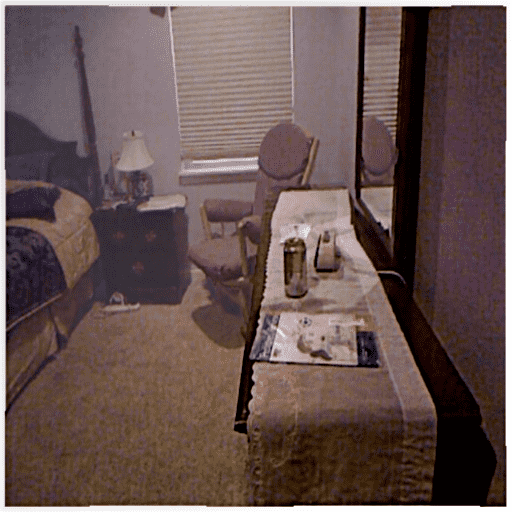}}
		\vspace{3pt}
		\centerline{\includegraphics[width=\textwidth]{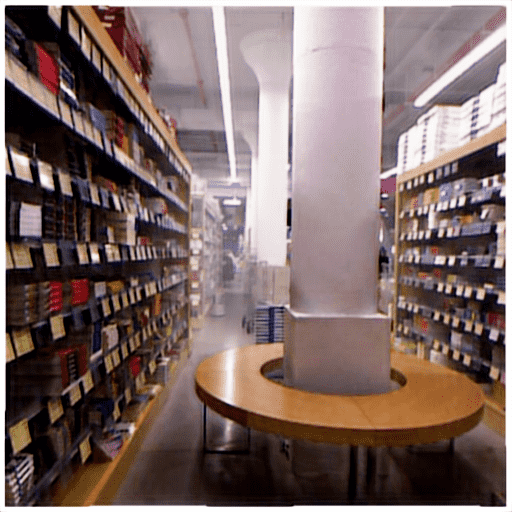}}
		\vspace{3pt}
		\centerline{\includegraphics[width=\textwidth]{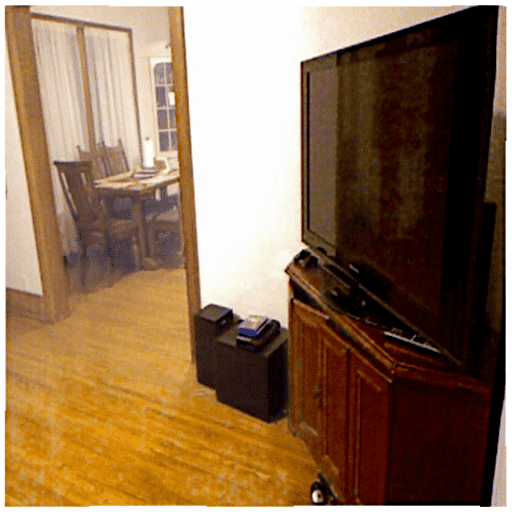}}
		\vspace{3pt}
		\centerline{(c)}
	\end{minipage}
	\begin{minipage}{0.1\linewidth}
		\vspace{3pt}
		\centerline{\includegraphics[width=\textwidth]{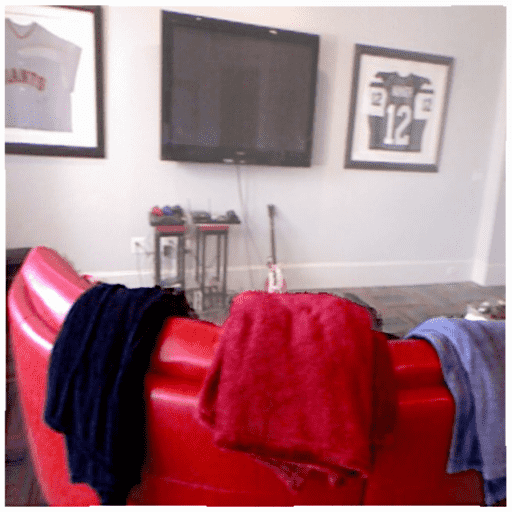}}
		\vspace{3pt}
		\centerline{\includegraphics[width=\textwidth]{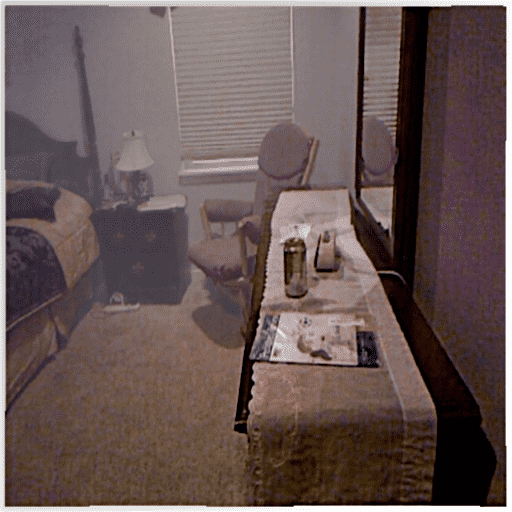}}
		\vspace{3pt}
		\centerline{\includegraphics[width=\textwidth]{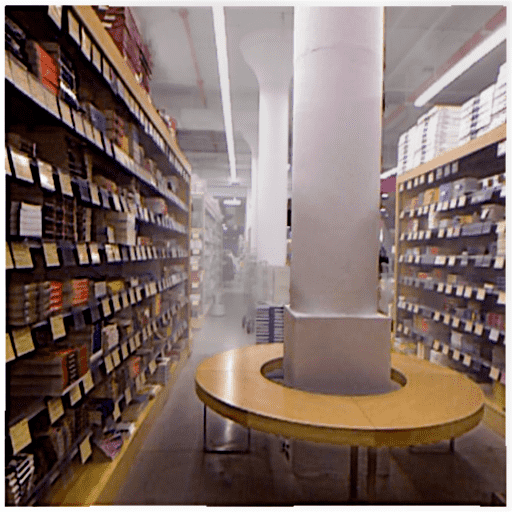}}
		\vspace{3pt}
		\centerline{\includegraphics[width=\textwidth]{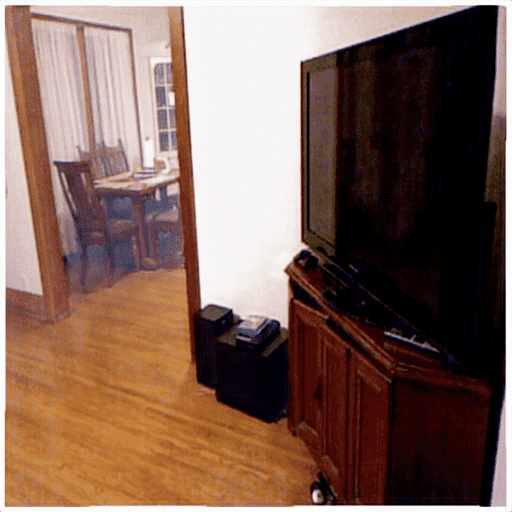}}
		\vspace{3pt}
		\centerline{(d)}
	\end{minipage}
	\begin{minipage}{0.1\linewidth}
		\vspace{3pt}
		\centerline{\includegraphics[width=\textwidth]{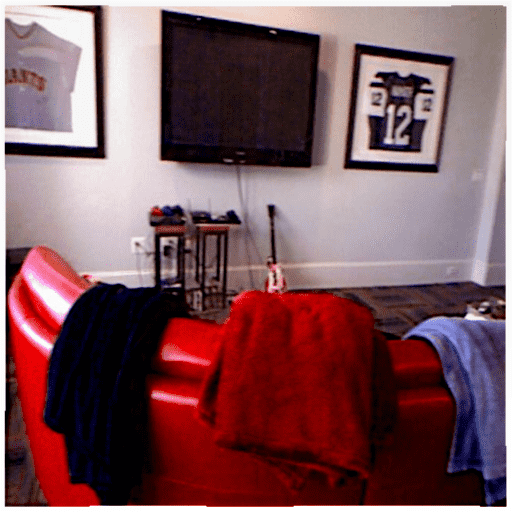}}
		\vspace{3pt}
		\centerline{\includegraphics[width=\textwidth]{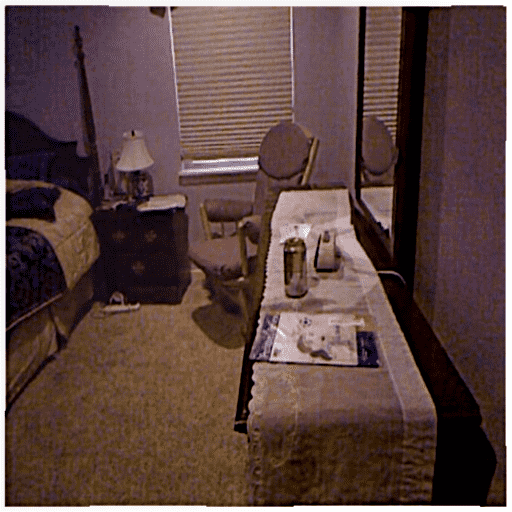}}
		\vspace{3pt}
		\centerline{\includegraphics[width=\textwidth]{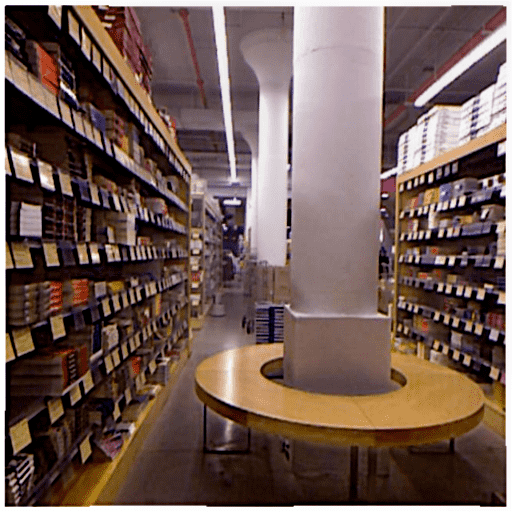}}
		\vspace{3pt}
		\centerline{\includegraphics[width=\textwidth]{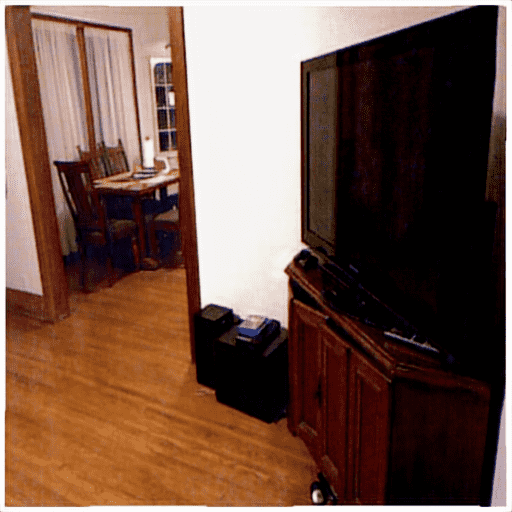}}
		\vspace{3pt}
		\centerline{(e)}
	\end{minipage}
	\begin{minipage}{0.1\linewidth}
		\vspace{3pt}
		\centerline{\includegraphics[width=\textwidth]{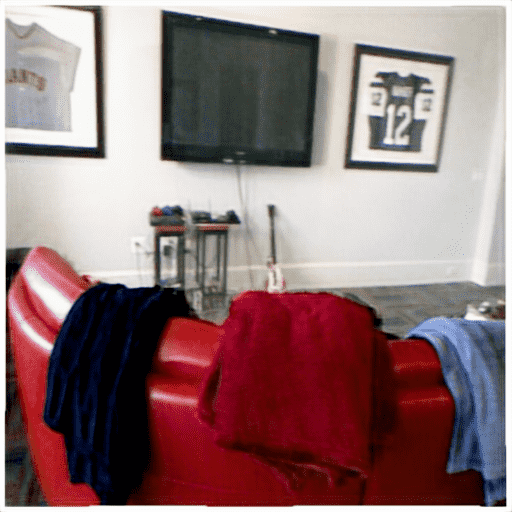}}
		\vspace{3pt}
		\centerline{\includegraphics[width=\textwidth]{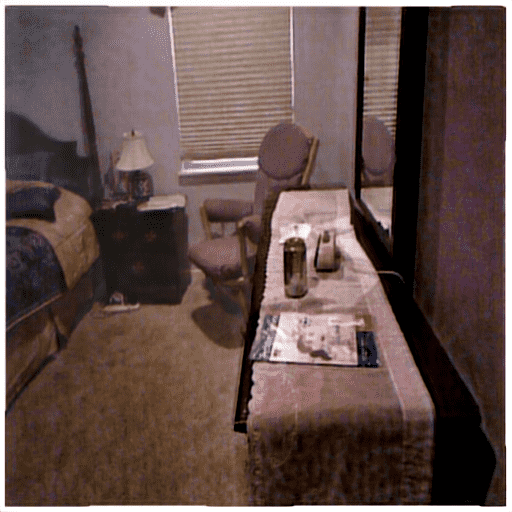}}
		\vspace{3pt}
		\centerline{\includegraphics[width=\textwidth]{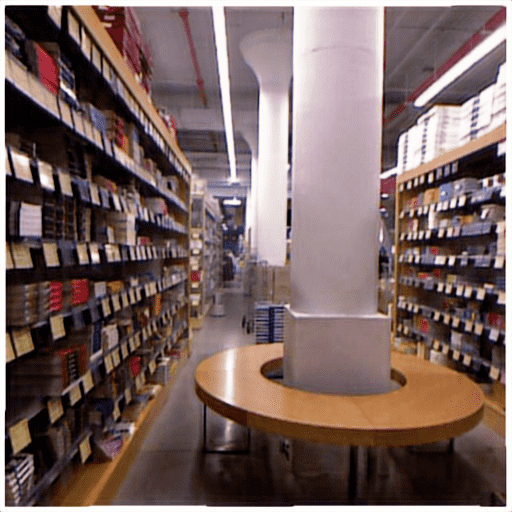}}
		\vspace{3pt}
		\centerline{\includegraphics[width=\textwidth]{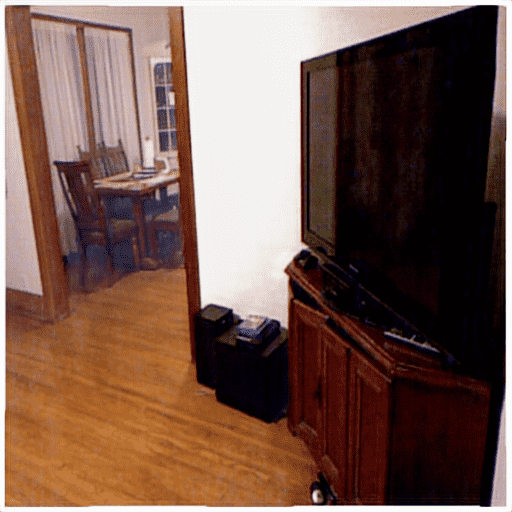}}
		\vspace{3pt}
		\centerline{(f)}
	\end{minipage}
	\begin{minipage}{0.1\linewidth}
		\vspace{3pt}
		\centerline{\includegraphics[width=\textwidth]{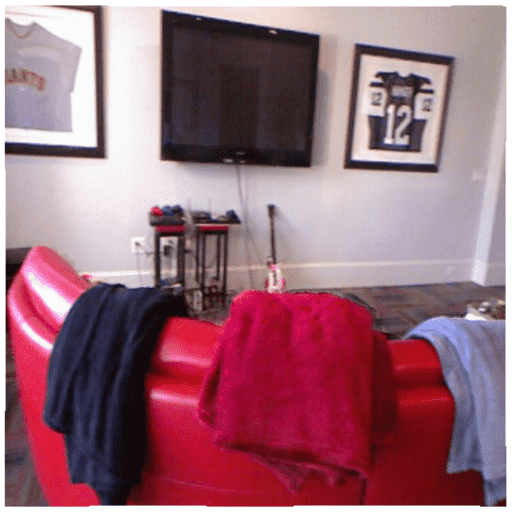}}
		\vspace{3pt}
		\centerline{\includegraphics[width=\textwidth]{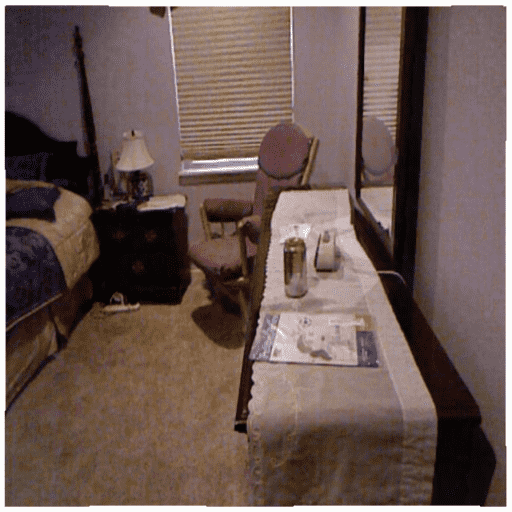}}
		\vspace{3pt}
		\centerline{\includegraphics[width=\textwidth]{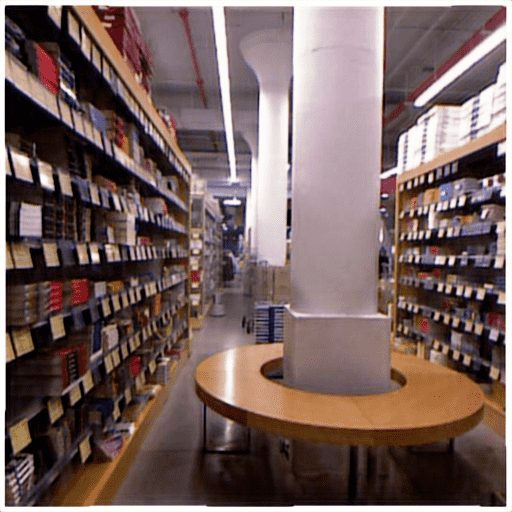}}
		\vspace{3pt}
		\centerline{\includegraphics[width=\textwidth]{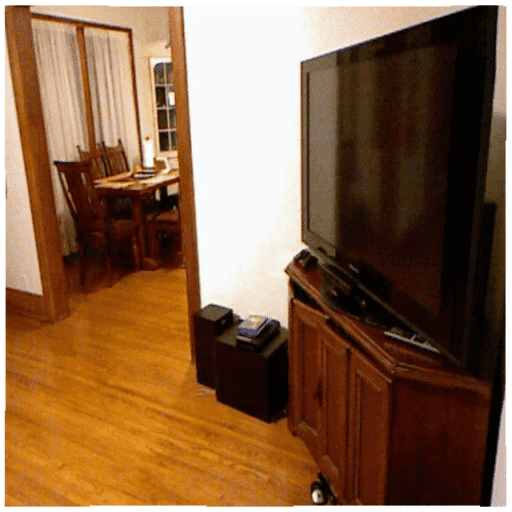}}
		\vspace{3pt}
		\centerline{(g)}
	\end{minipage}
	\begin{minipage}{0.1\linewidth}
		\vspace{3pt}
		\centerline{\includegraphics[width=\textwidth]{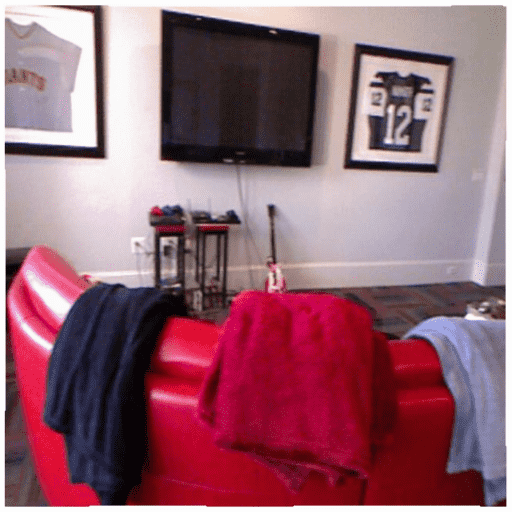}}
		\vspace{3pt}
		\centerline{\includegraphics[width=\textwidth]{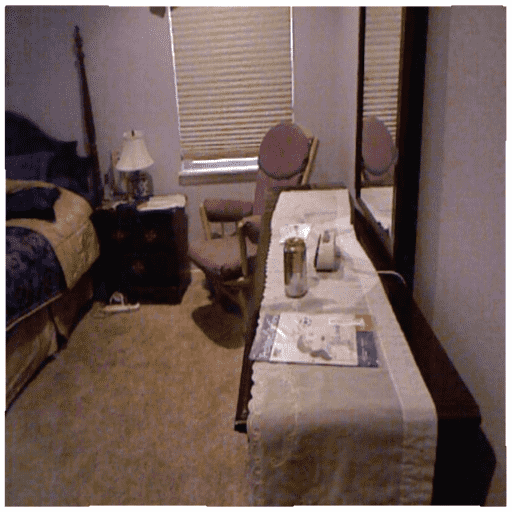}}
		\vspace{3pt}
		\centerline{\includegraphics[width=\textwidth]{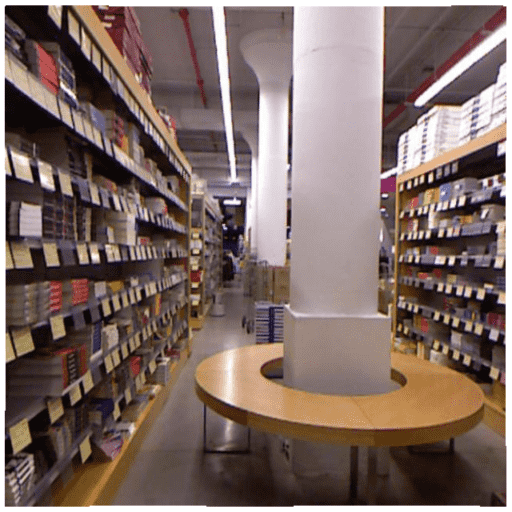}}
		\vspace{3pt}
		\centerline{\includegraphics[width=\textwidth]{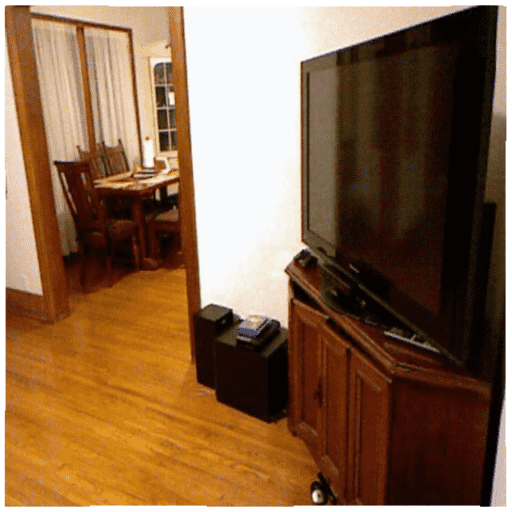}}
		\vspace{3pt}
		\centerline{(h)}
	\end{minipage}
	\begin{minipage}{0.1\linewidth}
		\vspace{3pt}
		\centerline{\includegraphics[width=\textwidth]{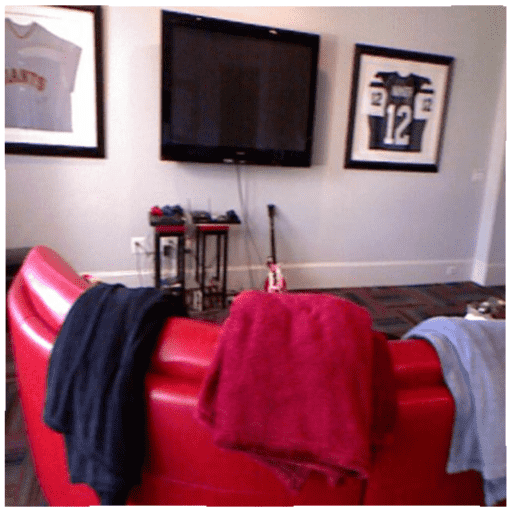}}
		\vspace{3pt}
		\centerline{\includegraphics[width=\textwidth]{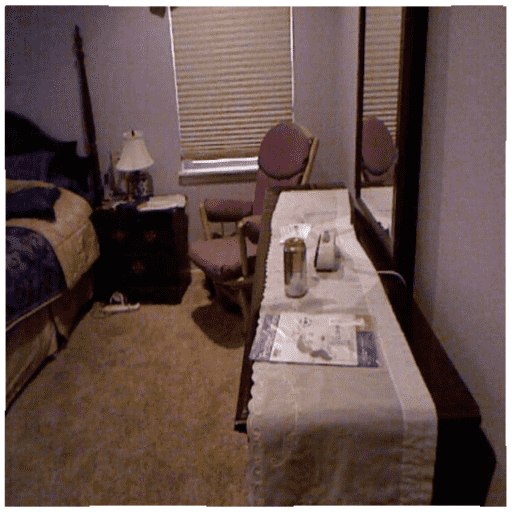}}
		\vspace{3pt}
		\centerline{\includegraphics[width=\textwidth]{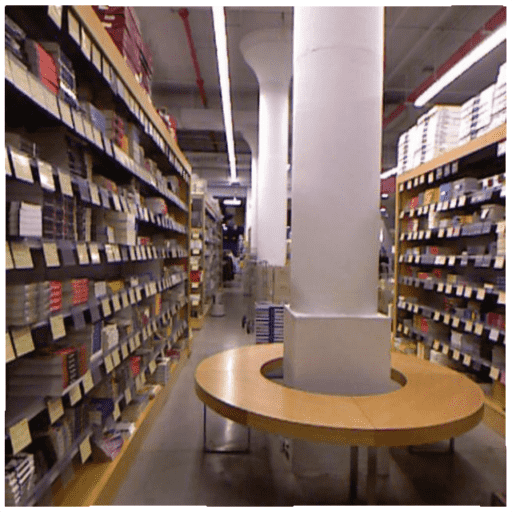}}
		\vspace{3pt}
		\centerline{\includegraphics[width=\textwidth]{compare-DCPDN-dataset-4-1.png}}
		\vspace{3pt}
		\centerline{(i)}
	\end{minipage}
	\caption{Qualitative comparisons on TrainA-TestA for different methods. (a) Input hazy image, (b) DCP \cite{he2010single}, (c) DehazeNet \cite{cai2016dehazenet}, (d) MSCNN \cite{ren2016single}, (e) GFN \cite{ren2018gated}, (f) AOD-Net \cite{li2017aod}, (g) DCPDN \cite{zhang2018densely}, (h) Ours, (i) Ground Truth.}
	\label{TestA}
\end{figure*}

\begin{figure}[htp]
	\centering  
	\begin{minipage}{0.3\linewidth}
		\vspace{3pt}
		\centerline{\includegraphics[width=\textwidth]{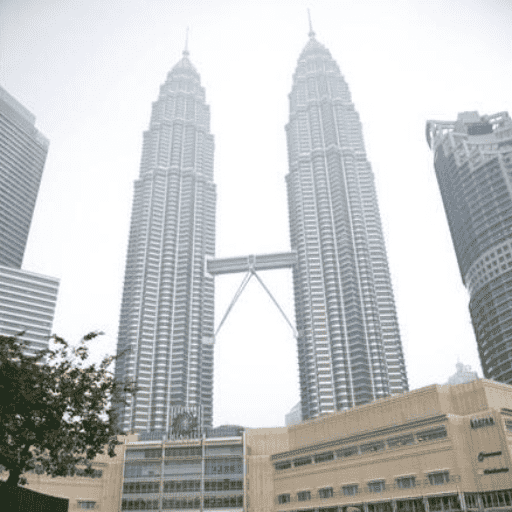}}
		\vspace{3pt}
		\centerline{\includegraphics[width=\textwidth]{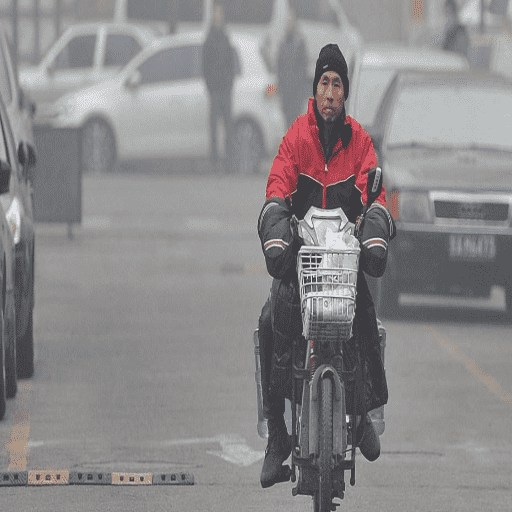}}
		\vspace{3pt}
		\centerline{\includegraphics[width=\textwidth]{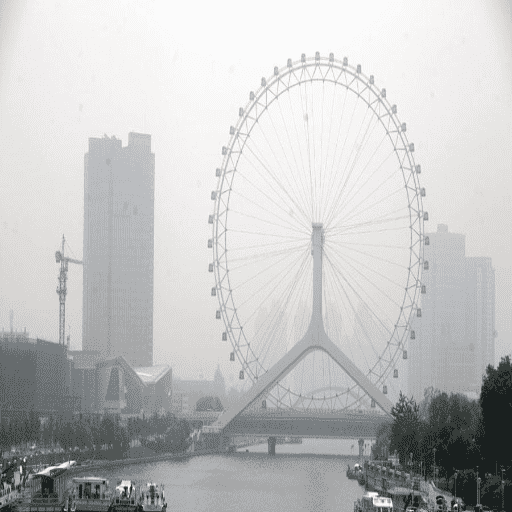}}
		\centerline{(a)}
	\end{minipage}
	\begin{minipage}{0.3\linewidth}
		\vspace{3pt}
		\centerline{\includegraphics[width=\textwidth]{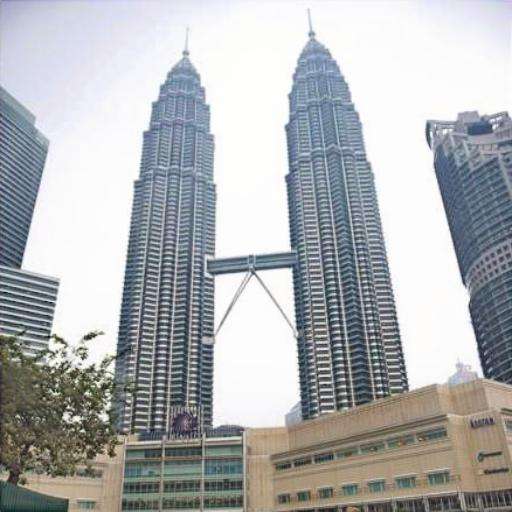}}
		\vspace{3pt}
		\centerline{\includegraphics[width=\textwidth]{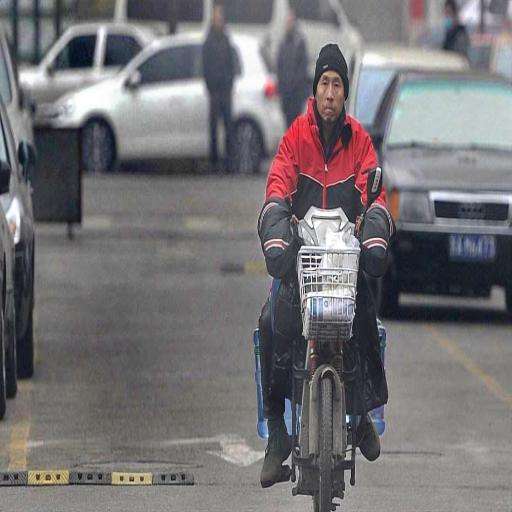}}
		\vspace{3pt}
		\centerline{\includegraphics[width=\textwidth]{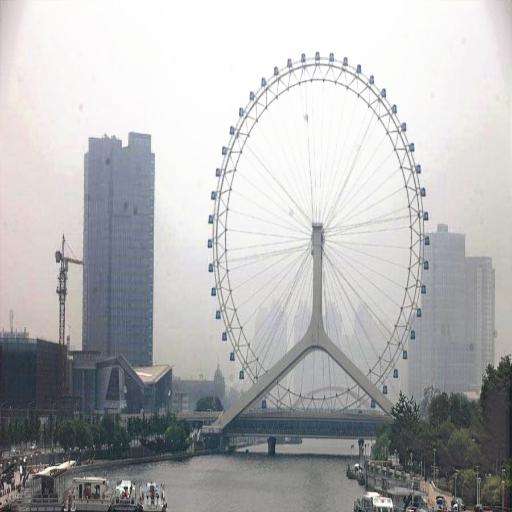}}
		\centerline{(b)}
	\end{minipage}
	\begin{minipage}{0.3\linewidth}
		\vspace{3pt}
		\centerline{\includegraphics[width=\textwidth]{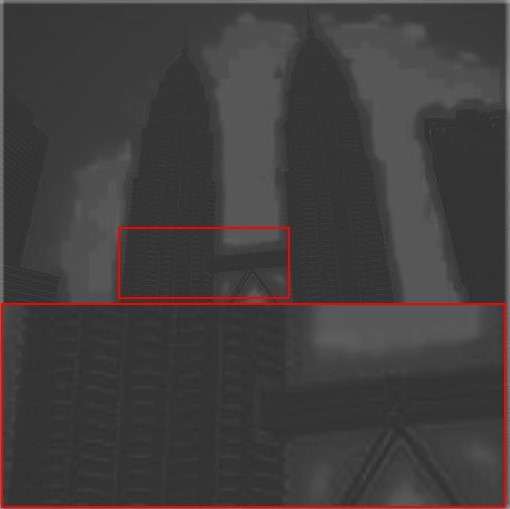}}
		\vspace{3pt}
		\centerline{\includegraphics[width=\textwidth]{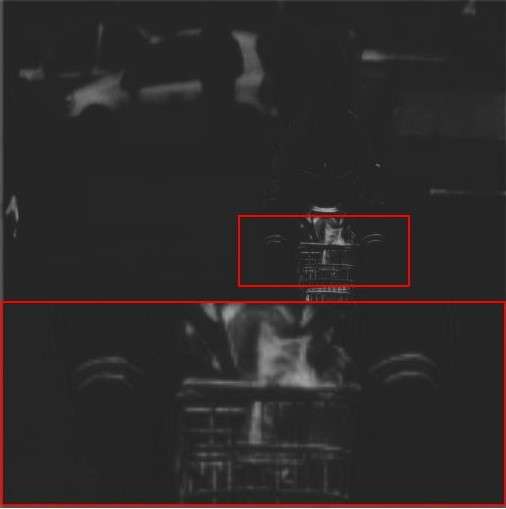}}
		\vspace{3pt}
		\centerline{\includegraphics[width=\textwidth]{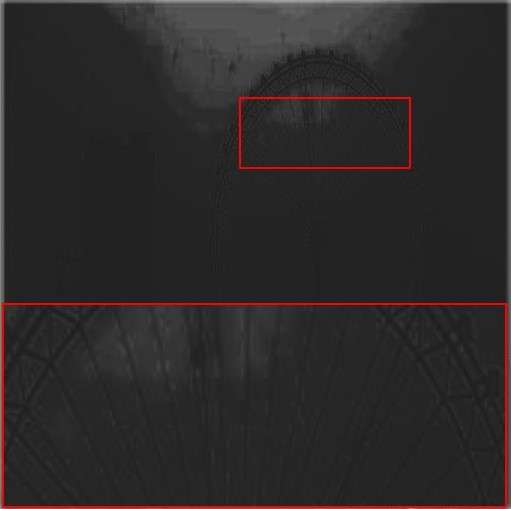}}
		\centerline{(c)}
	\end{minipage}
	\caption{The visualization of detail feature map. (a) Input hazy image, (b) Dehazed image, (c) Detail feature map.}
	\label{detail_map}
	\vspace{-10pt}
\end{figure}

\begin{figure*}[]
	\footnotesize
	\begin{minipage}{0.105\linewidth}
		\vspace{3pt}
		\centerline{\includegraphics[width=\textwidth]{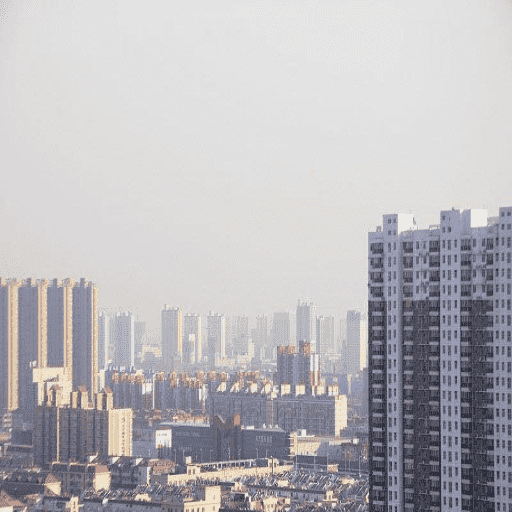}}
		\vspace{3pt}
		\centerline{\includegraphics[width=\textwidth]{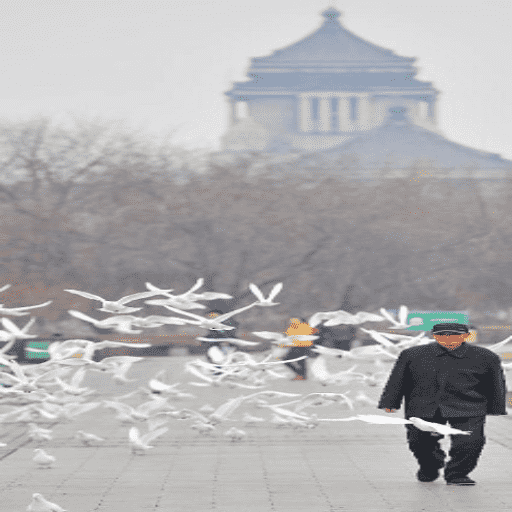}}
		\vspace{3pt}
		\centerline{\includegraphics[width=\textwidth]{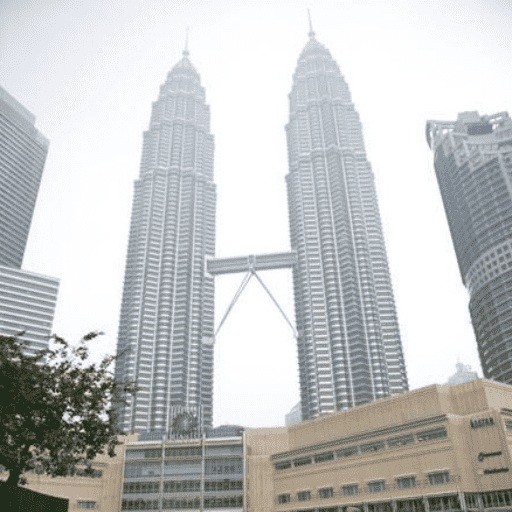}}
		\vspace{3pt}
		\centerline{(a)}
	\end{minipage}
	\begin{minipage}{0.105\linewidth}
		\vspace{3pt}
		\centerline{\includegraphics[width=\textwidth]{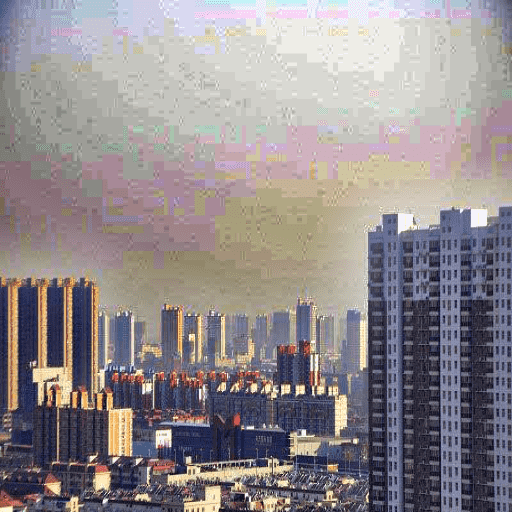}}
		\vspace{3pt}
		\centerline{\includegraphics[width=\textwidth]{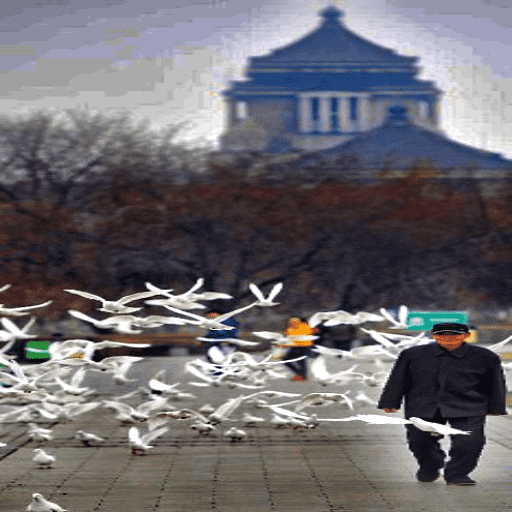}}
		\vspace{3pt}
		\centerline{\includegraphics[width=\textwidth]{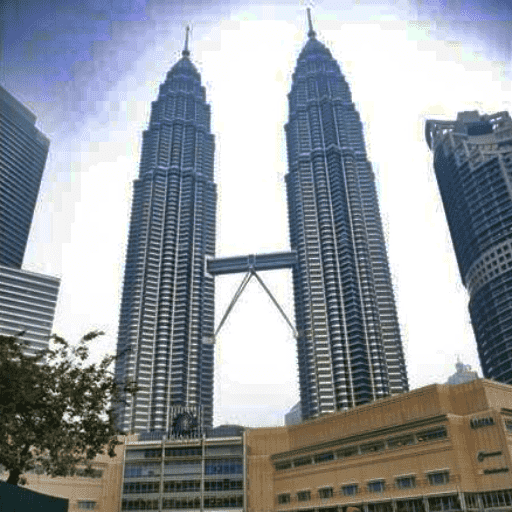}}
		\vspace{3pt}
		\centerline{(b)}
	\end{minipage}
	\begin{minipage}{0.105\linewidth}
		\vspace{3pt}
		\centerline{\includegraphics[width=\textwidth]{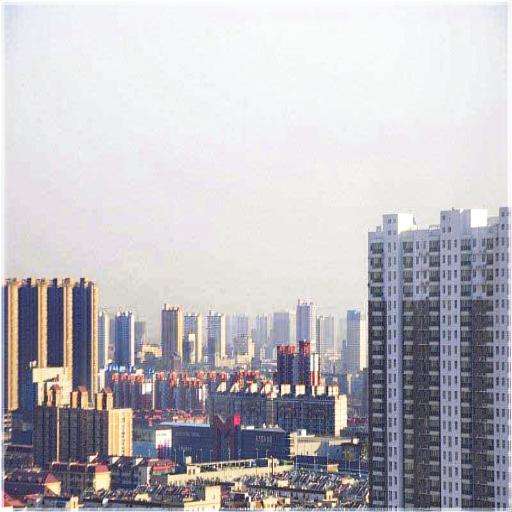}}
		\vspace{3pt}
		\centerline{\includegraphics[width=\textwidth]{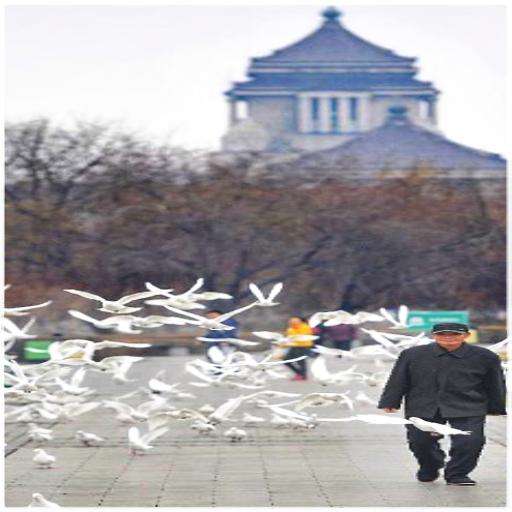}}
		\vspace{3pt}
		\centerline{\includegraphics[width=\textwidth]{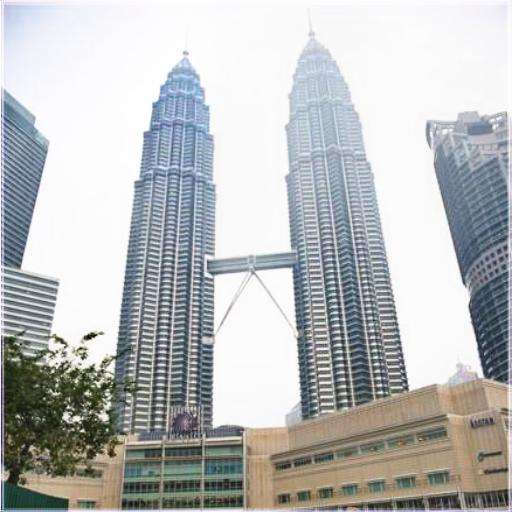}}
		\vspace{3pt}
		\centerline{(c)}
	\end{minipage}
	\begin{minipage}{0.105\linewidth}
		\vspace{3pt}
		\centerline{\includegraphics[width=\textwidth]{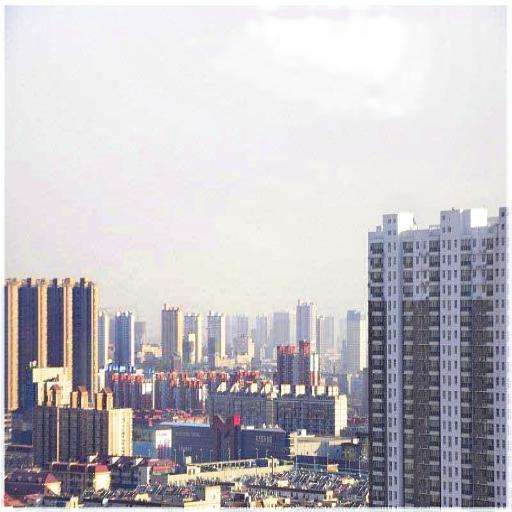}}
		\vspace{3pt}
		\centerline{\includegraphics[width=\textwidth]{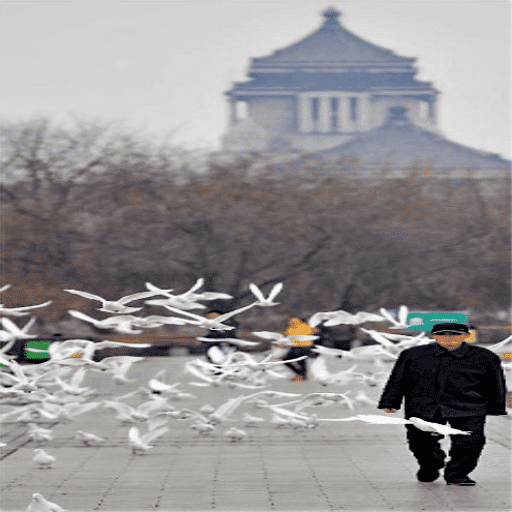}}
		\vspace{3pt}
		\centerline{\includegraphics[width=\textwidth]{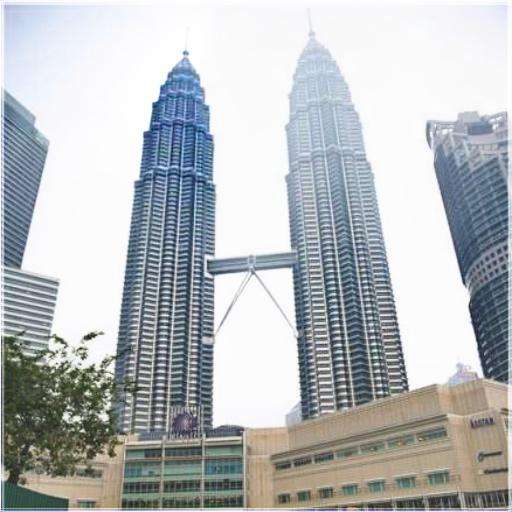}}
		\vspace{3pt}
		\centerline{(d)}
	\end{minipage}
	\begin{minipage}{0.105\linewidth}
		\vspace{3pt}
		\centerline{\includegraphics[width=\textwidth]{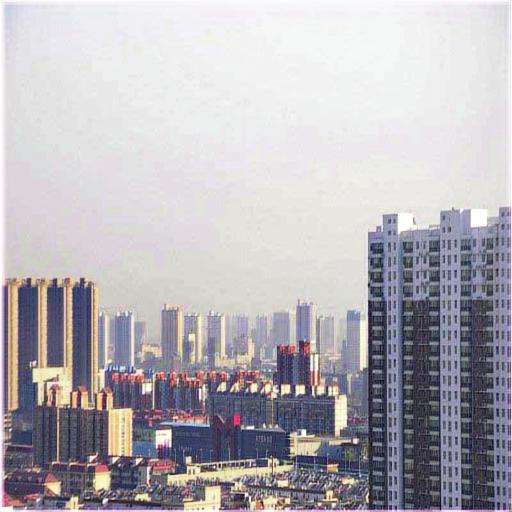}}
		\vspace{3pt}
		\centerline{\includegraphics[width=\textwidth]{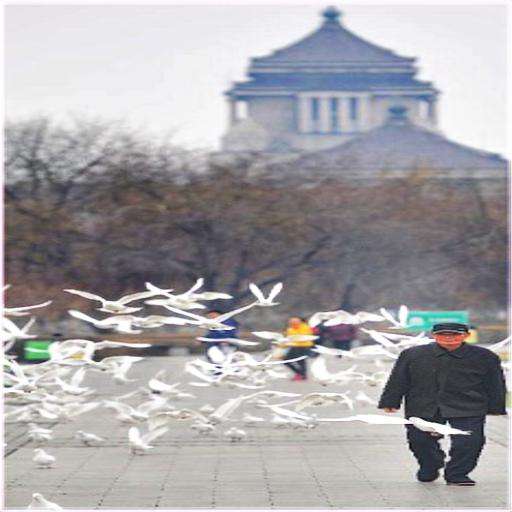}}
		\vspace{3pt}
		\centerline{\includegraphics[width=\textwidth]{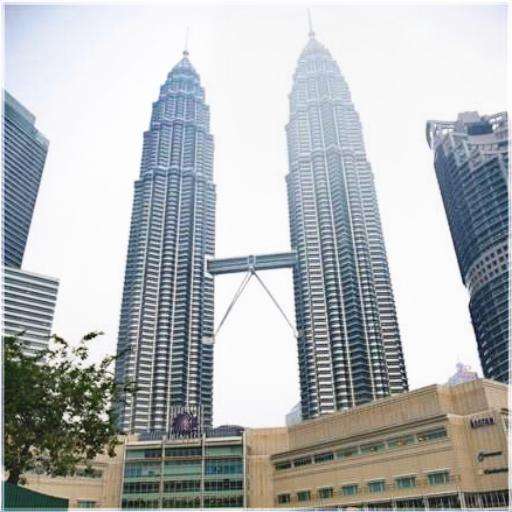}}
		\vspace{3pt}
		\centerline{(e)}
	\end{minipage}
	\begin{minipage}{0.105\linewidth}
		\vspace{3pt}
		\centerline{\includegraphics[width=\textwidth]{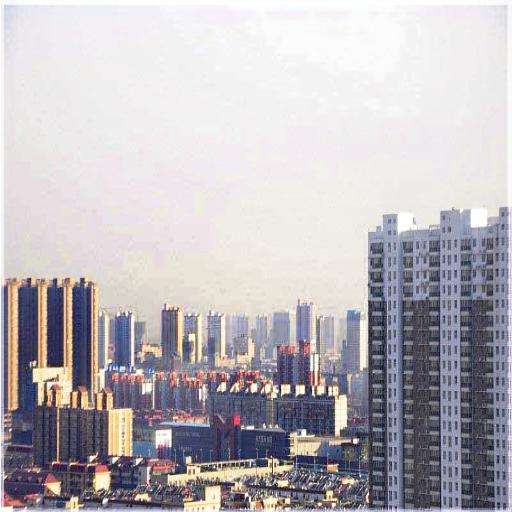}}
		\vspace{3pt}
		\centerline{\includegraphics[width=\textwidth]{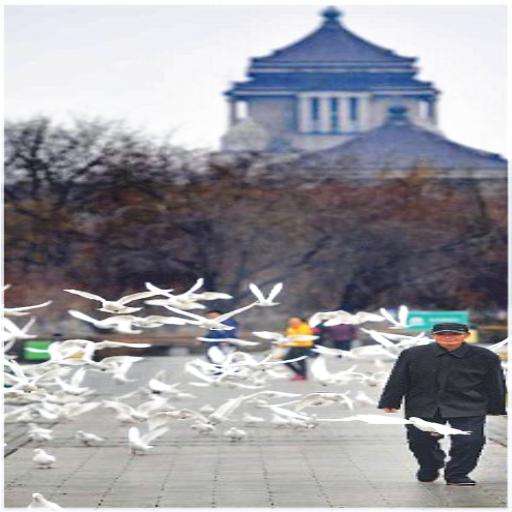}}
		\vspace{3pt}
		\centerline{\includegraphics[width=\textwidth]{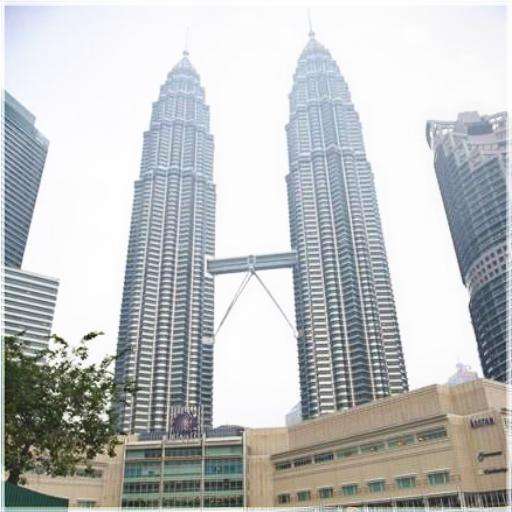}}
		\vspace{3pt}
		\centerline{(f)}
	\end{minipage}
	\begin{minipage}{0.105\linewidth}
		\vspace{3pt}
		\centerline{\includegraphics[width=\textwidth]{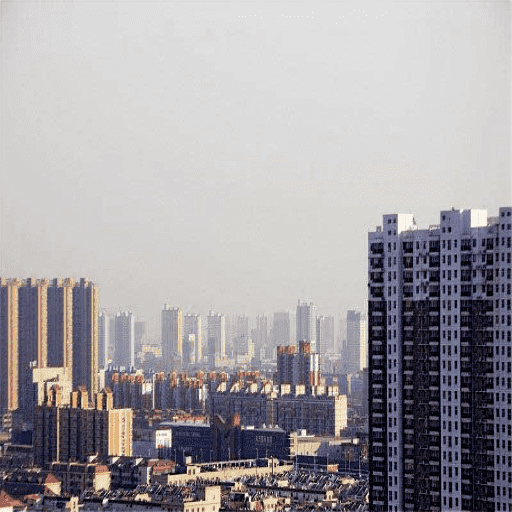}}
		\vspace{3pt}
		\centerline{\includegraphics[width=\textwidth]{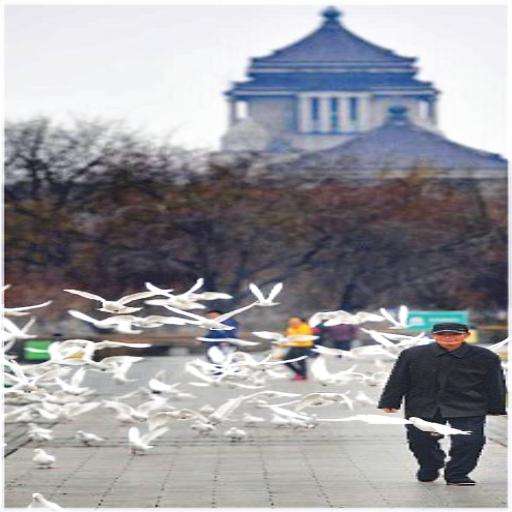}}
		\vspace{3pt}
		\centerline{\includegraphics[width=\textwidth]{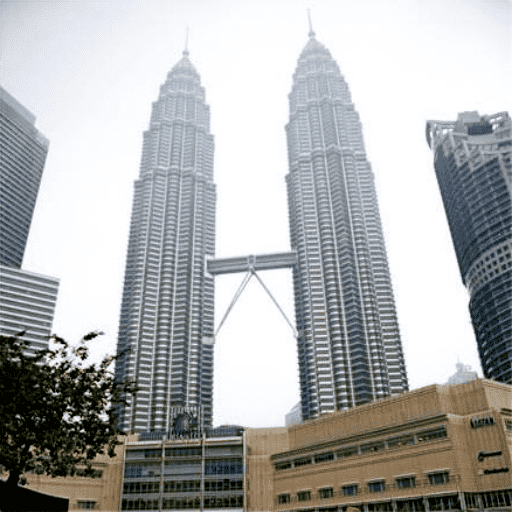}}
		\vspace{3pt}
		\centerline{(g)}
	\end{minipage}
	\begin{minipage}{0.105\linewidth}
		\vspace{3pt}
		\centerline{\includegraphics[width=\textwidth]{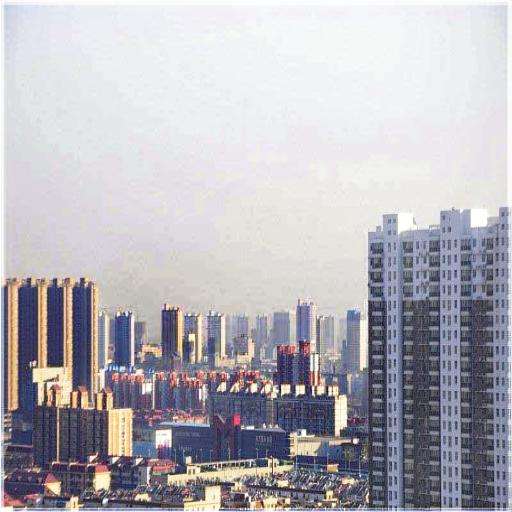}}
		\vspace{3pt}
		\centerline{\includegraphics[width=\textwidth]{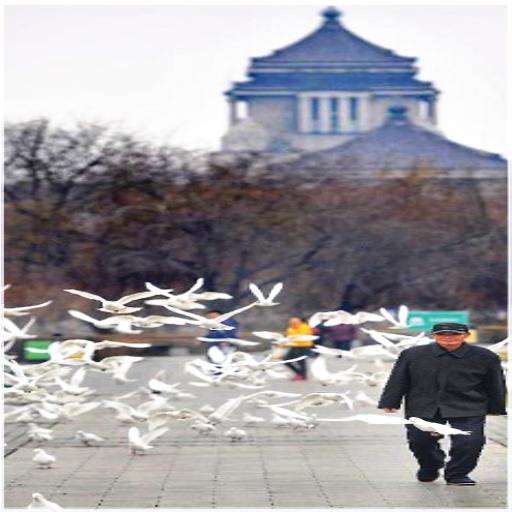}}
		\vspace{3pt}
		\centerline{\includegraphics[width=\textwidth]{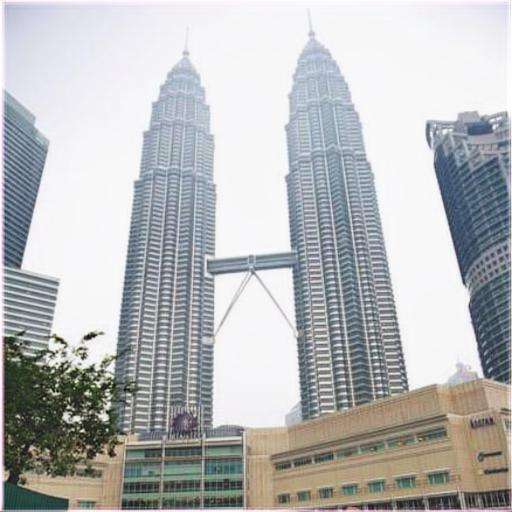}}
		\vspace{3pt}
		\centerline{(h)}
	\end{minipage}
	\begin{minipage}{0.105\linewidth}
		\vspace{3pt}
		\centerline{\includegraphics[width=\textwidth]{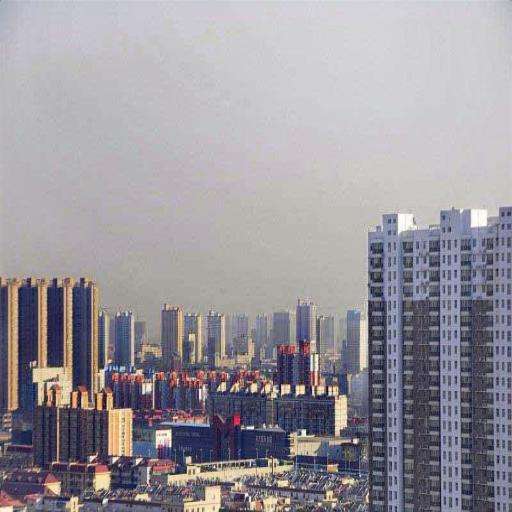}}
		\vspace{3pt}
		\centerline{\includegraphics[width=\textwidth]{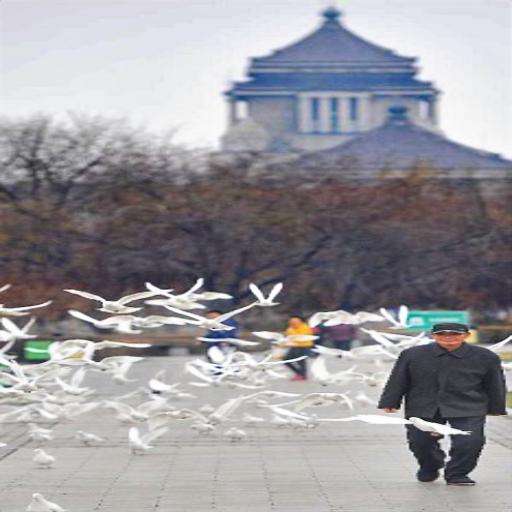}}
		\vspace{3pt}
		\centerline{\includegraphics[width=\textwidth]{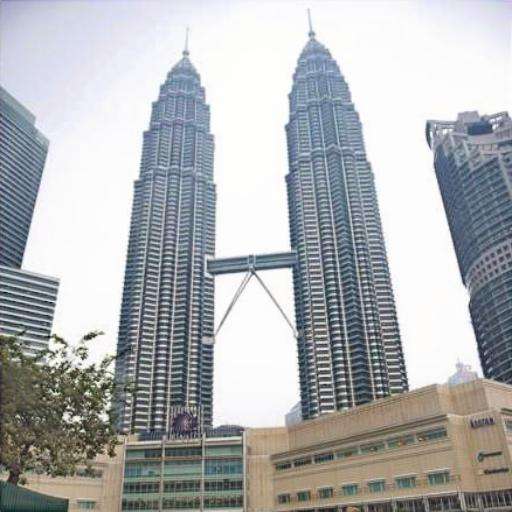}}
		\vspace{3pt}
		\centerline{(i)}
	\end{minipage}
	\caption{Qualitative comparisons on URHI for different methods. (a) Input hazy image, (b) DCP \cite{he2010single}, (c) DehazeNet \cite{cai2016dehazenet}, (d) MSCNN \cite{ren2016single}, (e) AOD-Net \cite{li2017aod}, (f) DCPDN \cite{zhang2018densely}, (g) GFN \cite{ren2018gated}, (h) EPDN \cite{qu2019enhanced}, (i) Ours.}
	\label{real}
\end{figure*}

\section{Experiments}
\subsection{Datasets and Metrics}
\textbf{RESIDE} We adopt the RESIDE dataset to train and test our method, which is a large-scale synthetic hazy image dataset proposed in \cite{li2019benchmarking}. The training set of RESIDE contains 13,990 hazy images which are synthesized using 1,399 clear images from the NYU Depth Dataset V2 \cite{silberman2012indoor} and the Middlebury stereo \cite{scharstein2003high}. The testing set, named Synthetic Objective Testing Set (SOTS), selects 500 indoor images and 500 outdoor ones from the NYU Depth Dataset V2 to synthesize hazy images. Here we name them as RESIDE-Indoor and RESIDE-Outdoor, respectively. In order to test the dehazing effect on the real hazy images, we use URHI (Unannotated Real Hazy Images) as the test dataset, which is a subset of the RESIDE dataset.

\textbf{TrainA-TestA} We train and test our method on the dataset prepared by \cite{zhang2018densely} against the state-of-the-art methods.
The dataset consists of TrainA and TestA. Specifically, TrainA contains 4,000 synthetic hazy images which are generated by using 1,000 clear images randomly selected from the NYU Depth Dataset V2. Correspondingly, the testing set, named TestA, consists of 400 (100 $\times$ 4) images and follows the same process to generate the training set. Note that TestA non-overlap with TrainA. These images are synthesized with the following parameters: $A$ $\in$ [0.5, 1.0] and $\gamma \in$ [0.4, 1.6].

The performance of dehazing methods is evaluated with PSNR and SSIM, which are the most widely used metrics in dehazing methods.

\subsection{Implementation Details}
The proposed network is end-to-end trainable without pre-training for sub-modules. Our experiments are carried out on one NVIDIA GeForce RTX 2080Ti. The proposed method is implemented with PyTorch.
During training, we use Adam \cite{kingma2014adam} as optimizer, where $\beta_{1}$ and $\beta_{2}$ take the default values of 0.9 and 0.999, respectively. The initial learning rate is set to 1 $\times$ $10^{-4}$. The learning rate decreases uniformly with every epoch and drops to 0 at the end of training.

\begin{figure}[h]
	\footnotesize
	\begin{minipage}{0.19\linewidth}
		\vspace{3pt}
		\centerline{\includegraphics[width=\textwidth]{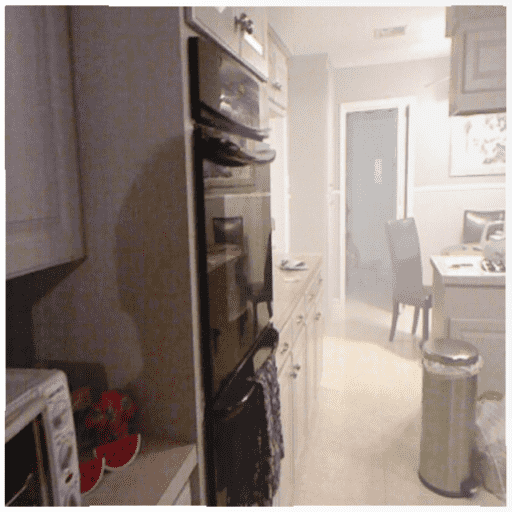}}
		\vspace{3pt}
		\centerline{\includegraphics[width=\textwidth]{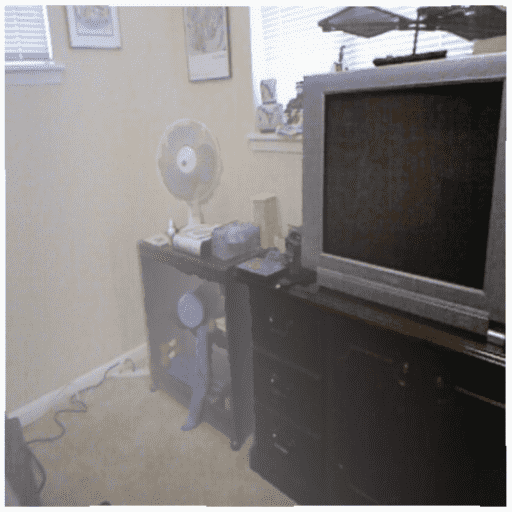}}
		\vspace{3pt}
		\centerline{\includegraphics[width=\textwidth]{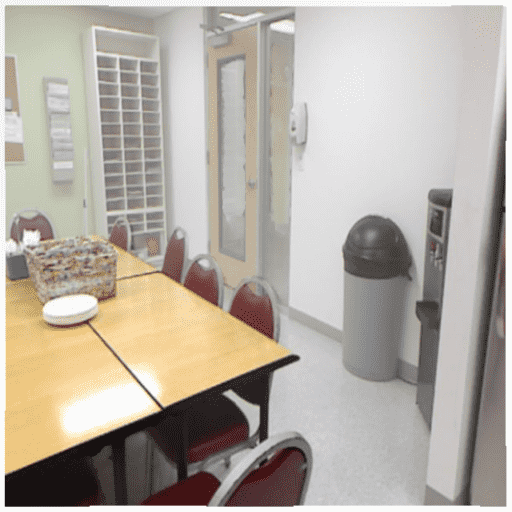}}
		\vspace{3pt}
		\centerline{Input}
	\end{minipage}
	\begin{minipage}{0.19\linewidth}
		\vspace{3pt}
		\centerline{\includegraphics[width=\textwidth]{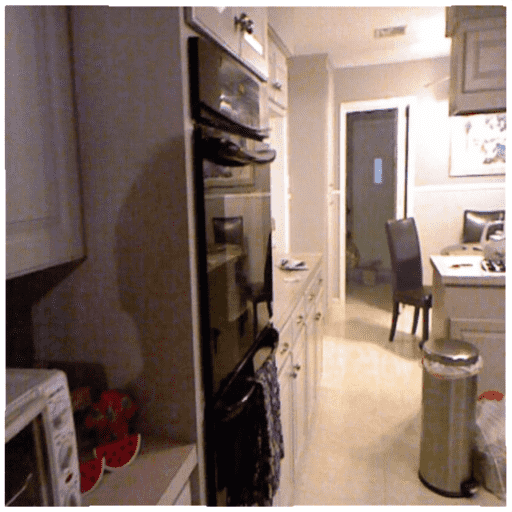}}
		\vspace{3pt}
		\centerline{\includegraphics[width=\textwidth]{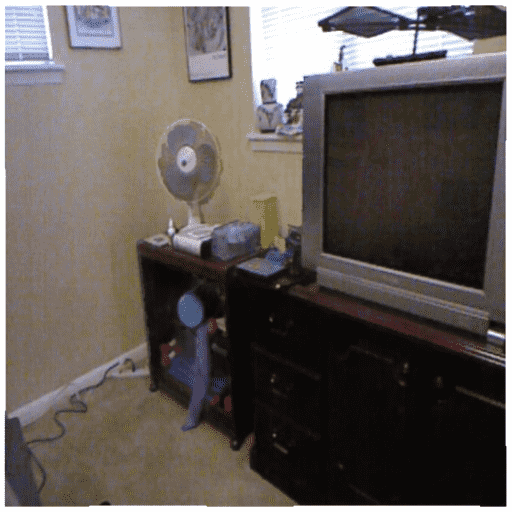}}
		\vspace{3pt}
		\centerline{\includegraphics[width=\textwidth]{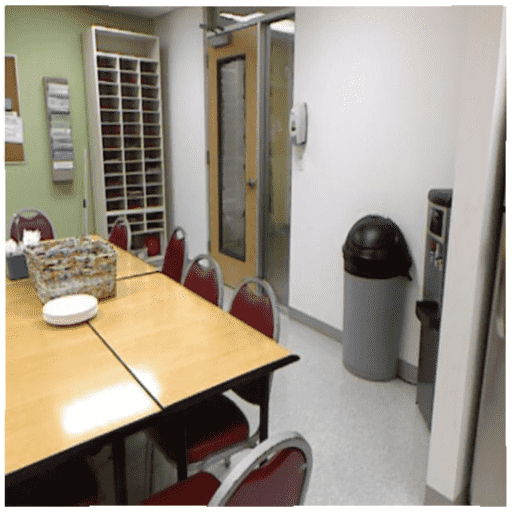}}
		\vspace{3pt}
		\centerline{(1)}
	\end{minipage}
	\begin{minipage}{0.19\linewidth}
		\vspace{3pt}
		\centerline{\includegraphics[width=\textwidth]{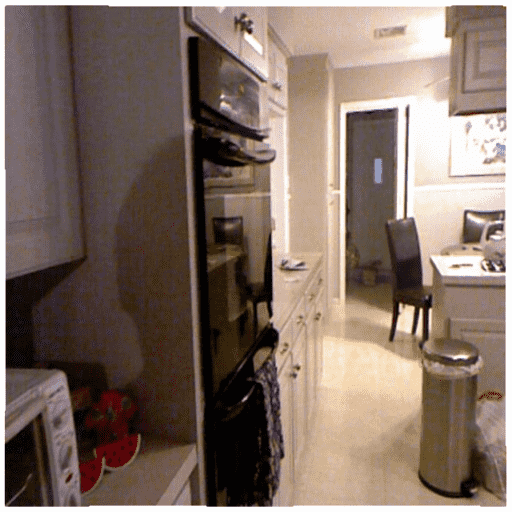}}
		\vspace{3pt}
		\centerline{\includegraphics[width=\textwidth]{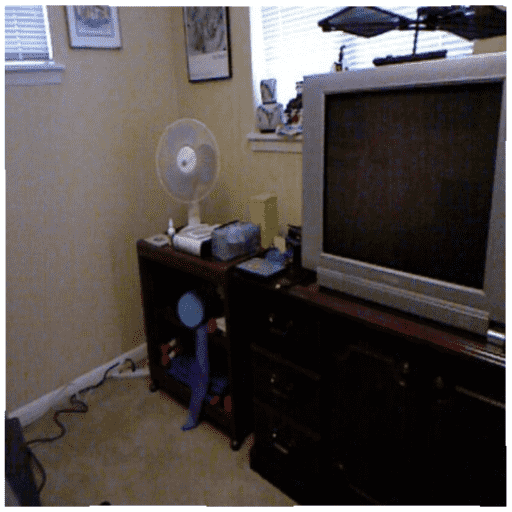}}
		\vspace{3pt}
		\centerline{\includegraphics[width=\textwidth]{ablation-3-2.png}}
		\vspace{3pt}
		\centerline{(2)}
	\end{minipage}
	\begin{minipage}{0.19\linewidth}
		\vspace{3pt}
		\centerline{\includegraphics[width=\textwidth]{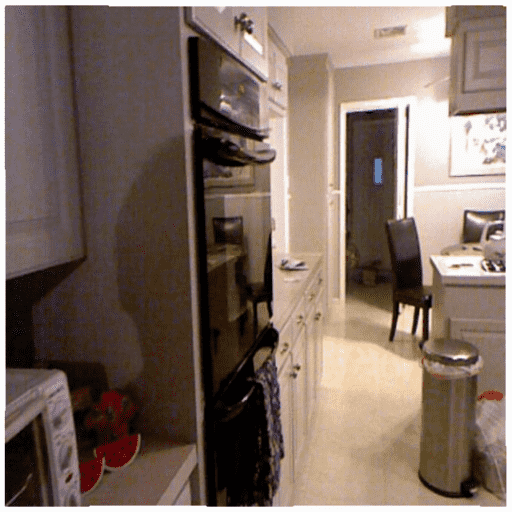}}
		\vspace{3pt}
		\centerline{\includegraphics[width=\textwidth]{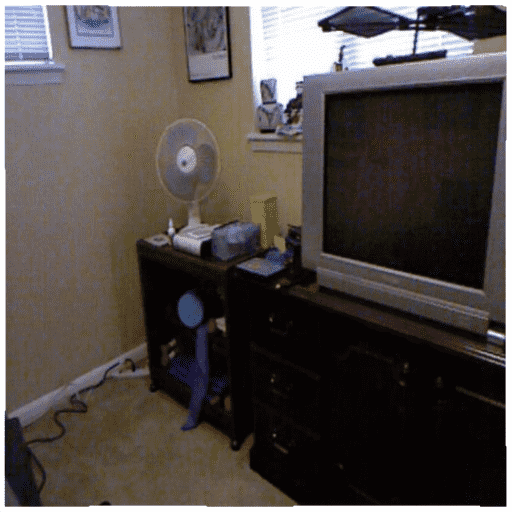}}
		\vspace{3pt}
		\centerline{\includegraphics[width=\textwidth]{ablation-3-3.png}}
		\vspace{3pt}
		\centerline{(3)}
	\end{minipage}
	\begin{minipage}{0.19\linewidth}
		\vspace{3pt}
		\centerline{\includegraphics[width=\textwidth]{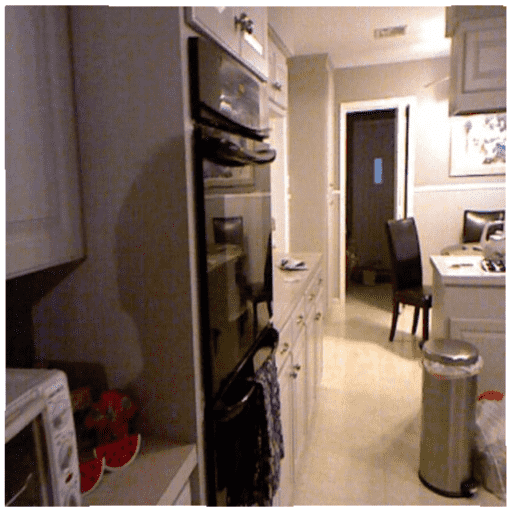}}
		\vspace{3pt}
		\centerline{\includegraphics[width=\textwidth]{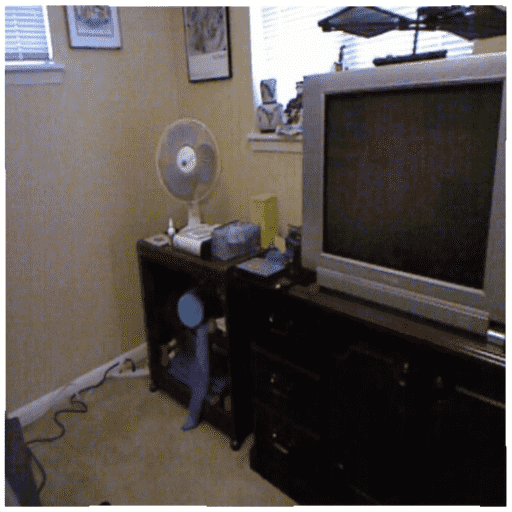}}
		\vspace{3pt}
		\centerline{\includegraphics[width=\textwidth]{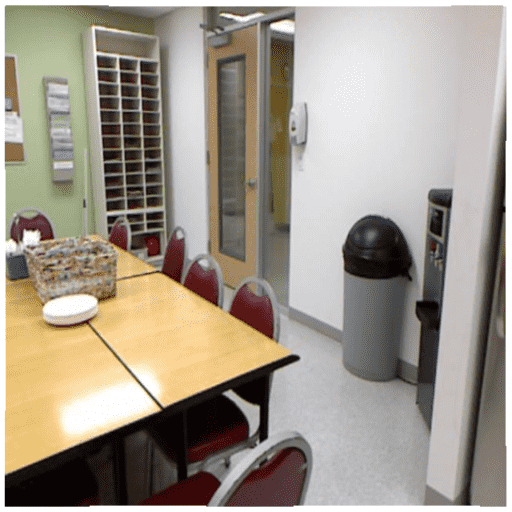}}
		\vspace{3pt}
		\centerline{GT}
	\end{minipage}
	\caption{Qualitative comparisons for different configurations of our network in ablation study. (1) Base, (2) Base+DRN (w/o SDC), (3) Ours (Base+DRN).}
	\label{ablation-image}
\end{figure}

\begin{table}[]
	\footnotesize
	\centering
	\caption{The ablation experiments by considering different configurations of our network on TrainA-TestA.}
	\setlength\tabcolsep{8pt}
	\vspace{10pt}
	\scalebox{1.0}[1.0]{
		\begin{tabular}{lccc}
			\toprule
			Name& SDC & PSNR & SSIM \\
			\midrule
			Base & --- & 28.95 & 0.9568 \\
			Base+DRN (w/o SDC)&---& 29.85 & 0.9691 \\
			\textbf{Ours (Base+DRN)} & \textbf{\checkmark} &\textbf{30.02} & \textbf{0.9726} \\
			\bottomrule
	\end{tabular}}
	\label{ablation-table}
\end{table}

\begin{table}[]
	\footnotesize
	\centering
	\caption{Comparison of loss functions used to train our model on TrainA-TestA.}
	\setlength\tabcolsep{4pt}
	\vspace{10pt}
	\scalebox{1.0}[1.0]{
		\begin{tabular}{ccccc}
			\toprule
			Pixel-wise Loss ($L_{d}+L_{t}+L_{a}$)&\checkmark &\checkmark  &\checkmark\\
			Perceptual Loss     &   &  \checkmark   &  \checkmark \\
			Reconstruction Loss &   &               &  \checkmark \\
			\midrule
			SSIM                & 0.9705  & 0.9720   & \textbf{0.9726}  \\
			PSNR                & 29.77  & 29.98  &  \textbf{30.02} \\
			\bottomrule
	\end{tabular}}
	\label{loss}
\end{table}

\begin{figure*}[t]
	\footnotesize
	\centering
	\begin{minipage}{0.3\linewidth}
		\centerline{\includegraphics[width=\textwidth]{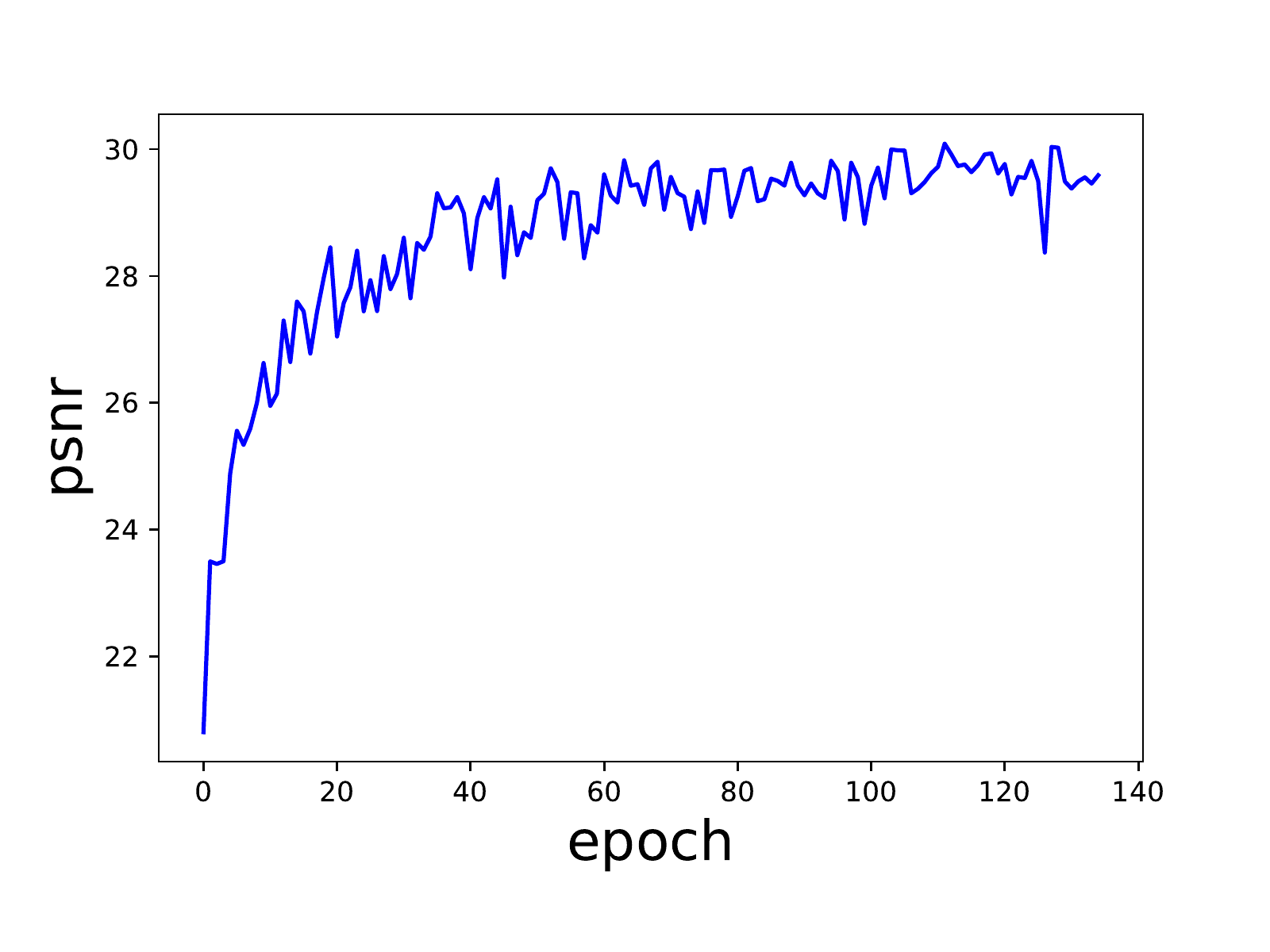}}
		\centerline{(a)} 
	\end{minipage} 
	\begin{minipage}{0.3\linewidth}
		\centerline{\includegraphics[width=\textwidth]{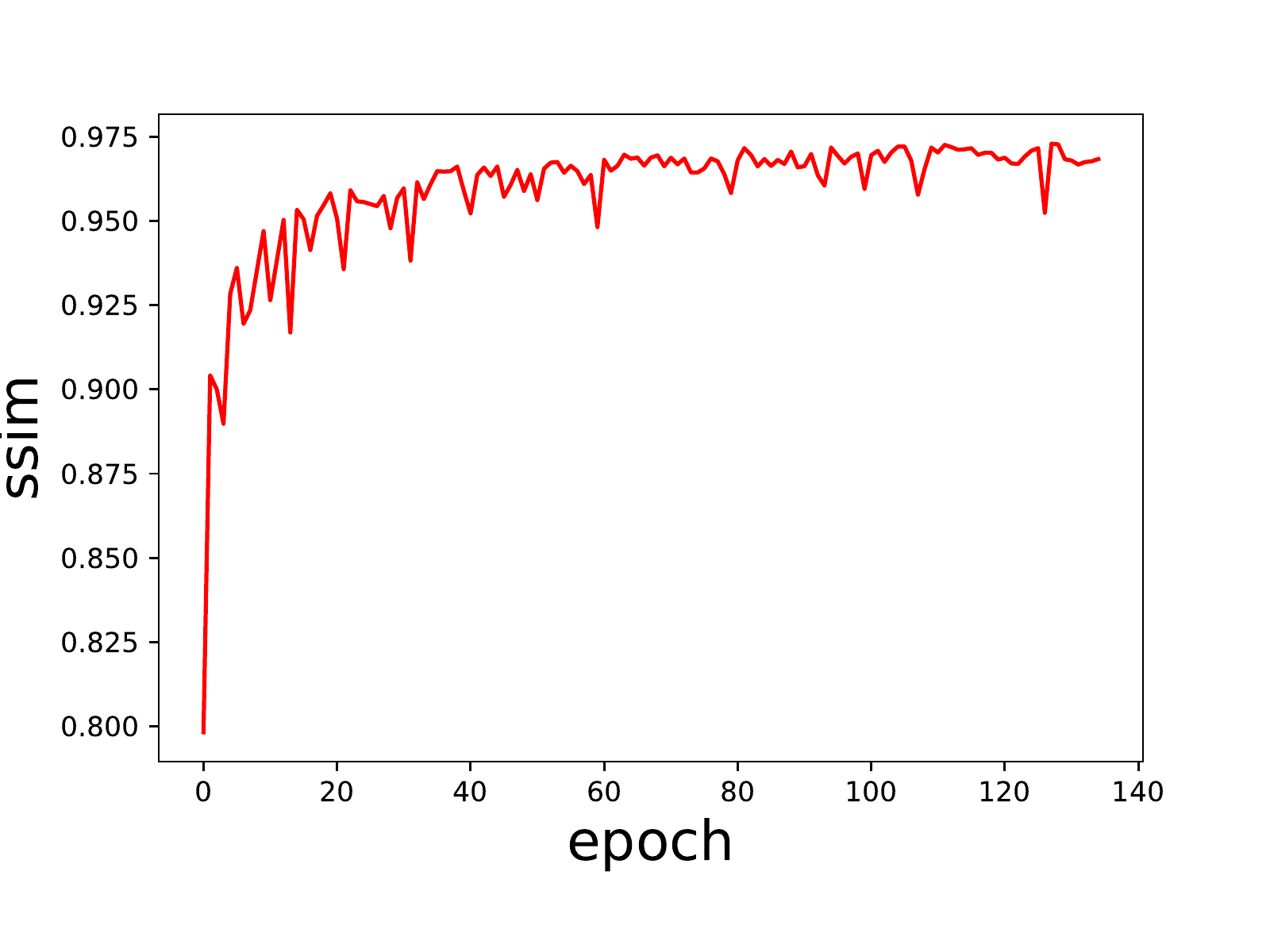}}
		\centerline{(b)} 
	\end{minipage} 
	\begin{minipage}{0.3\linewidth}
		\centerline{\includegraphics[width=\textwidth]{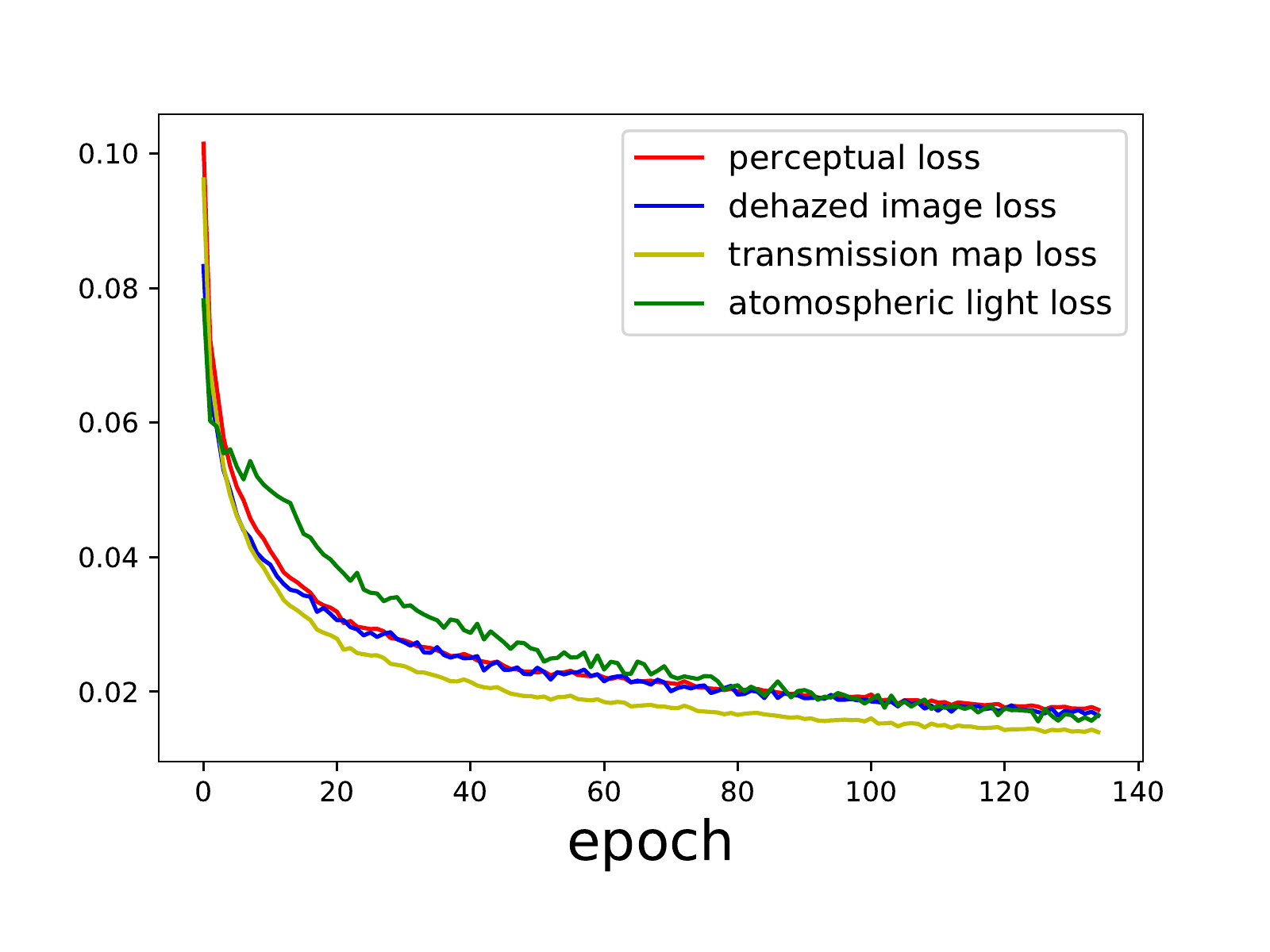}}
		\centerline{(c)} 
	\end{minipage} 
	\caption{Network convergence analysis. (a) Diagram of convergence process of PSNR with epoch. (b) Diagram of convergence process of SSIM with epoch. (c) The convergence process of the four loss functions with epoch, and the four loss functions correspond to Eq. (\ref{eq:1}), Eq. (\ref{eq:2}), Eq. (\ref{eq:3}) and Eq. (\ref{eq:4}).}
	\label{convergence} 
\end{figure*}

\subsection{Results on Synthetic Images}
We compare our method with previous state-of-the-art image dehazing methods both quantitatively and qualitatively, including DCP \cite{he2010single}, DehazeNet \cite{cai2016dehazenet}, AOD-Net \cite{li2017aod}, DCPDN \cite{zhang2018densely}, GFN \cite{ren2018gated}, PFFNet \cite{mei2018progressive}, EPDN \cite{qu2019enhanced}, FDU \cite{dong2020physics}, MSCNN \cite{ren2016single}. Most of them are data-driven methods except for DCP. We carry out these experiments on three datasets, RESIDE-Indoor, RESIDE-Outdoor and TrainA-TestA. 

\textbf{Quantitative Evaluation: }
Table \ref{SOTS-table} shows the quantitative comparisons of different methods on RESIDE-Indoor and RESIDE-outdoor. As shown in Table \ref{SOTS-table}, our method achieves the best performance with 33.01 dB PSNR and 0.98 SSIM compared with the other methods. FDU achieves the second-best performance on RESIDE-Indoor, which is 0.33 dB below us in terms of PSNR. We achieve the best SSIM on RESIDE-Outdoor and surpass the second 0.03. As mentioned above, our method outperforms previous methods on RESIDE in terms of SSIM and PSNR. 

Besides, we further conduct comparative experiments on TrainA-TestA. In Table \ref{TestA-table}, it is observed that our method is superior to the others, which achieves the best performance with 30.02 dB PSNR and 0.97 SSIM. TrainA-TestA is proposed by DCPDN which achieves the second-best based on the results reported in their paper, with 0.75 dB PSNR and 0.01 SSIM lower than us respectively.

\textbf{Qualitative Evaluation: }
Fig. \ref{SOTS} shows the subjective comparisons on RESIDE. The first three lines are RESIDE-Indoor images and the last three lines are RESIDE-Outdoor ones.
For DCP, we figure out that it alleviates the color distortion problem to some extent. Nevertheless, it tends to result in halo artifacts, which makes the dehazed images look artificial. It is observed that there remain lots of haze residuals and renders in the dehazed images of DehazeNet, AOD-Net and DCPDN. Although GFN and PFFNet improve the dehazing performance compared with the methods just mentioned, the effect of local dehazing is still incomplete. EPDN seems to have better dehazing results, but there is a local gap with ground truth because of its exposure. 
In contrast, our proposed method is the best one and it can generate more natural dehazed images with more realistic brightness and color fidelity. At the same time, the dehazed images produced by our method are free of major artifacts and are able to preserve more detail information. So the dehazed images restored by our method are more visually faithful to the ground truth.

Then we conduct the qualitative analysis of the dehazing effect on the TrainA-TestA dataset. As can be seen intuitively in Fig. \ref{TestA}, our method still shows the best effect on TrainA-TestA. We can observe that DCP still has the phenomenon of color distortion and the tone is rather dark. Both DehazeNet and MSCNN can not remove the haze completely, and the latter has the problem of color distortion. As we can see, GFN can improve the dehazing performance, but it is much darker than the ground truth, and the dehazing effect of local locations is still unsatisfactory. The dehazed images from AOD-Net still have a lot of haze residue and appear fuzzy phenomenon. Compared with other methods, the performance of DCPDN is relatively excellent, but there is still a certain gap between it and ground truth because of the color and details. Our dehazed images can be more realistic to the ground truth than these compared state-of-the-art methods.   

\subsection{Results on Real-world Images}
Fig. \ref{real} visually compares the dehazing effect on real-world hazy images from the URHI dataset. It can be seen that DCP also produces the color distortion problem for real hazy images. In the dehazed images of DehazeNet, MSCNN and AOD-Net, we can find a lot of unremoved haze. In contrast, the dehazing effect of DCPDN, GFN and EPDN is improved, but it is still not satisfactory. Especially, GFN makes local areas dark. Our method can ensure the dehazing effect and make the image more natural. In order to further demonstrate the effectiveness of the proposed DRN, we show the detail feature maps generated on the URHI dataset images in Fig. \ref{detail_map}. By enlarging the detail feature map, the learned image details can be clearly observed, which is beneficial for image reconstruction.

\subsection{Ablation Study}
To better demonstrate the superiority of our method, we conduct ablation experiments by considering different configurations of our network. The different configurations are given in Table \ref{ablation-table}, which simultaneously shows the corresponding performance in terms of SSIM and PSNR. (1) \textbf{Base:} Only a dehazing network. (2) \textbf{Base+DRN (w/o SDC):} Add DRN without SDC into Base. (3) \textbf{Ours:} Add DRN with SDC into the Base.

In order to illustrate the effectiveness of each component in the network, ablation experiment results are presented in Table \ref{ablation-table}. We first add DRN without SDC to Base and the results improved somewhat. Then, in order to verify the effectiveness of SDC, we replace the general dilated convolution in DRN with SDC, in which the experimental performance is optimal. As shown in Fig. \ref{ablation-image}, by comparing the images in the last three columns, we can find that DRN makes the dehazed images closer to the ground truth, but the output image has an exposure phenomenon which is brighter without SDC. All in all, larger gains can be achieved by combing all the designed components in our method.

And beyond that, we perform the ablation experiments to validate the necessity of the loss functions. From the results given in Table \ref{loss}, we can see that the perceptual loss contributes 0.21 dB PSNR and 0.0015 SSIM. The reconstruction loss further boosts the performance by 0.04 dB PSNR and 0.0006 SSIM. To sum up, the combination of loss functions ensures the effectiveness of haze removal.

\subsection{Convergence Analysis}
In Fig. \ref{convergence}, (a) and (b) show the convergence process of PSNR and SSIM with the epoch, respectively. The convergence process of the four loss functions with epoch is shown in (3). These figures show that our proposed method can converge at 140 epoch.

\section{Conclusion}
We propose a single image dehazing method with an independent detail-recovery network, which can give good consideration to both haze removal and detail recovery. The overall network is implemented through the independent DRN and the dehazing network. On the one hand, the independent DRN is exploited to recover the intrinsic image details contained in the hazy images, which consists of a local branch and a global branch. The local branch can obtain more local detail information through a convolution layer and the global branch can capture more global information by SDC with larger reception fields. On the other hand, the dehazing network is developed to generate the coarse dehazed image. It is composed of transmission map estimated network and atmospheric light estimated network, where the former build the encoder-decoder structure with residual connections and takes the dense block as the basic unit, and the latter one adopts the U-Net structure to estimate the atmospheric light. Finally, the high-quality dehazed image can be obtained through the enhancement of the detail feature map to the coarse dehazed image. Besides, we formulate the DRN, the physical-model-based dehazing network and the reconstruction loss into an end-to-end joint learning framework. Moreover, qualitative and quantitative experiment results demonstrate the superiority of our method in single image dehazing.

\section*{Acknowledgment}
This work is partially supported by the Joint Project for Smart Computing of Shandong Natural Science Foundation (ZR2020LZH015) and the National Natural Science Foundation of China (No. 62176198).

\bibliographystyle{IEEEtran}
\bibliography{IEEEexample}

\begin{thebibliography}{10}
\providecommand{\url}[1]{#1}
\csname url@samestyle\endcsname
\providecommand{\newblock}{\relax}
\providecommand{\bibinfo}[2]{#2}
\providecommand{\BIBentrySTDinterwordspacing}{\spaceskip=0pt\relax}
\providecommand{\BIBentryALTinterwordstretchfactor}{4}
\providecommand{\BIBentryALTinterwordspacing}{\spaceskip=\fontdimen2\font plus
\BIBentryALTinterwordstretchfactor\fontdimen3\font minus
  \fontdimen4\font\relax}
\providecommand{\BIBforeignlanguage}[2]{{%
\expandafter\ifx\csname l@#1\endcsname\relax
\typeout{** WARNING: IEEEtran.bst: No hyphenation pattern has been}%
\typeout{** loaded for the language `#1'. Using the pattern for}%
\typeout{** the default language instead.}%
\else
\language=\csname l@#1\endcsname
\fi
#2}}
\providecommand{\BIBdecl}{\relax}
\BIBdecl

\bibitem{min2020two}
W.~Min, S.~Mei, Z.~Li, and S.~Jiang, ``A two-stage triplet network training
  framework for image retrieval,'' \emph{IEEE Transactions on Multimedia},
  vol.~22, no.~12, pp. 3128--3138, 2020.

\bibitem{ren2020salient}
Q.~Ren, S.~Lu, J.~Zhang, and R.~Hu, ``Salient object detection by fusing local
  and global contexts,'' \emph{IEEE Transactions on Multimedia}, vol.~23, pp.
  1442--1453, 2020.

\bibitem{li2020bottom}
M.~Li, L.~Peng, T.~Wu, and Z.~Peng, ``A bottom-up and top-down integration
  framework for online object tracking,'' \emph{IEEE Transactions on
  Multimedia}, vol.~23, pp. 105--119, 2020.

\bibitem{yao2020adaptive}
X.~Yao, D.~She, H.~Zhang, J.~Yang, M.-M. Cheng, and L.~Wang, ``Adaptive deep
  metric learning for affective image retrieval and classification,''
  \emph{IEEE Transactions on Multimedia}, 2020.

\bibitem{he2010single}
K.~He, J.~Sun, and X.~Tang, ``Single image haze removal using dark channel
  prior,'' \emph{IEEE Transactions on Pattern Analysis and Machine
  Intelligence}, vol.~33, no.~12, pp. 2341--2353, 2010.

\bibitem{cai2016dehazenet}
B.~Cai, X.~Xu, K.~Jia, C.~Qing, and D.~Tao, ``Dehazenet: An end-to-end system
  for single image haze removal,'' \emph{IEEE Transactions on Image
  Processing}, vol.~25, no.~11, pp. 5187--5198, 2016.

\bibitem{wang2017fast}
W.~Wang, X.~Yuan, X.~Wu, and Y.~Liu, ``Fast image dehazing method based on
  linear transformation,'' \emph{IEEE Transactions on Multimedia}, vol.~19,
  no.~6, pp. 1142--1155, 2017.

\bibitem{ling2017optimal}
Z.~Ling, J.~Gong, G.~Fan, and X.~Lu, ``Optimal transmission estimation via fog
  density perception for efficient single image defogging,'' \emph{IEEE
  Transactions on Multimedia}, vol.~20, no.~7, pp. 1699--1711, 2017.

\bibitem{mccartney1976optics}
E.~J. McCartney, ``Optics of the atmosphere: scattering by molecules and
  particles,'' \emph{New York}, 1976.

\bibitem{tan2008visibility}
R.~T. Tan, ``Visibility in bad weather from a single image,'' \emph{IEEE
  Conference on Computer Vision and Pattern Recognition}, pp. 1--8, 2008.

\bibitem{shen2018iterative}
L.~Shen, Y.~Zhao, Q.~Peng, J.~C.-W. Chan, and S.~G. Kong, ``An iterative image
  dehazing method with polarization,'' \emph{IEEE Transactions on Multimedia},
  vol.~21, no.~5, pp. 1093--1107, 2018.

\bibitem{li2019pdr}
C.~Li, C.~Guo, J.~Guo, P.~Han, H.~Fu, and R.~Cong, ``Pdr-net:
  Perception-inspired single image dehazing network with refinement,''
  \emph{IEEE Transactions on Multimedia}, vol.~22, no.~3, pp. 704--716, 2019.

\bibitem{dong2020multi}
H.~Dong, J.~Pan, L.~Xiang, Z.~Hu, X.~Zhang, F.~Wang, and M.-H. Yang,
  ``Multi-scale boosted dehazing network with dense feature fusion,''
  \emph{IEEE Conference on Computer Vision and Pattern Recognition}, pp.
  2157--2167, 2020.

\bibitem{fattal2014dehazing}
R.~Fattal, ``Dehazing using color-lines,'' \emph{ACM Transactions on Graphics},
  vol.~34, no.~1, pp. 1--14, 2014.

\bibitem{zhu2015fast}
Q.~Zhu, J.~Mai, and L.~Shao, ``A fast single image haze removal algorithm using
  color attenuation prior,'' \emph{IEEE Transactions on Image Processing},
  vol.~24, no.~11, pp. 3522--3533, 2015.

\bibitem{berman2016non}
D.~Berman, S.~Avidan \emph{et~al.}, ``Non-local image dehazing,'' \emph{IEEE
  conference on Computer Vision and Pattern Recognition}, pp. 1674--1682, 2016.

\bibitem{li2017aod}
B.~Li, X.~Peng, Z.~Wang, J.~Xu, and D.~Feng, ``Aod-net: All-in-one dehazing
  network,'' \emph{IEEE International Conference on Computer Vision}, pp.
  4770--4778, 2017.

\bibitem{zhang2018multi}
H.~Zhang, V.~Sindagi, and V.~M. Patel, ``Multi-scale single image dehazing
  using perceptual pyramid deep network,'' \emph{IEEE Conference on Computer
  Vision and Pattern Recognition Workshops}, pp. 902--911, 2018.

\bibitem{liu2019griddehazenet}
X.~Liu, Y.~Ma, Z.~Shi, and J.~Chen, ``Griddehazenet: Attention-based
  multi-scale network for image dehazing,'' \emph{IEEE International Conference
  on Computer Vision}, pp. 7314--7323, 2019.

\bibitem{li2018single}
R.~Li, J.~Pan, Z.~Li, and J.~Tang, ``Single image dehazing via conditional
  generative adversarial network,'' \emph{IEEE Conference on Computer Vision
  and Pattern Recognition}, pp. 8202--8211, 2018.

\bibitem{ren2018gated}
W.~Ren, L.~Ma, J.~Zhang, J.~Pan, X.~Cao, W.~Liu, and M.-H. Yang, ``Gated fusion
  network for single image dehazing,'' \emph{IEEE Conference on Computer Vision
  and Pattern Recognition}, pp. 3253--3261, 2018.

\bibitem{engin2018cycle}
D.~Engin, A.~Gen{\c{c}}, and H.~Kemal~Ekenel, ``Cycle-dehaze: Enhanced cyclegan
  for single image dehazing,'' \emph{IEEE Conference on Computer Vision and
  Pattern Recognition Workshops}, pp. 825--833, 2018.

\bibitem{mao2016image}
X.~Mao, C.~Shen, and Y.-B. Yang, ``Image restoration using very deep
  convolutional encoder-decoder networks with symmetric skip connections,''
  \emph{Advances in Neural Information Processing Systems}, vol.~29, pp.
  2802--2810, 2016.

\bibitem{zhang2018densely}
H.~Zhang and V.~M. Patel, ``Densely connected pyramid dehazing network,''
  \emph{IEEE Conference on Computer Vision and Pattern Recognition}, pp.
  3194--3203, 2018.

\bibitem{li2020zero}
B.~Li, Y.~Gou, J.~Z. Liu, H.~Zhu, J.~T. Zhou, and X.~Peng, ``Zero-shot image
  dehazing,'' \emph{IEEE Transactions on Image Processing}, vol.~29, pp.
  8457--8466, 2020.

\bibitem{ronneberger2015u}
O.~Ronneberger, P.~Fischer, and T.~Brox, ``U-net: Convolutional networks for
  biomedical image segmentation,'' \emph{International Conference on Medical
  Image Computing and Computer-Assisted Intervention}, pp. 234--241, 2015.

\bibitem{ren2016single}
W.~Ren, S.~Liu, H.~Zhang, J.~Pan, X.~Cao, and M.-H. Yang, ``Single image
  dehazing via multi-scale convolutional neural networks,'' \emph{European
  conference on Computer Vision}, pp. 154--169, 2016.

\bibitem{li2020deep}
P.~Li, J.~Tian, Y.~Tang, G.~Wang, and C.~Wu, ``Deep retinex network for single
  image dehazing,'' \emph{IEEE Transactions on Image Processing}, vol.~30, pp.
  1100--1115, 2020.

\bibitem{gao2018detail}
Y.~Gao, H.-M. Hu, B.~Li, Q.~Guo, and S.~Pu, ``Detail preserved single image
  dehazing algorithm based on airlight refinement,'' \emph{IEEE Transactions on
  Multimedia}, vol.~21, no.~2, pp. 351--362, 2018.

\bibitem{ju2019idgcp}
M.~Ju, C.~Ding, Y.~J. Guo, and D.~Zhang, ``Idgcp: Image dehazing based on gamma
  correction prior,'' \emph{IEEE Transactions on Image Processing}, vol.~29,
  pp. 3104--3118, 2019.

\bibitem{qu2019enhanced}
Y.~Qu, Y.~Chen, J.~Huang, and Y.~Xie, ``Enhanced pix2pix dehazing network,''
  \emph{IEEE Conference on Computer Vision and Pattern Recognition}, pp.
  8160--8168, 2019.

\bibitem{fattal2008single}
R.~Fattal, ``Single image dehazing,'' \emph{ACM Transactions on Graphics},
  vol.~27, no.~3, pp. 1--9, 2008.

\bibitem{mei2018progressive}
K.~Mei, A.~Jiang, J.~Li, and M.~Wang, ``Progressive feature fusion network for
  realistic image dehazing,'' \emph{Asian Conference on Computer Vision}, pp.
  203--215, 2018.

\bibitem{qin2020ffa}
X.~Qin, Z.~Wang, Y.~Bai, X.~Xie, and H.~Jia, ``Ffa-net: Feature fusion
  attention network for single image dehazing,'' \emph{AAAI Conference on
  Artificial Intelligence}, vol.~34, no.~07, pp. 11\,908--11\,915, 2020.

\bibitem{dong2020physics}
J.~Dong and J.~Pan, ``Physics-based feature dehazing networks,'' \emph{European
  Conference on Computer Vision}, pp. 188--204, 2020.

\bibitem{dong2020fd}
Y.~Dong, Y.~Liu, H.~Zhang, S.~Chen, and Y.~Qiao, ``Fd-gan: Generative
  adversarial networks with fusion-discriminator for single image dehazing,''
  \emph{AAAI Conference on Artificial Intelligence}, vol.~34, no.~07, pp.
  10\,729--10\,736, 2020.

\bibitem{deng2020detail}
S.~Deng, M.~Wei, J.~Wang, Y.~Feng, L.~Liang, H.~Xie, F.~L. Wang, and M.~Wang,
  ``Detail-recovery image deraining via context aggregation networks,''
  \emph{IEEE Conference on Computer Vision and Pattern Recognition}, pp.
  14\,560--14\,569, 2020.

\bibitem{lin2013network}
M.~Lin, Q.~Chen, and S.~Yan, ``Network in network,'' \emph{arXiv preprint
  arXiv:1312.4400}, 2013.

\bibitem{wang2018smoothed}
Z.~Wang and S.~Ji, ``Smoothed dilated convolutions for improved dense
  prediction,'' \emph{ACM Conference on Knowledge Discovery and Data Mining},
  pp. 2486--2495, 2018.

\bibitem{johnson2016perceptual}
J.~Johnson, A.~Alahi, and L.~Fei-Fei, ``Perceptual losses for real-time style
  transfer and super-resolution,'' \emph{European Conference on Computer
  Vision}, pp. 694--711, 2016.

\bibitem{simonyan2014very}
K.~Simonyan and A.~Zisserman, ``Very deep convolutional networks for
  large-scale image recognition,'' \emph{arXiv preprint arXiv:1409.1556}, 2014.

\bibitem{ILSVRC15}
O.~Russakovsky, J.~Deng, H.~Su, J.~Krause, S.~Satheesh, S.~Ma, Z.~Huang,
  A.~Karpathy, A.~Khosla, M.~Bernstein, A.~C. Berg, and L.~Fei-Fei, ``Imagenet
  large scale visual recognition challenge,'' \emph{International Journal of
  Computer Vision}, vol. 115, no.~3, pp. 211--252, 2015.

\bibitem{lim2017enhanced}
B.~Lim, S.~Son, H.~Kim, S.~Nah, and K.~Mu~Lee, ``Enhanced deep residual
  networks for single image super-resolution,'' \emph{IEEE Conference on
  Computer Vision and Pattern Recognition Workshops}, pp. 136--144, 2017.

\bibitem{huynh2008scope}
Q.~Huynh-Thu and M.~Ghanbari, ``Scope of validity of psnr in image/video
  quality assessment,'' \emph{Electronics Letters}, vol.~44, no.~13, pp.
  800--801, 2008.

\bibitem{wang2004image}
Z.~Wang, A.~C. Bovik, H.~R. Sheikh, and E.~P. Simoncelli, ``Image quality
  assessment: from error visibility to structural similarity,'' \emph{IEEE
  Transactions on Image Processing}, vol.~13, no.~4, pp. 600--612, 2004.

\bibitem{li2019benchmarking}
B.~Li, W.~Ren, D.~Fu, D.~Tao, D.~Feng, W.~Zeng, and Z.~Wang, ``Benchmarking
  single-image dehazing and beyond,'' \emph{IEEE Transactions on Image
  Processing}, vol.~28, no.~1, pp. 492--505, 2019.

\bibitem{silberman2012indoor}
N.~Silberman, D.~Hoiem, P.~Kohli, and R.~Fergus, ``Indoor segmentation and
  support inference from rgbd images,'' \emph{European Conference on Computer
  Vision}, pp. 746--760, 2012.

\bibitem{scharstein2003high}
D.~Scharstein and R.~Szeliski, ``High-accuracy stereo depth maps using
  structured light,'' \emph{IEEE Conference on Computer Vision and Pattern
  Recognition}, vol.~1, pp. I--I, 2003.

\bibitem{kingma2014adam}
D.~P. Kingma and J.~Ba, ``Adam: A method for stochastic optimization,''
  \emph{arXiv preprint arXiv:1412.6980}, 2014.

\end{thebibliography}


\end{document}